\documentclass[11pt,twoside]{book}
\usepackage[T1]{fontenc}
\usepackage[dvips]{graphicx} 
\usepackage{epsfig}
\usepackage{latexsym}
\usepackage{amssymb}
\usepackage{amsmath}

\usepackage[spanish,USenglish]{babel}
\usepackage[ansinew]{inputenc}
\usepackage{makeidx}
\usepackage{algorithmic}
\usepackage[]{algorithm2e}
\usepackage[usenames]{color}
\usepackage{datetime}

\usepackage{url}
\usepackage{indentfirst}
\usepackage{helvet}
\usepackage{mathrsfs}
\usepackage[dvipsnames,table,xcdraw]{xcolor}            
\usepackage{subfigure}        
\usepackage{amsfonts}          
\usepackage{lineno}
\usepackage{epsfig}
\usepackage{gensymb}
\usepackage{indentfirst}
\usepackage{acronym}

\usepackage{amsmath,epsfig}
\usepackage{float}

\usepackage{booktabs}           
\usepackage{multirow}           
\usepackage{array,hhline}   

\usepackage{subfigure}

\usepackage{lmodern}

\usepackage{fancyhdr}

\usepackage{multicol}

\newcommand\myeq{\stackrel{\text{\tiny def}}{=}}

\makeindex

\begin{document}
\sloppy
%
%
\newcommand{\paginalimpia}{\clearpage{\pagestyle{empty} \cleardoublepage}}

\makeatletter
\def\cleardoublepage{\clearpage\if@twoside \ifodd\c@page\else
\hbox{} \vspace*{\fill} \vspace{\fill} \thispagestyle{empty}
\newpage
\if@twocolumn\hbox{}\newpage\fi\fi\fi}
\makeatother

\pagestyle{empty}

\begin{figure}[t]
    \begin{center}
      \includegraphics[width=.4\textwidth]{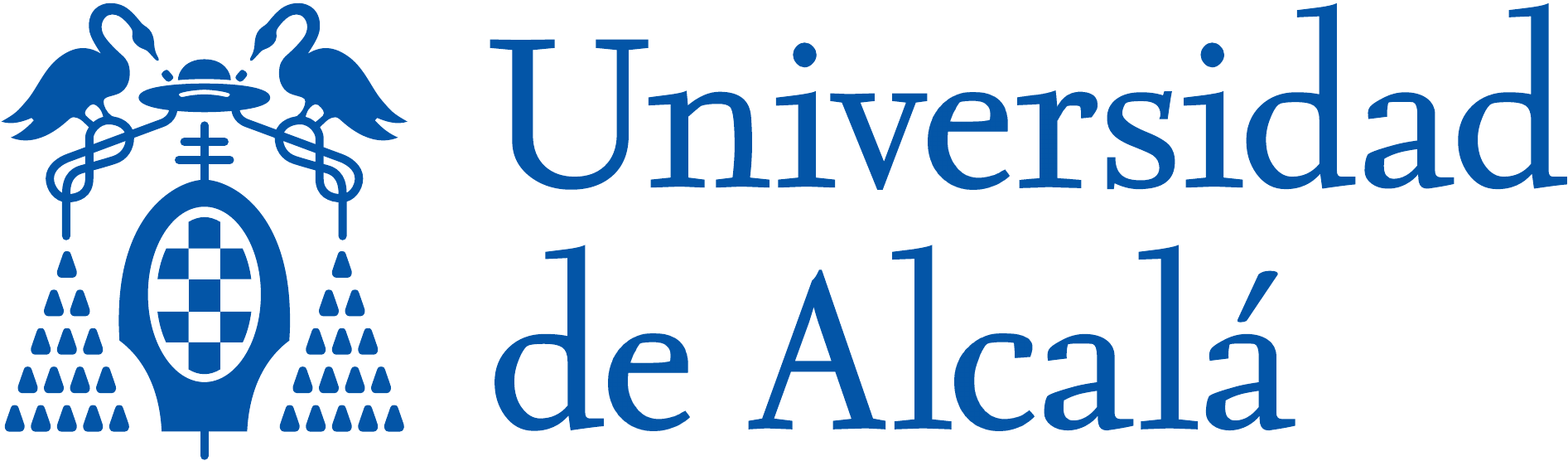}\\
    \end{center}
\end{figure}

\begin{center}
\textbf{\LARGE \textcolor{Blue}{Programa de Doctorado en Tecnolog\'ias de la Informaci\'on y las Comunicaciones}}\\
\vspace{2.0cm}
\huge \textbf{\textsc{\textcolor{Blue}{{NEW HYBRID NEURO-EVOLUTIONARY ALGORITHMS FOR RENEWABLE ENERGY AND FACILITIES MANAGEMENT PROBLEMS}}}}\\
\vspace{2.0cm}
\textbf{\LARGE \textcolor{Blue}{Tesis Doctoral presentada por}}\\
\vspace{2.0cm}
\textbf{\LARGE \textcolor{Blue}{LAURA M\textordfeminine\hspace{0.008cm} CORNEJO BUENO}}\\
\vspace{2.0cm}
\end{center}

\begin{flushleft}
\textbf{Director:}\\
\textbf{DR. SANCHO SALCEDO SANZ}\\
\vspace{2.0cm}
\end{flushleft}

\begin{center}
\textbf{Alcal\'a de Henares, 2018}\\
\end{center}
\paginalimpia

\vspace*{8.0cm}
\begin{flushright}
\emph{No te rindas, por favor no cedas, aunque el fr\'io queme, aunque el miedo muerda, aunque el sol se esconda y se calle el viento, a\'un hay fuego en tu alma, a\'un hay vida en tus sue\~{n}os.}\\
\vspace{1cm}
\emph{Mario Benedetti}\\
\end{flushright}
\paginalimpia

\chapter*{Abstract}\label{Abstract}
This Ph.D. thesis deals with the optimization of several renewable energy resources development as well as the improvement of facilities management in oceanic engineering and airports, using computational hybrid methods belonging to \ac{AI} to this end. These problems will be summarized hereafter with the technical solutions proposed at the end of the section.

Energy is essential to our society in order to ensure a good quality of life. Nowadays, fossil fuels are the most important energy source in the planet. However they contribute to Climate Change greatly, affecting the ecosystems severely. For this reason, there is a trend to gradually use renewable resources which guarantee a sustainable development. In fact, a penetration of these resources over the 50\% are expected in fifty years. Obviously, that process shall not be the same in all countries due to renewable energy resources are not uniformly distributed throughout the World. It is important to note that different regions rely on different renewable technologies, so they can contribute to regional development in a larger or lesser degree. The main drawback of renewable energies is the natural variability inherent to the resource. This means that predictions over the characteristics on which renewable energies depend are necessary, in order to know the amount of energy that will be obtained at any time.

The second topic tackled in this thesis is related to the basic parameters that influence in different marine activities and airports, whose knowledge is necessary to develop a proper facilities management in these environments. For instance, the \ac{$H_s$} is a basic parameter in wave characterization, important to different problems in marine activities such as the design and management of vessels, marine structures, \ac{WECs}, etc. On the other hand, the low-visibility events at airports, normally caused by fog events, is another fundamental issue in airport activities which can cause flight delays, diversions and cancellations or accidents in the worst cases.

Within this work, a study of the state-of-the-art \ac{ML} have been performed to solve the problems associated with the topics above-mentioned, and several contributions have been proposed:
\begin{itemize}
  \item One of the pillars of this work is focused on the estimation of the most important parameters in the exploitation of renewable resources. For this purpose, \ac{SVR}, \ac{NN} (\ac{MLP} and \ac{ELM}) and \ac{GP} algorithms are used in several practical problems. The performance of these algorithms is discussed in every experiment carried out, and also the specific settings of the algorithms, as well as internal characteristics of the models.
  \item The second contribution of this thesis is related to feature selection problems. More specifically, the use of EAs as \ac{GGA} or \ac{CRO} hybridized with others \ac{ML} approaches as classifiers and regressors. Regarding this, the \ac{GGA} or \ac{CRO} looks for several subsets of features important to solve the problem, and the regressor employed provides the prediction in terms of the features selected by the \ac{GA}, reducing the computational cost with a good accuracy.
\end{itemize}

The proposed methodologies are applied to multiple problems: the prediction of $H_s$, relevant for marine energy applications and marine activities, the estimation of \ac{WPREs}, undesirable variations in the electric power produced by a wind farm, the prediction of global solar radiation in areas from Spain and Australia, really important in terms of solar energy, and the prediction of low-visibility events at airports. All of these practical issues are developed with the consequent previous data analysis, normally, in terms of meteorological variables.

\chapter*{Resumen en Castellano}\label{Resumen}
Esta tesis tiene como objetivo la optimizaci\'on de la explotaci\'on de recursos energ\'eticos renovables, as\'i como la mejora en la gesti\'on de instalaciones en ingenier\'ia oce\'anica y aeropuertos, usando m\'etodos computacionales h\'ibridos pertenecientes a una rama de la Inteligencia Artificial (IA), denominada aprendizaje m\'aquina, para este fin. Estos problemas ser\'an resumidos a continuaci\'on con las soluciones t\'ecnicas propuestas al final de la secci\'on.

La energ\'ia es esencial en nuestra sociedad para asegurar una buena calidad de vida. Hoy en d\'ia, los combustibles f\'osiles constituyen la fuente energ\'etica m\'as importante del planeta, sin embargo, estas formas de energ\'ia contribuyen al Cambio Clim\'atico en gran medida, afectando los ecosistemas severamente. Por esta raz\'on, se tiende gradualmente al uso de fuentes de energ\'ia renovables que garanticen un desarrollo sostenible. De hecho, se prev\'e en 50 a\~nos, una penetraci\'on de estos recursos por encima del 50\%. Obviamente, este proceso no ser\'a igual en todos los pa\'ises, debido a que las fuentes de energ\'ia renovables no est\'an uniformemente distribuidas a lo largo del mundo. El hecho m\'as importante es que cada \'area cuenta con alguna de ellas, y pueden contribuir al desarrollo regional en mayor o menor medida, gracias a lo cual las fuentes de energ\'ia convencionales ser\'an sustituidas progresivamente.
Sin embargo, se observa un lento desarrollo en este sentido, y la \'unica cuesti\'on que cabe preguntarse es cu\'ando las energ\'ias renovables tendr\'an mayor penetraci\'on en el sistema que los actuales combustibles f\'osiles. Para responder a esta pregunta, una buena manera es centrarse en el principal inconveniente de este tipo de energ\'ias: la variabilidad natural inherente al recurso. Esto significa que las predicciones sobre los par\'ametros m\'as importantes de los que dependen las energ\'ias renovables son necesarias para conocer la cantidad de energ\'ia que ser\'a obtenida en un momento dado.

El otro tema abordado en esta tesis est\'a relacionado con los par\'ametros que influyen en diferentes actividades marinas y aeropuertos, cuyo conocimiento de su comportamiento es necesario para desarrollar una correcta gesti\'on de las instalaciones en estos entornos. Por ejemplo, la altura significativa de las olas ($H_s$) es un par\'ametro b\'asico en la caracterizaci\'on de las olas, muy importante para el desarrollo de actividades marinas como el dise\~no y mantenimiento de barcos, estructuras marinas, convertidores energ\'eticos de ola, etc. Por otro lado, la escasa visibilidad en los aeropuertos, normalmente causada por la niebla, es otro aspecto fundamental para el correcto desarrollo de actividades aeroportuarias, y que puede causar retrasos en los vuelos, desv\'ios y cancelaciones, o accidentes en el peor de los casos.

En este trabajo se ha realizado un an\'alisis del estado del arte de los modelos de aprendizaje m\'aquina que se utilizan actualmente, con el objetivo de resolver los problemas asociados a los temas tratados con anterioridad. Diferentes contribuciones han sido propuestas:
\begin{itemize}
  \item Uno de los pilares esenciales de este trabajo est\'a centrado en la estimaci\'on de los par\'ametros m\'as importantes en la explotaci\'on de energ\'ias renovables. Con este prop\'osito, los algoritmos Vectores Soporte para Regresi\'on (VSR), Redes Neuronales (RN) (Perceptrones Multicapa (MLP) y M\'aquinas de Aprendizaje Extremo (MAE)) y Procesos Gaussianos son utilizados en diferentes problemas pr\'acticos.  El rendimiento de estos algoritmos es analizado en cada uno de los experimentos realizados, tanto la precisi\'on de los mismos como la especificaci\'on de las caracter\'isticas internas.
  \item Otro de los aspectos tratados est\'a relacionado con problemas de selecci\'on de caracter\'isticas. Concretamente, con el uso de algoritmos evolutivos como Algoritmos de Agrupaci\'on Gen\'etica (AAG) o los algoritmos de Optimizaci\'on de Arrecife de Coral (OAC) hibridizados con otros m\'etodos de aprendizaje m\'aquina como  clasificadores y regresores. En este sentido, el AAG o OAC analizan diferentes conjuntos de caracter\'isticas para obtener aquel que resuelva el problema con la mayor precisi\'on, y el regresor empleado proporciona la predicci\'on en funci\'on de las caracter\'isticas obtenidas por el Algoritmo Gen\'etico (AG), reduciendo el coste computacional con gran fiabilidad en los resultados.
\end{itemize}

La metodolog\'ia mencionada es aplicada a m\'ultiples problemas: predicci\'on de $H_s$, relevante en aplicaciones energ\'eticas y actividades marinas, estimaci\'on de eventos puntuales como son las rampas de viento (ERV), variaciones indeseables en la potencia el\'ectrica producidas por un parque e\'olico, predicci\'on de la radiaci\'on solar global en \'areas de Espa\~na y Australia, realmente importante en t\'erminos de energ\'ia solar, y la estimaci\'on de eventos de baja visibilidad en aeropuertos. Los casos pr\'acticos citados son desarrollados con el consecuente an\'alisis previo de la base de datos empleada, normalmente, en t\'erminos de variables meteorol\'ogicas.
\paginalimpia

\chapter*{Agradecimientos}
Seguramente necesitar\'ia otro libro para expresar el enorme agradecimiento que siento hacia todos vosotros. El esfuerzo y esp\'iritu de sacrificio siempre dan su fruto, pero una cosa es segura y es la importancia de poder contar con el apoyo incondicional de las personas que tienes alrededor. Porque de una forma u otra todos aport\'ais vuestro granito de arena y contribu\'is a que hoy pueda seguir creciendo como estudiante y como persona.

En primer lugar quiero agradecer a mi Director de Tesis, Sancho Salcedo la confianza depositada en m\'i. Por ense\~narme a ganar seguridad en uno mismo desde la mejor de las humildades, y por supuesto por ense\~narme tanto y tan constante, porque sin \'el est\'a claro que todo este trabajo no habr\'ia visto la luz. Gracias por todo el apoyo y por hacer de esta etapa una de las mejores vividas hasta el momento.

Tambi\'en quiero agradecer a Enrique Alexandre, Silvia Jim\'enez, Jos\'e Antonio Portilla, Lucas Cuadra, Jos\'e Carlos Nieto y Raquel Criado el haberme acogido como una m\'as, y permitirme aprender tanto de ellos. Porque adem\'as de poder trabajar en lo que te entusiasma, es un gusto poder hacerlo en un ambiente tan agradable como el que consegu\'is crear en el laboratorio.

Y a Carlos Casanova con el que he tenido el placer de poder trabajar codo con codo en uno de los art\'iculos de esta tesis, y cuya colaboraci\'on ha sido crucial para su publicaci\'on. Aprovecho tambi\'en para agredecer a todas las personas que me he cruzado en estos a\~nos y de las que he podido tanto aprender como coloborar en numerosos trabajos.

Por supuesto a mis ``mindundis'' Carlos Camacho, Freddy Pinto y Adri\'an Aybar, mis compis de fatiga. Gracias por las innumerables comidas, caf\'es, charlas y quedadas; por ese intercambio de conocimiento y sobre todo por las risas dif\'iciles de olvidar. !`Chicos ya se va viendo la luz al final del t\'unel!
Y a mi compa\~nera de la planta de arriba Inma Mohino, cuya sonrisa te alegra el d\'ia.

El doctorado adem\'as me ha permitido vivir una de las mejores experiencias de mi vivida. Mi estancia de 3 meses en Australia. All\'i cont\'e con el apoyo del Profesor Ravinesh C. Deo, quien me recibi\'o con los brazos abiertos y contriby\'o en mi formaci\'on. Adem\'as aprend\'i que se puede conocer el verdadero significado de amistad aunque 2 personas est\'en separadas por m\'as de 17.000 km. Kavina Dayal gracias por convertir esta estancia en algo inolvidable; nos vemos en alguna parte del mundo.

Quer\'ia agradecer tambi\'en a una persona muy especial, a un amigo que me conoce desde mucho antes de estar aqu\'i y que me ha apoyado tanto desde dentro. Enrique Garc\'ia, Kike, gracias por tus visitas, por la alegr\'ia que consigues despertarme a\'un en los momentos que parec\'ian no tenerla. Ha sido muy importante poder contar tan de cerca con alguien de mi familia, alguien como t\'u.

Y ahora s\'i, las personas que me han visto crecer, y que tanto han cre\'ido en mi, incluso ni cuando yo misma cre\'ia.

Quiero empezar por la persona que me lo ha dado todo, mi madre, Carmen Bueno. No se puede explicar con palabras todo lo que te debo. Mil gracias por estar ah\'i al pie del ca\~n\'on y sacar fuerzas de donde no las hay para mostrar siempre una sonrisa. En especial quiero destacar tu enorme valor y la fuerza que has demostrado siempre, sobre todo frente a la adversidad de este \'ultimo a\~no. Eres toda una inspiraci\'on y verte me hace sentir que puedo ser capaz de cualquier cosa. Eres mi luz.

A mi padre, Juan Carlos Cornejo, que no ha dejado de trabajar ni un solo d\'ia para que hoy haya podido llegar hasta aqu\'i. Gracias por todo tu esfuerzo y voluntad, y por formar parte de lo que somos mi hermana y yo.

Mi peque\~na hermanita, Sara Cornejo, que es muy grande. No creo que haya alguien que pueda conocerme mejor. Siempre est\'as pendiente de lo que necesito en cada momento, cuando la hermana mayor soy yo. Siempre sabes qu\'e decir, y tus consejos nunca pueden ser m\'as acertados. No podr\'ia imaginarme una vida sin t\'i, porque no habr\'ia una sin una de las partes. Gracias por tu condici\'on humana y por hacer que no me sienta sola por muy lejos que estemos la una de la otra. Tu fuerza tambi\'en hace que hoy pueda decir, !`he llegado!, !`estoy aqu\'i!.

A David Do\~noro, mi pilar, mi compa\~nero de viaje en esta aventura. Son 8 a\~nos los que llevo a tu lado y consigues hacerme sentir como si estuvi\'eramos empezando cada d\'ia. Es reconfortante poder llegar a ``mi sitio'' y sentirme en casa. Gracias por creer en mi, por ser fuerte cuando lo necesito, y por no dejarme caer. Juntos podemos con lo que nos echen.

Por supuesto a mi yayi, Teresa Montes, una luchadora innata, un ejemplo de vida. Una persona que es capaz de transmitir AMOR en el m\'as profundo sentido de la palabra. Te debemos todo, y no creo que podamos estar m\'as orgullosos de tener una madre, esposa, abuela y bisabuela como T\'U. Gracias por no decaer y seguir a nuestro lado con tanto tes\'on.

Y a mi otra abuela, Magdalena Mac\'ias, cuya pasi\'on por los estudios nos anim\'o siempre a seguir luchando por nuestro futuro. Gracias por ser igualmente una luchadora de esta vida, y mantenerte entera pese a todo lo vivido. Eres un ejemplo de constancia.

A mis t\'ios Francisco Jos\'e L\'opez, Teresa Bueno e Isabel Bueno, gracias por hacer que pueda contar con vosotros y estar a mi lado en este camino. Destacar las comiditas de la tita Beli, que tanto ayudan cuando no hay tiempo ni de cocinar, los consejos de la tita Mari, y las provechas conversaciones del tito Francisco.

No puedo olvidarme de mis primos, Jos\'e Gabriel del Prado, Eduardo Gonz\'alez, Israel Gonz\'alez y Jes\'us del Prado, que no son primos sino hermanos. Gracias por toda esta vida de cari\~no y diversi\'on, sois parte esencial de este camino. Y como no Mariv\'i, Ana y Andrea que se han convertido en las mejores primas inesperadas que se puede tener.

Y siguiendo la l\'inea sucesoria es turno de mis sobrinitos. Isabel, F\'atima, Gabriel y Alejandro, las personitas m\'as peque\~nas y que m\'as pueden llenar de luz un d\'ia gris.

Como hay una que s\'i sabe leer, quer\'ia dedicarle unas palabras, pues creo que no puede imaginarse como me cambi\'o la vida. Isabel eres mi motor, el empuje ma\~nanero que me anima cada d\'ia. Ni loca me perder\'ia esas conversaciones de ni\~na de 7 a\~nos a adulto en las que a veces dudo de qui\'en es el adulto. Porque aunque digas ``eres la mejor tita del mundo'', y reconozco que me derrito cada vez que te escucho, eres t\'u la \'unica capaz de hacer olvidar todo lo malo de alrededor, y encima lo haces sin darte cuenta. No he podido tener m\'as suerte contigo, y solo quisiera poder transmitirte la mitad de lo que t\'u me das. No te rindas nunca y lucha, lucha porque yo siempre estar\'e a tu lado apoy\'andote como t\'u (siendo tan peque\~na) has hecho conmigo. Recuerda, nada es imposible. Te quiero.

Y no puedo terminar esta parte sin agradecer enormemente a los que por desgracia no han podido verme acabar. Mi abuelo, Jos\'e Luis Cayuela, y Manuel Garc\'ia. Me quedo con todo lo que me hab\'eis ense\~nado, que es mucho, ech\'andoos de menos cada d\'ia, pero agradeciendo enormemente el haber coincidido en esta vida, y que hay\'ais formado parte de mi familia. Padri nadie podr\'a llamarme ``chata'' de la forma en que t\'u lo hac\'ias, y no es comprable a nada la forma burlona de llamar a ``la Lauri'' que t\'u Manuel ten\'ias. Gracias por vuestro AMOR. Os quiero.

A mi otra familia, Cati, Pablo, Estefan\'ia, Juli\'an, To\~ni, Jes\'us \'Angel, Juan Carlos, Abuelos, Manolo y desde el m\'as profundo cari\~no Grego. Por hacerme sentir parte de vuestras vidas y contribuir con vuestro cari\~no y valores desde que comenc\'e esta etapa tan importante. Gracias por estar a mi lado y acogerme como lo hic\'isteis.

Y no pod\'ia olvidarme de vosotros, mis amigos y compa\~neros desde que empezamos la carrera. Casi 10 a\~nos ya y tan unidos como al principio. Casillas, Pascu, Dan, Jenny, Gallo, Guille, Sara, Samu, Paloma, Jes\'us, Manu, Mar, Pastor, Alvarito, V\'ictor, Jesica y Susana. !`Gracias! Porque sab\'eis lo importante que sois, y hab\'eis demostrado estar en todo momento. !`Qu\'e aburridos habr\'ian sido los d\'ias sin vosotros! Espero que mantegamos esta bonita amistad por muchos a\~nos m\'as.

Por \'ultimo quiero dar las gracias a ``las ni\~nas del cole'' Marta, Rosana y Lidia, con las que he compartido mi ni\~nez y adolescencia y con las que sigo creciendo y afrontando etapas. Despu\'es de casi 20 a\~nos es \'increible poder contar con amigas como vosotras.
Y a Roc\'io, siempre la vecinita. Por todas las tardes de estudio que nos ha amenizado con su alegr\'ia, y ser todo un apoyo por muchos d\'ias que pasen sin que nos veamos.

Siento si me dejo a alguien, pero esto es gracias a TODOS, a los que aparec\'eis y a los que no he puesto. Porque a lo largo de los a\~nos se conocen muchas personas que dejan huella y forman parte de lo que ahora somos. Qui\'en sabe cu\'ando volver\'e a escribir un libro, al menos en \'este puedo reflejar el esfuerzo de muchos a\~nos y el fruto obtenido, que tambi\'en es vuestro.

A todos, OS QUIERO.
\paginalimpia

\pagenumbering{roman}


\tableofcontents
\listoffigures
\listoftables
\paginalimpia
\chapter*{LIST OF ACRONYMS}
\addcontentsline{lof}{section}{List of acronyms}

\begin{acronym}
\acro{AI}{Artificial Intelligent}
\acro{ARMA}{Autoregressive-Moving-Average}
\acro{ANN}{Artificial Neural Network}
\acro{BO}{Bayesian Optimization}
\acro{CI}{Computational Intelligence}
\acro{CNN}{Convolutional Neural Network}
\acro{CRO}{Coral Reef Optimization}
\acro{DFT}{Discrete Fourier Transform}
\acro{EA}{Evolutionary Algorithm}
\acro{EC}{Evolutionary Computation}
\acro{ELM}{Extreme-Learning Machine}
\acro{EI}{Expected Improvement}
\acro{EV}{Electric Vehicles}
\acro{FC}{Fuzzy Computation}
\acro{FFT}{Fast Fourier Transform}
\acro{FL}{Fuzzy Logic}
\acro{FS}{Feature Selection}
\acro{GA}{Genetic Algorithm}
\acro{GGA}{Grouping Genetic Algorithm}
\acro{GP}{Gaussian process}
\acro{GS}{Grid Search}
\acro{MAE}{Mean Absolute Error}
\acro{ML}{Machine Learning}
\acro{MLP}{Multi-Layer Perceptron}
\acro{MSE}{Mean Squared Error}
\acro{NC}{Neural Computation}
\acro{NN}{Neural Network}
\acro{RMSE}{Root Mean Squared Error}
\acro{SAR}{Synthetic Aperture Radar}
\acro{SC}{Soft-Computing}
\acro{SM}{Standard Method}
\acro{SVM}{Support Vector Machine}
\acro{SVR}{Support-Vector Regression}
\acro{$H_s$}{Significant Wave Height}
\acro{V2G}{Vehicle-to-Grid}
\acro{$P$}{Wave Energy Flux}
\acro{WECs}{Wave Energy Converters}
\acro{WPF}{Wind Power Forecasting}
\acro{WPREs}{Wind Power Ramps Events}

\end{acronym}
\paginalimpia

\pagenumbering{arabic}
\pagestyle{fancy}
\fancyhf{}
\fancyhf{}
\fancyhead[LO]{\rightmark} 
\fancyhead[RE]{\leftmark} 
\fancyhead[RO,LE]{\thepage} 
\renewcommand{\chaptermark}[1]{\markboth{{Chapter \thechapter. #1}}{}} 
\renewcommand{\sectionmark}[1]{\markright{{\thesection. #1}}} 
\renewcommand{\headrulewidth}{0.0pt}
\setlength{\headheight}{15pt}

\part{Motivation and state-of-the-art}\label{part:introduccion}
\chapter{Introduction}\label{cap:introduction}
\section{Motivation}
The challenges of renewable energies in the near future, as well as the associated facilities, will require new computational tools for the optimization and exploitation of the available resources. In a World where the climate change is a recognized fact, it is necessary a re-evaluation of the energy use. For this reason, this work is focused on renewable energy sources, with almost zero emissions of both air pollutants and greenhouse gases. Currently, renewable energy sources supply 19\% of the total world energy demand \cite{REN21}, but it is expected an increasing in the near future. In consequence, a sustainable development must be guaranteed, defined by the World Commission on Environment and Development as ``development that meets the needs of the present without compromising the ability of future generations to meet their own needs''. The main goal then is to conciliate energy production, that satisfies social welfare, and the environmental protection, achieving economic growth.

The technology is the best way to meet the objectives proposed. There are many renewable energy technologies but most of them are still at an early stage of development and not technically mature. The aim of this work is to contribute to the progress in this field by means of AI.

AI is a term that indicates in its broadest sense the ability of a machine to perform the human learning. Specifically, it is the part of the computer science tasked with the design of intelligent computer systems. This kind of intelligence can be associated with human behavior, i.e., understanding, language, learning \cite{Kal06} and whose skills can be applied in diverse applications in forecasting, patter recognition, optimization and many more. That is possible because AI covers different areas like \ac{NC}, \ac{EC} and \ac{FC}, among others, that can be used or hybridized to solve several problems in our society.

Some of these algorithms are used, in this work, in the estimation of really important parameters in renewable energy area, taking into account not only the attainment of energy but also how these parameters can affect in determined tasks of facilities management. In this regard, facilities management are related with the necessary infrastructures in renewable energy environments and another fields where meteorological variables affect in the same way, as the study developed in airports to estimate the visibility in runways which will be explained in depth in Part \ref{part:aplicaciones2}.

Two fields compose the core of this Ph.D Thesis: \ac{ML} regression algorithms and evolutionary optimization. Pattern recognition is a branch of AI focused on systems that are able to associate multidimensional data to labels. Using this method it will be possible to develop others systems based on the available data to obtain predictions and classifications in many fields in Engineering, Sciences, Economy, etc. The second pillar of this thesis is the use of the \ac{EC} approaches to solve features selection problems and thus optimize the accuracy in the future regressions.

In the next sections it will be provided a more detailed description of the \ac{ML} techniques applied in this work as well as \ac{EA}, providing a review of the state-of-the-art in \ac{SC} techniques.

\section{State of the art} \label{sec:state_of_the_art}
\begin{figure}[ht]
  \centering
    \includegraphics[width=0.85\textwidth]{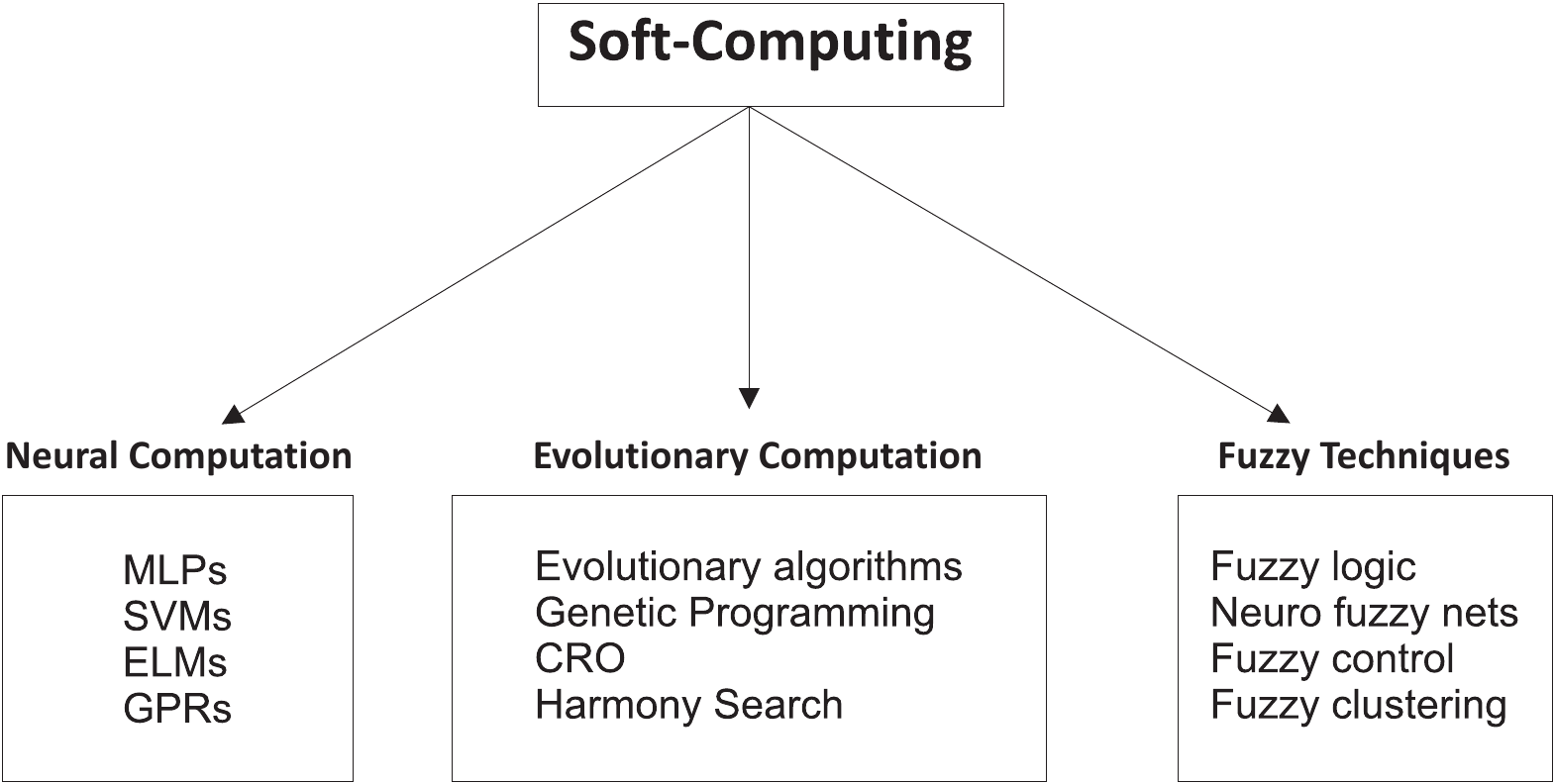}
  \caption{Structure of SC, including \ac{NC}, EC and FC.}
  \label{soft_computing}
\end{figure}

This section presents a description of the state-of-the-art in the technological fields addressed in this thesis. Figure \ref{soft_computing} shows a scheme of different areas of \ac{SC}. \ac{SC} is an essential part of AI and many of its methods can be classified in the field of Knowledge called ``Natural Computing''. The algorithms that can be found in this category are inspired by the way Nature solves complex problems. In this regard, EC is inspired in the theory of evolution or ANNs find their behaviour in human brain. Because of the variety of techniques used, the structure of this section has been chosen to properly cover the areas included in AI that come in handy in this work.

\subsection{Neural Computation-based Approaches}
NC is the part of \ac{SC} that includes algorithms inspired on how the human brain learns. They have been mainly used in classification and regression problems. In the next points four of the most used NC approaches will be described: Feed-forward NNs (MLPs and ELMs), GPs for Regression and SVR algorithms.

\subsubsection{Multi-layer perceptrons}
\begin{figure}[ht]
  \centering
    \includegraphics[width=0.65\textwidth]{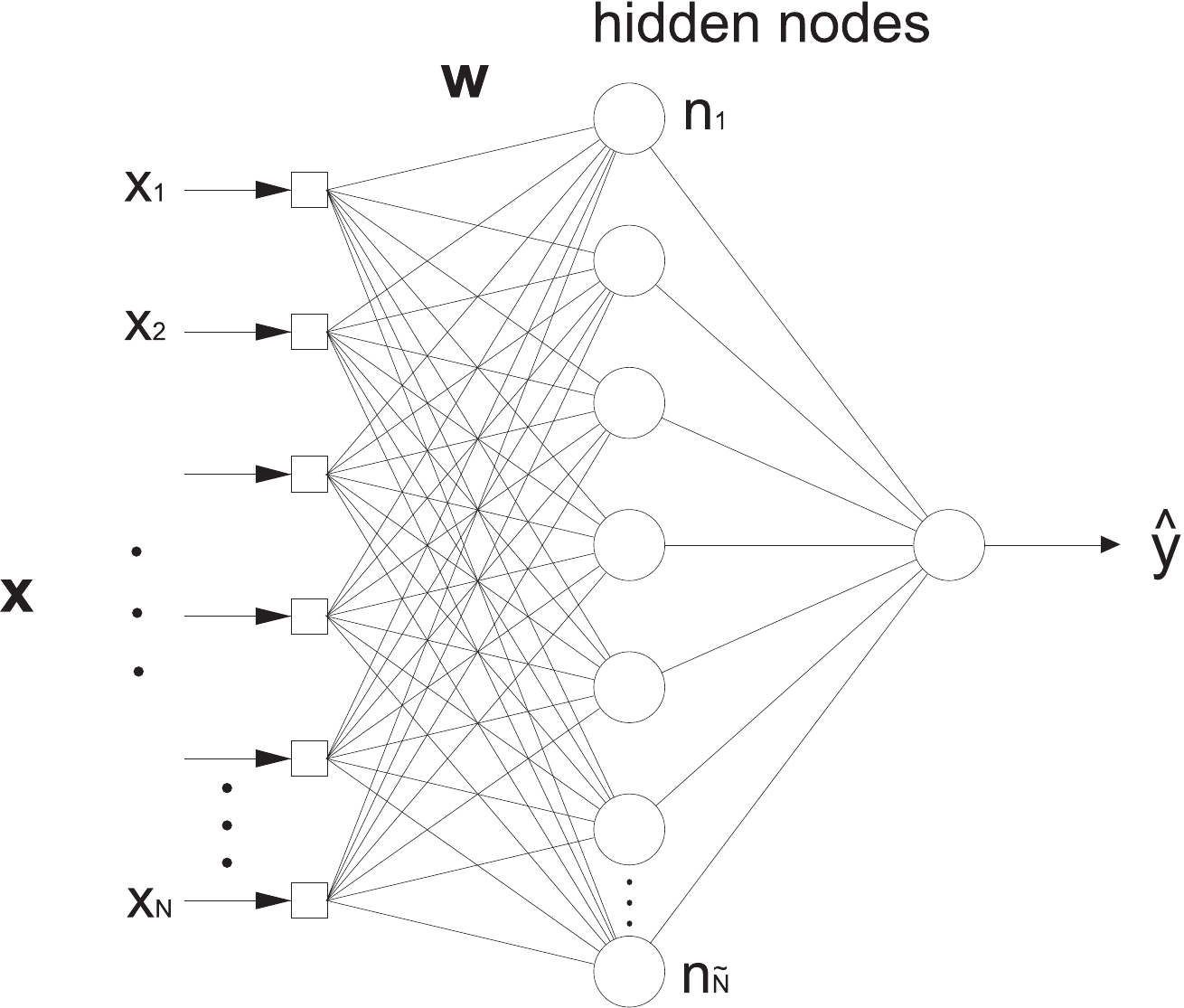}
  \caption{Artificial neural network.}
  \label{red}
\end{figure}

A MLP is a particular kind of \ac{ANN} that is massively parallel. It is considered a distributed information-processing system, which has been successfully applied in modelling a large variety of nonlinear problems \cite{Haykin98,Bishop95}. The MLP consists of an input layer, a number of hidden layers, and an output layer, all of which are basically composed of a number of special processing units called neurons, as Figure \ref{red} shows. Just as important as the processing units themselves is their connectivity, whereby the neurons within a given layer are connected to those of other layers by means of weighted links, whose values are related to the ability of the MLP to learn and generalize from a sufficiently long number of examples. Such a learning process demands a proper database containing a variety of input examples or patterns with the corresponding known outputs. The adequate values of the weights minimize the error between the output generated by the MLP (when fed with input patterns in the database), and the corresponding expected output in the database. The number of neurons in the hidden layer is a parameter to be optimized when using this type of neural network \cite{Haykin98,Bishop95}.

The input data for the MLP consists of a number of samples arranged as input vectors, {\bf x}=$\{x_1, \ldots, x_N\}$. Once a MLP has been properly trained, validated and tested using an input vector different from those contained in the database, it is able to generate a proper output $y$. The relationship between the output and the input signals of a neuron is

\begin{equation}
y=\varphi\left(\sum_{j=1}^n w_j x_j- \theta\right),
\end{equation}
where $y$ is the output signal, $x_j$ for $j=1,\ldots,n$ are the input signals, $w_j$ is the weight associated with the $j$-th input, and $\theta$  is a threshold \cite{Haykin98,Bishop95}. The transfer function $\varphi$ is usually considered as the logistic function,

\begin{equation}
\varphi(x)=\frac{1}{1+e^{-x}}.
\end{equation}

The process to obtain an accuracy output is related with the training procedure as it was mentioned before. During the training process, the error between the estimated output and its corresponding real value in the database will determine what degree the weights in the network should be adjusted, and thanks to all neurons in the network are interconnected (feed-forward NN) MLP makes easy to get this purpose. Hence, The objective of training is to find the combination of weights which result in the smallest error. There are many algorithms that can be used to train a MLP. One possible technique is the backpropagation training algorithm which uses the procedure known as gradient descent to try to locate the global minimum of the error \cite{Gardner}. Another approach is the well-known Levenberg-Marquardt algorithm which is often applied to train the MLP \cite{Hagan94}.

\subsubsection{Extreme Learning Machine}\label{sec:elm}
An ELM \cite{huang2015trends,Huang06} is a novel and fast learning method based on the structure of MLPs, similar to the one shown in Figure \ref{red}. In addition, the ELM approach presents a novel way of training feed-forward NN. The most significant characteristic of the ELM training is that it is carried out just by randomly setting the network weights, and then obtaining a pseudo-inverse of the hidden-layer output matrix. The advantages of this technique are its simplicity, which makes the training algorithm extremely fast, and also its outstanding performance when compared to avant-garde learning methods, usually better than other established approaches such as classical MLPs or SVRs.

Moreover, the universal approximation capability of the ELM network, as well as its classification capability, have been already proven \cite{Huang12}.

The ELM algorithm can be summarized as follows: given a training set $\mathbb{T} = {(\mathbf{x}_i,\boldsymbol{y}_i) | \mathbf{x}_i \in \mathbb{R}^n, \boldsymbol{y}_i \in \mathbb{R}, i=1, \cdots, l},$ an activation function $g(x)$, which a sigmoidal function is usually used, and number of hidden nodes ($\tilde{N}$),

\begin{enumerate}
\item Randomly assign inputs weights $\mathbf{w}_i$ and bias $b_i$, $i = 1, \cdots ,\tilde{N}$.
\item Calculate the hidden layer output matrix $\mathbf{H}$, defined as

\begin{equation}
\mathbf{H} = \left [ \begin{array}{ccc}
g( \mathbf{w}_1 \mathbf{x}_1 + b_1) & \cdots & g(\mathbf{w}_{\tilde{N}} \mathbf{x}_1 + b_{\tilde{N}}) \\
\vdots & \cdots & \vdots \\
g(\mathbf{w}_1 \mathbf{x}_l + b_1) & \cdots & g(\mathbf{w}_{\tilde{N}} \mathbf{x}_N + b_{\tilde{N}})
\end{array}
\right ]_{l \times \tilde{N}}
\end{equation}

\item Calculate the output weight vector $\beta$ as
\begin{equation}
\beta = \mathbf{H}^\dagger \mathbf{y_t},
\end{equation}
where $\mathbf{H}^\dagger$ stands for the Moore-Penrose inverse of matrix $\mathbf{H}$ \cite{Huang06}, and $\mathbf{y_t}$ is the training output vector, $\mathbf{y_t}=[\mathbf{y_t}_1,\cdots,\mathbf{y_t}_l]^\mathbb{T}$.
\end{enumerate}

Note that the number of hidden nodes ($\tilde{N}$) is a free parameter of the ELM training, and must be estimated for obtaining good results. Usually, scanning a range of $\tilde{N}$ values is the best solution.

\subsubsection{Gaussian Processes for Regression}
GPs for regression is a generic supervised-learning method primarily designed for solving regression problems, the advantages of which include the predictive interpolation of the observations. Here, the prediction is probabilistic (Gaussian), so that one computes the empirical confidence intervals and exceeded probabilities to be used in refitting the prediction in some region of interest. Moreover, different linear-regression and correlation models may be specified. Here a short description of the most important characteristics of the GP approach is given, for which the interested reader is referred to the more exhaustive reviews of \cite{L\'azaro12} and \cite{Rasmussen06} for further information.

Given a set of $N$-dimensional inputs $ \mathbf{x}_n$ and their corresponding scalar outputs $y_n$, for the dataset $\mathcal{D}_S \equiv \{ \mathbf{x}_n, y_n\}_{n=1}^l$, the regression task obtains the predictive distribution for the corresponding observation $y_*$ based on $\mathcal{D}_S$, given a new input $\mathbf{x}_*$.

The GP model assumes that the observations can be modelled as some noiseless latent function of the inputs in addition to an independent noise, $y = f(\mathbf{x}) + \varepsilon$, and then specifies a zero-mean GP for both the latent function $f(\mathbf{x})$ $\sim \mathcal{GP}$ $(0, k(\mathbf{x}, \mathbf{x}'))$ and the noise $\varepsilon  \sim \mathcal{N} (0, \sigma^2)$, where $k(\mathbf{x},\mathbf{x}')$ is a covariance function, and $\sigma^2$ is a hyper-parameter that specifies the noise power.

The covariance function $k(\mathbf{x}, \mathbf{x}')$ specifies the degree of coupling between $y(\mathbf{x})$ and $y(\mathbf{x}')$, and encodes the properties of the GP, such as the power level and smoothness. One of the best-known covariance functions is the anisotropic-squared exponential, which has the form of an unnormalized Gaussian function, $k(\mathbf{x}, \mathbf{x}') = \sigma_0^2 \exp \left ( -\frac{1}{2} \mathbf{x}^T \boldsymbol{\Lambda}^{-1} \mathbf{x} \right )$, and depends on the signal power $\sigma_o^2$ and length scales $\boldsymbol{\Lambda}$, where $\boldsymbol{\Lambda}$ is a diagonal matrix containing one length scale per input dimension. Each length scale controls the degree to which the correlation between outputs decay as the separation along the corresponding input dimension grows. All kernel parameters are collectively referred as $\boldsymbol{\theta}$.

The joint distribution of the available observations (collected in $\boldsymbol{y}$) and some unknown outputs $y(\mathbf{x}_*)$ form a multivariate Gaussian distribution, with parameters specified by the covariance function

\begin{equation}
\label{eq:covfun}
\left [ \begin{array}{c} \boldsymbol{y} \\ y_* \end{array} \right ] \sim \mathcal{N} \left ( 0, \left [ \begin{array}{cc} \mathbf{K} + \sigma^2 \mathbf{I}_N & \mathbf{k}_* \\ \mathbf{k}_*^T & k_{**} + \sigma^2 \end{array} \right ] \right ),
\end{equation}
where $[\mathbf{K}]_{nn'} = k(\mathbf{x}_n, \mathbf{x}_{n'})$, $[\mathbf{k}_*]_n = k(\mathbf{x}_n, \mathbf{x}_?)$ and $k_{**} = k(\mathbf{x}_*, \mathbf{x}_*)$. Here, $\mathbf{I}_N$ is used to denote the identity matrix of size $N$. The notation $[\mathbf{A}]_{nn'}$ refers to the entry at row $n$, column $n'$ of $\mathbf{A}$. Likewise, $[\mathbf{a}]_n$ is used to reference the $n$-th element of vector $\mathbf{a}$.

From (\ref{eq:covfun}) and the conditioning on the observed training outputs, the predictive distribution is obtained as

\begin{equation}
\begin{array}{l}
p_{GP}(y_* | \mathbf{x}_*, \mathcal{D}) = \mathcal{N}(y_* | \mu_{GP*}, \sigma_{GP*}^2) \\
\mu_{GP*} = \mathbf{k}_*^T (\mathbf{K} + \sigma^2 \mathbf{I}_N)^{-1} \boldsymbol{y} \\
\sigma_{GP*}^2 = \sigma^2 + k_{**} - \mathbf{k}_*^T (\mathbf{K} + \sigma^2 \mathbf{I}_N )^{-1} \mathbf{k}_* ,
\end{array}
\end{equation}
which is computed $\mathcal{O}(N^3)$ times, due to the inversion of the $N \times N$ matrix $\mathbf{K} + \sigma^2 \mathbf{I}_N$.

The hyper-parameters $\{\boldsymbol{\theta}, \sigma \}$ are typically selected by maximizing the marginal likelihood (also called ``evidence'') of the observations, which is

\[
\log p(\boldsymbol{y} | \boldsymbol{\theta}, \sigma) = - \frac{1}{2} \mathbf{y}^T (\mathbf{K} + \sigma^2 \mathbf{I}_N)^{-1} \boldsymbol{y} -
\]

\begin{equation}
\label{eq:evidence}
 - \frac{1}{2} |\mathbf{K} + \sigma^2 \mathbf{I}_N | - \frac{N}{2} \log(2 \pi).
\end{equation}

If analytical derivatives of (\ref{eq:evidence}) are available, optimization is carried out using gradient methods, with each gradient computed $\mathcal{O}(N^3)$ times. GPs regression algorithms can typically handle a few thousand data points on a desktop computer.

\subsubsection{Support Vector Regression}
SVR \cite{tutorial98} is one of the state-of-the-art algorithms for regression and function approximation. The SVR approach takes into account the error approximation to the data and also the generalization of the model, i.e. its capability to improve the prediction of the model when a new dataset is evaluated by it. Although there are several versions of the SVR, the classical model, $\epsilon$-SVR, described in detail in \cite{tutorial98} and used in a large number of application in Science and Engineering \cite{Salcedo14}, is considered in this work.

The $\epsilon$-SVR method for regression consists of, given a set of training vectors  $\mathbb{T}=\{({\bf x}_i,\boldsymbol{y}_i)),i=1, \ldots, l\}$, where ${\bf x}_i$ stands for a vector of predictive variables, and $\boldsymbol{y}_i$ is the target associated to the input, training a model of the form

\begin{equation}
\hat{\boldsymbol{y}}({\bf x})=g({\bf x})+b = {\bf w}^T\phi({\bf x}) + b,
\end{equation}
where $\hat{\boldsymbol{y}}({\bf x})$ stands for an estimation of $\boldsymbol{y}$, in such a way that a risk function is minimized. This risk function can be written as:

\begin{equation}
\label{eq_minfun}
	R[g] = \frac{1}{2} \left\| {\bf w} \right\|^2 + C \sum_{i=1}^l L\left(\boldsymbol{y}_i,g({\bf x}_i)\right)
\end{equation}
where the norm of ${\bf w}$ controls the smoothness of the model, $\phi({\bf x})$ is a function of projection of the input space to the feature space, $b$ is a parameter of bias, and $L\left(\boldsymbol{y}_i,g({\bf x}_i)\right)$ is the loss function selected. In this thesis, the L1-SVRr is used (L1 support vector regression), characterized by an $\epsilon$-insensitive loss function \cite{tutorial98}:

\begin{equation}
L\left(\boldsymbol{y}_i,g({\bf x}_i)\right)=\left\{
\begin{array}{l l}
0 & \mbox{if}~~|\boldsymbol{y}_i-g({\bf x}_i)| \leq \epsilon\\
|\boldsymbol{y}_i-g({\bf x}_i)|-\epsilon & \mbox{otherwise}\\
\end{array}
\right.
\end{equation}\\

In order to train this model, it is necessary to solve the following optimization problem \cite{tutorial98}:

\begin{equation}
\label{eq_minfun2}
	\min \left(\frac{1}{2} \left\| {\bf w} \right\|^2 + C \sum_{i=1}^l (\xi_i+\xi_i^*) \right)
\end{equation}

\noindent
subject to

\begin{equation}
\label{eq_restriccion1}
	\boldsymbol{y}_i - {\bf w}^T\phi({\bf x}_i) - b \leq \epsilon + \xi_i, \quad i=1,\ldots,l\\
\end{equation}
\begin{equation}
\label{eq_restriccion2}
	-\boldsymbol{y}_i + {\bf w}^T\phi({\bf x}_i) + b  \leq \epsilon + \xi_i^*, \quad i=1,\ldots,l\\
\end{equation}
\begin{equation}
	\xi_i,\xi_i^* \geq 0, \quad i=1,\ldots,l\\
\end{equation}

Figure~\ref{fig:SVR_epsilon} shows and example of the final solution for a given input variables. The dual form of this optimization problem is usually obtained through the minimization of the Lagrange function, constructed from the objective function and the problem constraints. In this case, the dual form of the optimization problem is the following:

\begin{figure}[ht]	
  \begin{center}
    \includegraphics[width=0.85\textwidth]{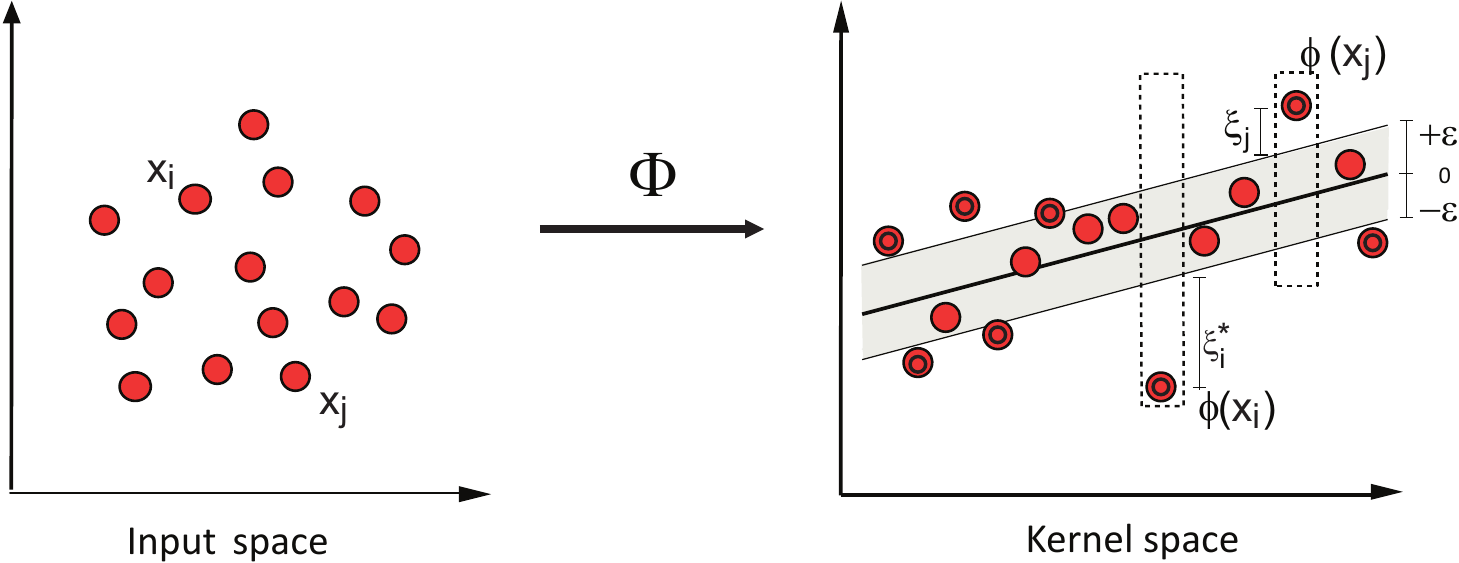}
  \end{center}
  \caption{Illustration of the SVR model. Samples in the original input space are first mapped to a Reproducing Kernel Hilbert Space, where a linear regression is performed. All samples outside a fixed tube of size $\epsilon$ are penalized, and are support vectors (double-circled symbols).}
  \label{fig:SVR_epsilon}
\end{figure}

\[
	\max \left(-\frac{1}{2} \sum_{i,j=1}^l (\alpha_i-\alpha_i^*)(\alpha_j-\alpha_j^*)K({\bf x}_i,{\bf x}_j)- \right.
\]
\begin{equation}
\left.- \epsilon \sum_{i=1}^l (\alpha_i+\alpha_i^*) + \sum_{i=1}^l \boldsymbol{y}_i(\alpha_i-\alpha_i^*)\right)
\end{equation}

\noindent
subject to

\begin{equation}
\label{eq:constraint1}
	\sum_{i=1}^l (\alpha_i-\alpha_i^*) = 0\\
\end{equation}
\begin{equation}
\label{eq:constC}
  \alpha_i,\alpha_i^* \in \left[0,C\right] \\
\end{equation}

In addition to these constraints, the Karush-Kuhn-Tucker conditions must be fulfilled, and also the bias variable, $b$,
must be obtained. The interested reader can consult \cite{tutorial98} for reference. In the dual formulation of the problem the function $K({\bf x}_i,{\bf x}_j)$ is the kernel matrix, which is formed by the evaluation of a kernel function, equivalent to the dot product $\left< \phi({\bf x}_i),\phi({\bf x}_j) \right>$. A usual election for this kernel function is a Gaussian function, as follows:

\begin{equation}
K({\bf x}_i,{\bf x}_j)=\exp(-\gamma \cdot \left\|{\bf x}_i-{\bf x}_j\right\|^2).
\end{equation}

The final form of function $g({\bf x})$ depends on the Lagrange multipliers $\alpha_i,\alpha_i^*$, as follows:

\begin{equation}
\label{eq:model}
	g({\bf x}) = \sum_{i=1}^l (\alpha_i-\alpha_i^*) K({\bf x}_i,{\bf x}) \mbox{.}
\end{equation}

\noindent
So finally, the estimation of the target under study will be carried out using the following expression:

\begin{equation}
\hat{\boldsymbol{y}}({\bf x})=g({\bf x})+b=\sum_{i=1}^l (\alpha_i-\alpha_i^*) K({\bf x}_i,{\bf x})+b \mbox{.}
\end{equation}

\noindent
In this way it is possible to obtain a SVR model by means of the training of a quadratic problem for a given hyper-parameters $C$, $\epsilon$ and $\gamma$. The estimation of these SVR hyper-parameters is a process usually carried out before the training of the algorithm. There are different methods to obtain $C$, $\epsilon$ and $\gamma$, but the most common approach consists of a \ac{GS} procedure. GS exhaustively considers all parameters combinations from a grid of possible pre-defined values. The quality of the SVR with these hyper-parameters' values is tested on a reduced validation set of data from the problem at hand. More information about GS and alternative techniques for SVR hyper-parameters estimation can be found in \cite{tutorial98}. A variant of the GS approach that includes lower and upper bounds to limit the tested values of the hyper-parameters can be used. This method was proposed in \cite{Ortiz2009}, and it is able to considerably reduce the time for hyper-parameters estimation with GS, without affecting the quality of the SVR.

\subsection{Evolutionary Computation-based Algorithms}
EC algorithms are used for solving continuous optimization challenges, working in discrete and search spaces. They are used also in features selection to improve the performance of the predictions in regression problems. All genetic and EAs are based on the evolution of the population of candidate solutions by applying a series of evolutionary operators. Part of this PhD. Thesis is based on the application of this kind of techniques, hence the explanation of different approaches will be carried out in the following points.

\subsubsection{The Grouping Genetic Algorithm}\label{sec:GGA}
There are many potential benefits for applying \ac{FS} in prediction problems: facilitating data visualization and data understanding, reducing the measurement and storage requirements, reducing training and utilization times or defying the curse of dimensionality to improve prediction performance, to mention some of them. Previous literature on this issue mainly focus on constructing and selecting subsets of features that are useful to build a good predictor. This process has been tackled before with EC \cite{Salcedo02,Salcedo14a}. The \ac{GGA} is a class of evolutionary algorithm especially modified to tackle grouping problems, i.e., problems in which a number of items must be assigned to a set of predefined groups (subsets of features, in the case of this work). It was first proposed by Falkenauer \cite{Falkenauer92,Falkenauer98}, who realized that traditional GAs had difficulties when they were applied to grouping problems (mainly, the standard binary encoding increases the space search size in this kind of problem). The \ac{GGA} has shown very good performance on different applications and problems \cite{Agus08,Lit00}. In the GGA, the encoding, crossover and mutation operators of traditional \ac{GA}s are modified to obtain a compact algorithm with very good performance in grouping problems.

\subsubsection{Problem encoding}\label{encoding}
The GGA initially proposed by Falkenauer is a variable-length GA. The encoding is carried out by separating each individual in the algorithm into two parts: the first one is an {\em assignment} part that associates each item to a given group. The second one is a {\em group} part, that defines which groups must be taken into account for the individual. In problems where the number of groups is not previously defined, it is easy to see why this is a variable-length algorithm: the group part varies from one individual to another. In the implementation of the GGA for \ac{FS}, an individual ${\bf c}$ has the form ${\bf c}=[{\bf a}|{\bf g}]$. An example of an individual in the proposed GGA for a FS problem, with 20 features and 4 groups, is the following:

1 1 2 3 1 4 1 4 3 4 4 1 2 4 4 2 3 1 3 2 $|$ 1 2 3 4

where the group 1 includes features $\{1,2,5,7,12,18\}$, group 2 features $\{3,13,16,20\}$, group 3 features \{4,9,17,19\} and finally group 4 includes features $\{6,8,10,11,14,15\}$.

\subsubsection{Genetic operators}
Tournament-based selection mechanism is usually used, similar to the one described in \cite{Xin99}. It has been shown to be one of the most effective selection operators, avoiding super-individuals and performing a excellent exploration of the search space. Regarding the crossover operator, it is implemented in the GGA as a modified version of the initially proposed by Falkenauer \cite{Falkenauer92,Falkenauer98}. The process to apply this operator follows the process outlined in Figure \ref{cruce_GGA}:

\begin{figure}[ht]
\centering
\includegraphics[width=0.65\textwidth]{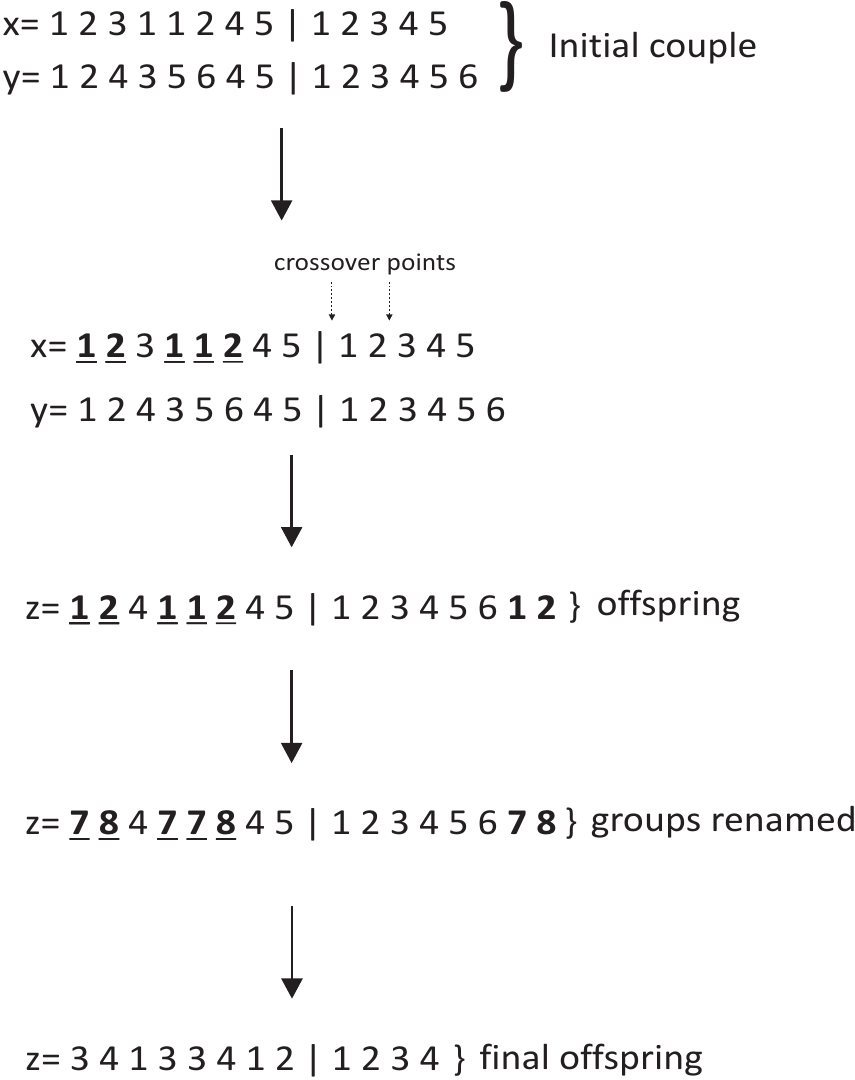} \\
\caption{\label{cruce_GGA} Outline of the grouping crossover implemented in the proposed example of GGA.}
\end{figure}

\begin{itemize}
\item[1.] Choose two parents from the current population, at random.
\item[2.] Randomly select two points for the crossover, from the ``Groups'' part of parent 1, then, all the groups between the two cross-points are selected. In the example of Figure \ref{cruce_GGA} the two crossover points are $G_1$ and $G_2$. Note that, in this case the items of parent1 belonging to group $G_{1}$ and $G_{2}$ are 1, 2, 4, 5, and 6.
\item[3.] Insert the selected section of the ``Groups'' part into the second parent. After the insertion in the example of Figure \ref{cruce_GGA}, the assignment of the nodes 1, 2, 4, 5 and 6 of the offspring individual will be those of parent 1, while the rest of the nodes' assignment are those of parent 2. The ``Groups'' part of the offspring individual is that of parent 2 plus the selected section of parent 1 (8 groups in total, in this case).
\item[4.] Modify the ``Groups'' part of the offspring individual with their corresponding number. In the example, $G$ = 1 \hspace{1mm} 2 \hspace{1mm} 3 \hspace{1mm} 4 \hspace{1mm} 5 \hspace{1mm} 6 \hspace{1mm} 1 \hspace{1mm} 2 is modified into $G$ = 1 \hspace{1mm} 2 \hspace{1mm} 3 \hspace{1mm} 4 \hspace{1mm} 5 \hspace{1mm} 6 \hspace{1mm} 7 \hspace{1mm} 8. Modify also the assignment part accordingly.
\item[5.] Remove any empty groups in the offspring individual. In the example considered, it is found that groups 1, 2, 3, and 6 are empty, these groups' identification number are eliminated and the rest are rearranged. The final offspring is then obtained.
\end{itemize}

Regarding mutation operator, note that standard mutation usually calls for an alteration of a small percentage of randomly selected parts of the individuals. This type of mutation may be too disruptive in the case of a grouping problem. In this case, a swapping mutation in which two items are interchanged (swapping this way the assignment of features to different groups), is taken into account. This procedure is carried out with a very low probability to avoid increasing of the random search in the process.

\subsubsection{The Coral Reef Optimization}
The CRO is a novel meta-heuristic approach for optimization, recently proposed in \cite{Sancho_CRO}, which is based on simulating the corals' reproduction and coral reefs' formation processes. Basically, the CRO is based on the artificial modeling of a coral reef $\mathcal{R}$, consisting of a $N \times M$ grid. It is assumed that each square \emph{(i,j)} of $\mathcal{R}$ is able to allocate a coral $\Omega_{i,j}$. Note that each of such corals represents a solution to a given optimization problem, for which it is encoded as a string of numbers, spanning a given alphabet $\mathcal{A}$. The CRO algorithm is first initialized at random by assigning some squares in $\mathcal{R}$ to be occupied by corals (i.e. solutions to the problem) and some other squares in the grid to be empty, i.e. holes in the reef where new corals can freely settle and grow in the future. The rate between free/occupied squares in $\mathcal{R}$ at the beginning of the algorithm is denoted as $\rho\in\mathbb{R}(0,1)$ and referred to as initial occupation factor. Each coral is labeled with an associated \emph{health} function $f(\Omega_{ij}):\mathcal{A} \rightarrow \mathbb{R}$ that corresponds to the problem's objective function. The CRO is based on the fact that the reef will evolve and develop as long as healthier or stronger corals (which represent better solutions to the problem at hand) survive, while less healthy corals perish.

After the reef initialization described above, the phase of reef formation is artificially simulated. This phase consists of \emph{K} iterations: at each of such iterations the corals' reproduction in the reef is emulated by applying different operators and processes as described in Algorithm \ref{alg:CRO}: a modeling of corals' sexual reproduction (broadcast spawning and brooding).

After the reproduction stage, the set of formed larvae (namely, newly produced solutions to the problem) attempts to find a place on the reef to develop and further reproduce. This deployment may occur in a free space inside the reef (hole), or in an occupied location, by fighting against the coral currently settled in that place. If larvae are not successful in locating a place to settle after a number of attempts, they are considered as preyed by animals in the reef. The coral builds a new reef layer in every iteration.

\begin{algorithm}[!h]
   \caption{Pseudo-code for the CRO algorithm}
   \label{alg:CRO}
   \begin{algorithmic}[1]
   \REQUIRE Valid values for the parameters controlling the CRO algorithm
   \ENSURE A single feasible individual with optimal value of its \emph{fitness}
   \STATE Initialize the algorithm
   \FOR{each iteration of the simulation}
   \STATE Update values of influential variables: predation probability, etc.
   \STATE Sexual reproduction processes (broadcast spawning and brooding)
   \STATE Settlement of new corals
   \STATE Predation process
   \STATE Evaluate the new population in the coral reef
   \ENDFOR
   \STATE Return the best individual (final solution) from the reef
   \end{algorithmic}
\end{algorithm}

The specific definition of the different operators that form the classical CRO algorithm is detailed here:

\begin{enumerate}

\item {\bf Sexual reproduction}: The CRO model implements two different kinds of sexual reproduction: external and internal.
\begin{enumerate}
\item {\bf External sexual reproduction} or \emph{broadcast spawning}: the corals eject their gametes to the water, from which male-female couples meet and combine together to produce a new larva by sexual crossover. In Nature, some species are able to combine their gametes to generate mixed polyps even though they are different from each other. In the CRO algorithm, external sexual reproduction is applied to a usually high fraction $F_b$ of the corals. The couple selection can be done uniformly at random or by resorting to any fitness proportionate selection approach (e.g. roulette wheel). In the original version of the CRO, standard crossover (one point or two-points) are applied in the broadcast spawning process.

\item {\bf Internal sexual reproduction} or \emph{brooding}: CRO applies this method to a fraction $(1-F_b)$ of the corals in the reef. The brooding process consists of the formation of a coral larva by means of a random mutation of the brooding-reproductive coral (self-fertilization considering hermaphrodite corals). The produced larvae is then released out to the water in a similar fashion than that of the larvae generated through broadcast spawning.
\end{enumerate}

\item {\bf Larvae settlement}: once all larvae are formed at iteration $k$ through reproduction, they try to settle down and grow in the reef. Each larva will randomly attempt at setting in a square $(i,j)$ of the reef. If the location is empty (free space in the reef), the coral grows therein no matter the value of its health function. By contrast, if another coral is already occupying the square at hand, the new larva will set only if its health function is better than the fitness of the existing coral. A number of attempts $\mathcal{N}_{att}$ for a larva to set in the reef is defined: after $\mathcal{N}_{att}$ unsuccessful tries, it will not survive to following iteration.

\item {\bf Depredation}: corals may die during the reef formation phase of the reef. At the end of each iteration, a small number of corals can be preyed, thus liberating space in the reef for the next iteration. The depredation operator is applied under a very small probability $P_d$, and exclusively to a fraction $F_d$ of the worse health corals.
\end{enumerate}

\section{Structure of the thesis} \label{structure_of_the_thesis}
The rest of this thesis is organized in two technical parts:

\begin{enumerate}
\item First, proposed contributions with numerical results in renewable energy problems is structured in two chapters: Ocean wave features prediction, and WPREs prediction.

\item The next part, proposed contributions with numerical results in facilities management is divided in two other chapters: Accurate estimation of $H_s$ with SVR and marine radar images, and efficient prediction of low-visibility events at airports.
\end{enumerate}

To conclude, some final remarks and future research lines are summarize in the last part of the document, with the list of publications shown in a final Appendix section.

\part{Proposed contributions with numerical results in renewable energy problems}\label{part:aplicaciones1}
\chapter{Ocean wave features prediction}\label{cap:ocean}
\section{Introduction}
The exploitation of marine energy resources is currently a hot topic in renewable energy research, since they have shown a clear potential for sustainable growth \cite{Defne09,Garcia14,Lenee11,Lopez2013,Rusu09,Rusu12,Gonzalves14}: marine energy resources do not generate CO$_2$, are potentially able to convert part of the huge energy of oceans into electricity \cite{Arinaga12,Esteban2012}, and reduce oil imports, a crucial geo-economical issue. However, in spite of this potential, the use of marine energy sources is nowadays still minor at global level. In spite of this, wave energy plays a key role for sustainable development in several offshore islands because it provides not only technical and economical benefits (to satisfy the demands of clean electricity) but also without significant environmental impact, a key concern in offshore islands, committed to the protection of ecological systems \cite{Fadaeenejad2014}. Some interesting reviews of the most important issues involved in the generation of electricity from oceans (including converters, their related economical aspects, and the potential of a number of ocean regions to be exploited worldwide) can be found in \cite{Bahaj2011,Chong2013,Kim2012,Hammar2012,cuadra2016computational}.

There are different technologies that can be considered within marine energy resources, including ocean wave, tidal and ocean thermal. This work is focused on wave energy, that uses \ac{WECs} to convert ocean energy into electricity \cite{Falcao2010,cuadra2016computational}. WECs transform the kinetic energy of wind-generated waves into electricity by means of either the vertical oscillation of waves or the linear motion of waves, and exhibit some important advantages when compared to alternatives based on tidal converters. However, waves are more difficult to characterize than tides, because of their stochastic nature. As a consequence of this complexity, both the design, deployment, and control of WECs \cite{Hong2014,Richter2013} become key topics that require a proper characterization and prediction of waves \cite{Larsen15,Reikard15,Wimmer06}. Maybe, the two most important wave parameters in this regard to characterize wave energy is the $H_s$ and the  \ac{$P$}, in which prediction this chapter is focused on.

As mentioned, waves' stochastic nature makes very difficult the prediction of wave energy resource, so the research work on this topic has been intense in the last few years. Focusing on ML approaches, one of the first algorithms proposed to predict $H_s$ is due to Deo et al. \cite{Deo98}, who use ANN to obtain an accurate prediction of $H_s$. Improvements on this prediction system were presented in a more recent work \cite{Agrawal04}. NN have also been applied to other problems of $H_s$ and $P$ prediction, such as \cite{ChingPiao2002}, where $H_s$ and $P$ are predicted from observed wave records using time series NN, \cite{Castro14}, where a neural network is applied to estimate the wave energy resource in the northern coast of Spain, or \cite{Morteza2009}, where a hybrid GA-adaptive network-based fuzzy inference system model was developed to forecast $H_s$ and the peak spectral period at Lake Michigan. Alternative proposals based on different approaches have been recently proposed like in \cite{Mahjoobi08}, where different SC techniques are tested for $H_s$ prediction, \cite{Mahjoobi09} where a SVR methodology is considered, \cite{Fernandez15} where different classifiers have been applied to analyze and predict $H_s$ and $P$ ranges in buoys for marine energy applications, \cite{Nitsure12}, that propose the use of genetic programming for $H_s$ reconstruction problems or \cite{Ozger11}, where \ac{FL}-based approaches were introduced for $H_s$ prediction problems.

In spite of this huge work dealing with ML algorithms in $H_s$ and $P$ prediction, there are not previous studies focussed on analyzing what are the best predictive variables to obtain an accurate prediction of these important parameters from neighbour buoys data. This problem is usually known in the ML community as \ac{FS} \cite{Weston00}, and it is an important task in supervised classification and regression problems. The reason for this is that irrelevant features, used as part of a training procedure in a classification or regression machine, can unnecessarily increase the cost and running time of a prediction system, as well as degrade its generalization performance \cite{Blum97,Salcedo02}. In this thesis a novel hybrid GGA--ELM for accurate prediction of $H_s$ and $P$ values is proposed. The GGA is a recently proposed algorithm especially modified to tackle grouping problems \cite{Falkenauer92,Falkenauer98}. In this case it is focussed on obtaining the best set of features (predictive variables) for the regressor machine (ELM). It will be shown how the GGA is able to produce different sets of good predictive variables for this problem, and how the ELM is able to obtain excellent $H_s$ or $P$ prediction from them. An experimental analysis of the proposed hybrid GGA-ELM approach in a real case of $H_s$ and $P$ prediction in buoys at the Western coast of the USA will be carried out. The application of alternative regression techniques such as SVR have been also analyzed in these experiments. Moreover, because of this hybrid prediction system has a number of parameters that may affect its final performance and they need to be previously specified by the practitioner, an automatic fine tuning of the prediction system's parameters is added to the study. In this case, the parameters of GGA-ELM approach include the probability of mutation in the GGA or the number of neurons in the ELM hidden layer, among others. It is proposed then to use a \ac{BO} approach to automatically optimize the parameters of the whole prediction system (GGA-ELM), with the aim of improving its performance in wave energy prediction problems. BO has been shown to obtain good results in the task of obtaining good parameter values for prediction systems \cite{Snoek12}.

The rest of the chapter is structured in the following parts: Section \ref{cal_P_H} where the calculation of $H_s$ and $P$ is done, Section \ref{sec:hyb_sys} that presents the prediction system considered, Section \ref{bo_section} which addresses the explanation of the BO method, Section \ref{exper_bo} that summarizes the experiments and results obtained and Section \ref{Conclusions1} that completes the study with some final remarks.

\section{Wave energy resource: calculation of $P$ and $H_s$}\label{cal_P_H}
In the evaluation of wave energy deployment systems such as WECs or WECs arrays, it is essential to previously characterize as accurately as possible the amount of wave energy available in a particular location, given by parameters such as $H_s$ and $P$. In order to obtain these parameters, note that the wave energy resource in a region is caused by both local and far winds blowing over the ocean surface, which transports the wave energy. Focusing thus the attention on the water surface, and within the framework of the linear wave theory, the vertical wave elevation, $\eta(\mathbf{r},t)$, at a point $\mathbf{r}=(x,y)$ on the sea surface at time $t$ can be assumed as a superposition of different monochromatic wave components \cite{Nieto13,Goda10}. This model is appropriate when the free wave components do not vary appreciably in space and time (that is, statistical temporal stationarity and spatial homogeneity can be assumed \cite{Goda10}).

In this model, the concept of ``sea state'' refers to the sea area and the time interval in which the statistical and spectral characteristics of the wave do not change considerably (statistical temporal stationarity and spatial homogeneity). The total energy of a sea state is the combined contribution of all energies from different sources. The ``wind sea'' occurs when the waves are caused by the energy transferred between the local wind and the free surface of the sea. The ``swell'' is the situation in which the waves have been generated by winds blowing on another far area (for instance, by storms), and propagate towards the region of observation. Usually, sea states are the composition of these two pure states, forming multi-modal or mixed seas.

In a given sea state, the wave elevation $\eta(\mathbf{r},t)$ with respect to the mean ocean level can be assumed as a zero-mean Gaussian \emph{stochastic process}, with statistical symmetry between wave maxima and minima. A buoy deployed at point $\mathbf{r}_B$ can take samples of this process, $\eta(\mathbf{r}_B,t_j)$ $j=1,2, \cdots , t_{\mathrm{MAX}}$, generating thus a time series of empirical vertical wave elevations. The \ac{DFT} of this sequence, using the \ac{FFT} algorithm, allows for estimating the \emph{spectral density} $S(f)$. Its spectral moments of order $n$ can be computed as follows:

\begin{equation}\label{equation_Spectral_Moments}
 m_n = \int_{0} ^{\infty} f^n S(f) df,
\end{equation}

The $P$ is a first indicator of the amount of wave energy available in a given area. $P$, or power density per meter of wave crest \cite{Cahill13} can be computed as

\begin{equation} \label{eq_P_Hs_Te}
 P =  \frac{\rho g^2}{4 \pi}\int_{0}^{\infty}\frac{S(f)}{f}  df  =   \frac{\rho g^2}{4 \pi}   m_{-1} =   \frac{\rho g^2}{64 \pi}   H_s^2  \cdot T_e
\end{equation}

where $\rho$ is the sea water density (1025 kg/m$^3$), $g$ the acceleration due to gravity, $H_s = 4  \sqrt{m_0}$ is the spectral estimation of the $H_s$, and $T_e \equiv T_{-1,0} = m_{-1} / m_0$ is an estimation of the mean wave period, normally known as the period of energy, which is used in the design of turbines for wave energy conversion. Expression (\ref{eq_P_Hs_Te}) (with $H_s$ in meters and $T_e$ in seconds) leads to $P =  0.49\cdot   H_s^2 \cdot T_e$ kW/m, and helps engineers estimate the amount of wave energy available when planning the deployment of WECs at a given location.

\section{The hybrid prediction system considered}\label{sec:hyb_sys}
The prediction system is a hybrid {\em wrapper} approach, formed by the GGA (explained in depth in Section \ref{sec:GGA}) for FS and the ELM to carry out the final prediction of $H_s$ or $P$ from a set of input data. The regressor chosen must be as accurate as possible, and also very fast in its training process, in order to avoid high computational burden for the complete algorithm. This is the main reason why the ELM is selected for the fitness function as well, and whose explanation is carried out in detail in Section \ref{sec:elm}. Since the ELM is hybridized with the GGA, there are different ways of calculating the final fitness associated with each individual. In this case the following fitness function is considered, that uses a measure of the \ac{RMSE} of the prediction for the best group of features in the GGA:

\begin{equation}
f({\bf x})=\sqrt{\frac{1}{N} \sum_{k=1}^N \left(Y(k)-\hat{Y}^b(k)\right)^2}
\end{equation}

where $Y$ stands for the $H_s$ or $P$ measured for sample $k$, and $\hat{Y}^b(k)$ stands for the $H_s$ or $P$ estimated by the ELM in the group of the individual with less error (best group of features), for sample $k$. Note that $N$ stands for the number of training samples.

\section{Bayesian optimization of the prediction system}\label{bo_section}
Every ML algorithm or prediction system has its own set of parameters that must be adjusted to obtain an optimal performance. An example is a deep neural network in which one has to specify parameters such as the learning rate, the number of layers, the number of neurons in
each layer, etc.  \cite{Bengio15}. Another example is stochastic gradient boosting in which one has to choose the number of terminal nodes in the ensemble trees, the number of trees, the regularization parameter, etc. \cite{Friedman02}. In the particular setting in this study, in an ELM the
number of units in the hidden layer has to be specified before training; and in the GA described in Section \ref{sec:GGA}, the probability of mutation and the number of epochs must be known initially.

Changing the parameter values of a prediction system may have a strong impact in its performance. Parameter tuning is hence defined as the problem of
finding the optimal parameter values of a prediction system on the problem considered. This task has traditionally been addressed by human experts, which often use prior knowledge to specify parameter values that are expected to perform well. However, such an approach can suffer from human bias. An alternative solution is to consider a grid or uniform search in the space of parameters to look for values that result in a good performance on a validation set. These methods, however, suffer when the dimensionality of the parameter space is very high \cite{Bergstra12}, requiring a large number of parameter evaluations.

BO has emerged as practical tool for parameter selection in prediction systems. These methods provide an efficient alternative to a grid or uniform search of the parameter space \cite{Snoek12}. Assume that the surface defined by the error of a prediction system that depends on some parameters is smooth. In that case, a search through the parameter space according to a criterion that exploits this smoothness property and avoids exhaustive exploration can be done. More precisely, BO methods are very useful for optimizing black-box objective functions that lack an analytical expression (which means no gradient information), are very expensive to evaluate, and in which the evaluations are potentially noisy \cite{Mockus78,Brochu10,Shahriari16}. The performance of a prediction system on a randomly chosen validation set, when seen as a function of the chosen parameters, has all these characteristics.

Consider a black-box objective $f(\cdot)$ with noisy evaluations of the form $y_i = f(\mathbf{x}_i)+\epsilon_i$, with $\epsilon_i$ some noise term. BO methods are very successful at reducing the number of evaluations of the objective function  needed to solve the optimization problem. At each iteration $t=1,2,3,\ldots$ of the optimization process, these methods fit a probabilistic model, typically a \ac{GP} to the observations of objective function $\{y_i\}_{i=1}^{t-1}$ collected so far. The uncertainty about the objective function provided by the GP is then used to generate an acquisition function $\alpha(\cdot)$, whose value at each input location indicates the expected utility of evaluating $f(\cdot)$ there. The next point $\mathbf{x}_t$ at which to evaluate the objective $f(\cdot)$ is the one that maximizes $\alpha(\cdot)$. Importantly,  $\alpha(\cdot)$ only depends on the probabilistic model and can hence be evaluated with very little cost. Thus, this function can be maximized very quickly using standard optimization techniques.
This process is repeated until enough data about the objective has been collected. When this is the case, the GP predictive mean for $f(\cdot)$ can be optimized to find the solution of the optimization problem. Algorithm \ref{alg:bo} shows the details of such a process.

\begin{figure}[ht]
\begin{algorithm}[H]
 \caption{BO of a black-box objective function.}
 \label{alg:bo}
 \For{$\text{t}=1,2,3,\ldots,\text{max\_steps}$}{
 	{\bf 1:} Find the next point to evaluate by optimizing the acquisition function:
	$\mathbf{x}_t = \underset{\mathbf{x}}{\text{arg max}} \quad \alpha(\mathbf{x}|\mathcal{D}_{1:t-1})$.
	
	{\bf 2:} Evaluate the black-box objective $f(\cdot)$ at $\mathbf{x}_t$: $y_t=f(\mathbf{x}_\text{t}) + \epsilon_t$.

	{\bf 3:} Augment the observed data $\mathcal{D}_{1:t}=\mathcal{D}_{1:t-1} \bigcup \{\mathbf{x}_t, y_t\}$.

	{\bf 4:} Update the GP model using $\mathcal{D}_{1:t}$.
 }
 \KwResult{ Optimize the mean of the GP to find the solution. }
 \vspace{.5cm}
\end{algorithm}
 \vspace{.5cm}
\end{figure}

The key for BO success is that evaluating the acquisition function $\alpha(\cdot)$ is very cheap compared to the evaluation of the actual objective $f(\cdot)$, which in this case requires re-training the prediction system. This is so because the acquisition function only depends on the GP predictive distribution for $f(\cdot)$ at a candidate point $\mathbf{x}$. Let the observed data until step $t-1$ of the algorithm be $\mathcal{D}_i=\{(\mathbf{x}_i,y_i)\}_{i=1}^{t-1}$. The GP predictive distribution for $f(\cdot)$ is given by a Gaussian distribution characterized by a mean $\mu(\mathbf{x})$ and a variance $\sigma^2(\mathbf{x})$. These values are:

\begin{align}
\mu(\mathbf{x}) & = \boldsymbol{k}_{*}^{T} (\mathbf{K}+\sigma_{n}^{2}I)^{-1}\boldsymbol{y}\,, \\
\sigma^2(\mathbf{x}) & = k(\boldsymbol{x},\boldsymbol{x}) - \boldsymbol{k}_{*}^T(\mathbf{K}+\sigma_{n}^{2}I)^{-1}\boldsymbol{k}_*\,.
\end{align}

where $y=(y_1,\ldots,y_{t-1})$ is a vector with the objective values observed so far; $\mathbf{k}_*$ is a vector with the prior covariances between $f(\mathbf{x})$ and each $y_i$; $\mathbf{K}$ is a matrix with the prior covariances among each $y_i$, for $i=1,\ldots,t-1$; and $k(\boldsymbol{x},\boldsymbol{x})$ is the prior variance at the candidate location $\mathbf{x}$. All these quantities are obtained from a covariance function $k(\cdot,\cdot)$ which is pre-specified and receives as an input two points, $\mathbf{x}_i$ and $\mathbf{x}_j$, at which the covariance between $f(\mathbf{x}_i)$ and $f(\mathbf{x}_j)$ has to be evaluated. A typical covariance function employed for BO is the Mat\'ern function \cite{Snoek12}.
For further details about GPs the reader is referred to \cite{Rasmussen06}.

\begin{figure}[ht]
\centering
\begin{tabular}{c}
\includegraphics[width=0.85\textwidth]{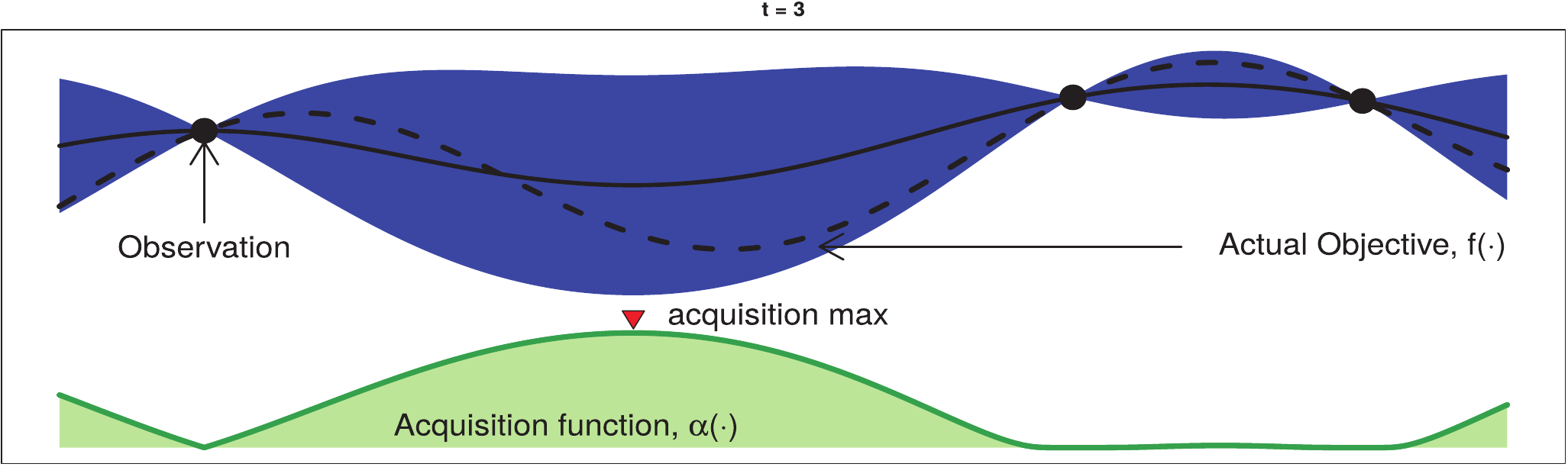} \\
\includegraphics[width=0.85\textwidth]{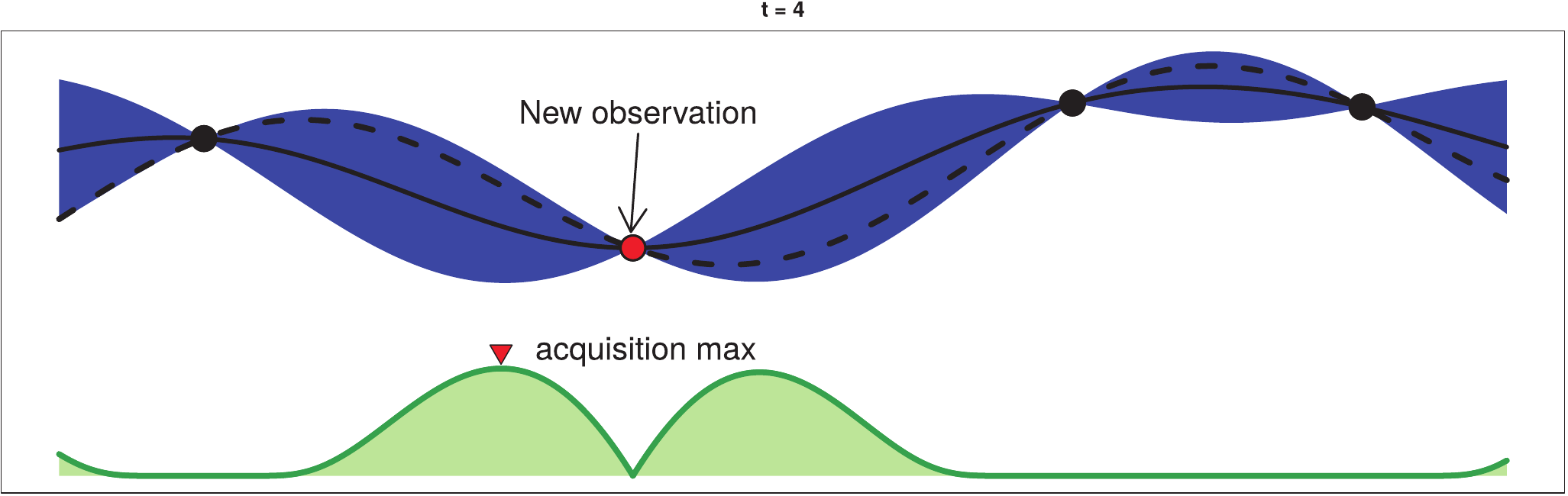} \\
\includegraphics[width=0.85\textwidth]{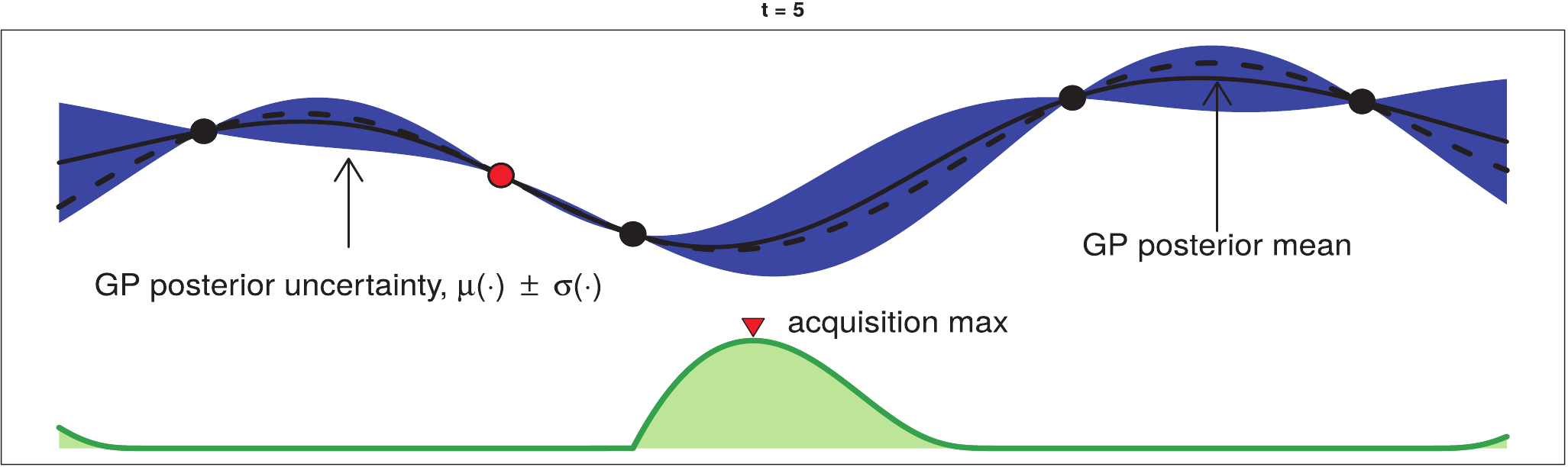}
\end{tabular}
\caption{\label{bo:illustration}
An example of BO on a toy 1D noiseless problem.}
\end{figure}

Thus, BO methods typically look for the best position very carefully to evaluate next the objective function with the aim of finding its optimum with the smallest number of evaluations. This is a very useful strategy when the objective function is very expensive to evaluate and it can save a lot of computational time. Three steps of the BO optimization process are illustrated graphically in Figure \ref{bo:illustration} for a toy minimization problem.

Figure \ref{bo:illustration} shows a GP estimation of the objective $f(\cdot)$ over three iterations. The acquisition function is shown in the lower part of the plot. The acquisition is high where the GP predicts a low objective and where the uncertainty is high. Those regions in which it is unlikely to find the global minimum of $f(\cdot)$ have low acquisition values and will not be explored.

Unlike BO methods, grid or uniform search strategies are based in a pure exploration of the search space. If the assumption that the objective function is smooth is made, doing a few evaluations in regions of the input space that look more promising (exploitation) is expected to give better results.
In BO methods the acquisition function $\alpha(\cdot)$ balances between exploration and exploitation in an automatic way. An example of an acquisition function is \ac{EI} \cite{Jones98}. EI is obtained as the expected value under the GP predictive distribution for $y_i$, of the utility function $u(y_i) = \text{max}{(0, \nu - y_i)}$, where $\nu=\text{min}(\{y_i\}_{i=1}^{t-1})$ is the best value observed so far. That is, EI measures on average how much the current best solution by evaluating the objective at each candidate point will be improved on. An advantage of EI is that the corresponding acquisition function $\alpha(\cdot)$ can be computed analytically: $\alpha(\mathbf{x}) = \sigma(\mathbf{x})(\gamma(\mathbf{x}) \Phi(\gamma(\mathbf{x}) + \phi(\gamma(\mathbf{x}))$, where $\gamma(\mathbf{x}) = (\nu - \mu(\mathbf{x})) /\sigma(\mathbf{x})$ and $\Phi(\cdot)$ and $\phi(\cdot)$ are respectively the c.d.f. and p.d.f. of a standard Gaussian. EI is the acquisition function displayed in Figure \ref{bo:illustration}.

BO has been recently applied with success in different prediction systems for finding good parameter values. For example, it has been used to find the parameters of topic models based on latent Dirichlet allocation, \ac{SVM}, or deep convolutional NN \cite{Snoek12}. Furthermore, BO methods have also been used to optimize a logistic regression model for labelling Amazon product reviews \cite{Dewancker16}, or to optimize the weights of a neural network to balance vertical poles and lengths on a moving cart \cite{Frean08}. Another applications of BO are found in the field of environmental monitoring, in the task of adjusting the parameters of a control system for robotics, in the optimization of recommender
systems, and in combinatorial optimization \cite{Brochu10,Shahriari16}. Finally, BO methods has been implemented in different software packages. An implementation in python is called Spearmint and is available at \cite{github}, which is the BO implementation used in this work.

\section{Experiments and results}\label{exper_bo}
\begin{figure}[ht]
\centering
\includegraphics[width=0.65\textwidth]{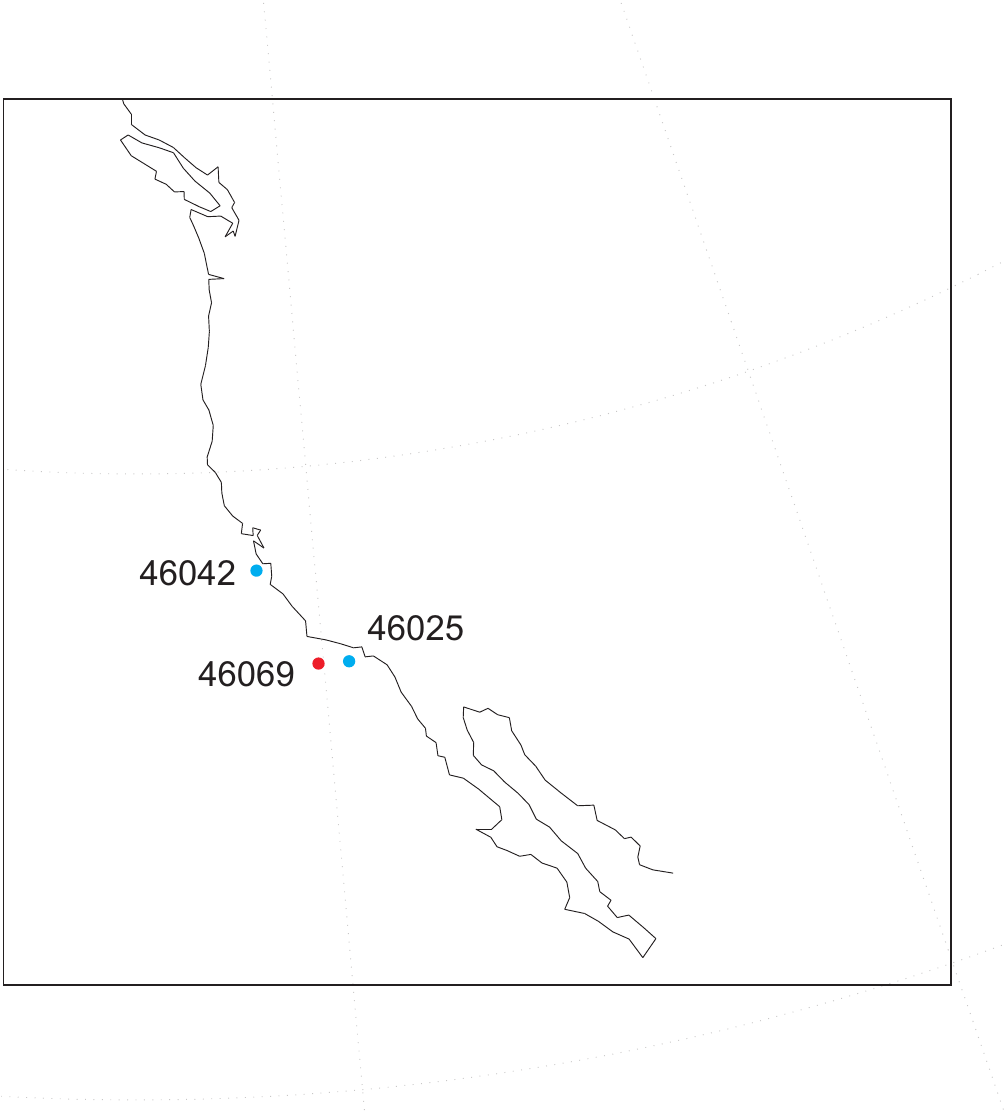} \\
\caption{\label{boyas_con_b} Western USA Buoys considered in this study. In red buoy where the $P$ prediction is carried out from data at blue ones.}
\end{figure}

This section describes some experiments with the aim of showing the improvements obtained in the performance of the prediction system when its parameters are optimized with the Bayesian techniques introduced before. A real problem of $P$ prediction ($P =  0.49\cdot   H_s^2 \cdot T_e$ kW/m, \cite{Goda10}) from marine buoys is considered. Figure \ref{boyas_con_b} shows the three buoys considered in this study at the Western coast of the USA, whose data bases are obtained from \cite{NOAA}, and their main characteristics are shown in Table \ref{Buoy_description}. The objective of the problem is to carry out the reconstruction of buoy 46069 from a number of predictive variables from the other two buoys. Thus, 10 predictive variables measured at each neighbor buoy are considered (a total of 20 predictive variables to carry out the reconstruction). Table \ref{tab:databaseSets} shows details of the predictive variables for this problem. Data for two complete years (1st January 2009 to 31st December 2010) are used, since complete data (without missing values in predictive and objective $P$) are available for that period in the three buoys. These data are divided into training set (year 2009) and test set (year 2010) to evaluate the performance of the proposed algorithm.

\begin{table}[ht]
\scriptsize{}
\begin{center}
\caption{\label{Buoy_description} Geographic coordinates and buoy's description.} \vspace{0.3cm}
\resizebox{14cm}{!} {
\begin{tabular}{cccc}
\hline
Characteristics         & Station 46025                 & Station 46042                 & Station 46069\\
 &33$^{\circ}$44'58"N 119$^{\circ}$3'10"W        &36$^{\circ}$47'29"N 122$^{\circ}$27'6"W        &33$^{\circ}$40'28"N 120$^{\circ}$12'42"W\\
\hline
Site elevation\        & sea level\                    & sea level                       & sea level\  \\
Air temp height\       & 4 m above site elevation\     & 4 m above site elevation        & 4 m above site elevation\    \\
Anemometer height\     & 5 m above site elevation\     & 5 m above site elevation        & 5 m above site elevation\    \\
Barometer elevation\   & sea level\                    & sea level                       & sea level\    \\
Sea temp depth\        & 0.6 m below water line\       & 0.6 m below water line          & 0.6 m below water line\    \\
Water depth\           & 905.3 m\                      & 2098 m                          & 1020.2 m\    \\
Watch circle radius\   & 1327 yards\                   & 2108 yards                      & 1799 yards\    \\
\hline
\end{tabular}}
\end{center}
\end{table}

\begin{table}[ht]
\begin{center}
\caption{Predictive variables used in the experiments.}
\label{tab:databaseSets}
\begin{tabular}{ccc}
       \hline
Acronym	& Predictive		& units \\
&variable&\\
\hline
\hline
WDIR\         & Wind direction\            & [degrees]\\
WSPD\         & Wind speed\                & [m/s]\\
GST\          & Gust speed\                & [m/s]\\
WVHT\         & Significant wave height\   &  [m]\\
DPD\          & Dominant wave period\      & [sec]\\
APD\          & Average period\            & [sec]\\
MWD\          & Direction DPD\             & [degrees]\\
PRES\         & Atmospheric pressure\      & [hPa]\\
ATMP\         & Air temperature\           & [Celsius]\\
WTMP\         & water temperature\         & [Celsius]\\
\hline
\end{tabular}
\end{center}
\end{table}

This experimental section is divided into two different subsections. First, the performance of the BO techniques proposed in the optimization of the specific GGA-ELM prediction algorithm is shown. Second, it will presented how the prediction performance is improved when the system is run with the parameters obtained by the BO techniques, i.e. by comparing the performance of the system before and after tuning the parameters with BO.

\subsection{Methodology}
The utility of the BO techniques described in Section \ref{bo_section}, for finding good parameters for the prediction system described in Section \ref{sec:hyb_sys}, will be evaluated. More precisely, the parameters that minimize the RMSE of the best individual found by the GGA on a validation set that contains $33\%$ of the total data available will be tried to find. The parameters of the GGA that are adjusted are the probability of mutation $p\in [0, 0.3]$, the percentage of confrontation in the tournament $q\in[0.5,1.0]$, and the number of epochs $e\in [50, 200]$. On the other hand, the parameters of the ELM that is used to evaluate the fitness in the GGA are also adjusted. These parameters are the number of hidden units $n\in[50,150]$ and the logarithm of the regularization constant of a ridge regression estimator, that is used to find the weights of the output layer $\gamma \in [-15,-3]$. Note that a ridge regression estimator for the output layer weights allows for a more flexible model than the standard ELM, as the standard ELM is retrieved when $\gamma$ is negative and large \cite{Albert72}.

The BO method is compared with two techniques. The first technique is a random exploration of the space of parameters. The second technique is a configuration specified by a human expert. Namely, $p=0.02$, $q=0.8$, $e=200$, $n=150$ and $\gamma = -10$. These are reasonable values that are expected to perform well in the specific application tackled. The computational budget to $50$ different parameter evaluations is set for both the BO and the random exploration strategy. After each evaluation, the performance of the best solution found is reported. The experiments are repeated for $50$ different random seeds and average results are informed. All BO experiments are carried out using the acquisition function EI and the software for BO Spearmint.

\subsection{Results I: Bayesian optimization of the wave energy prediction system parameters}
\begin{figure}[ht]
\centering
\includegraphics[width=0.75\textwidth]{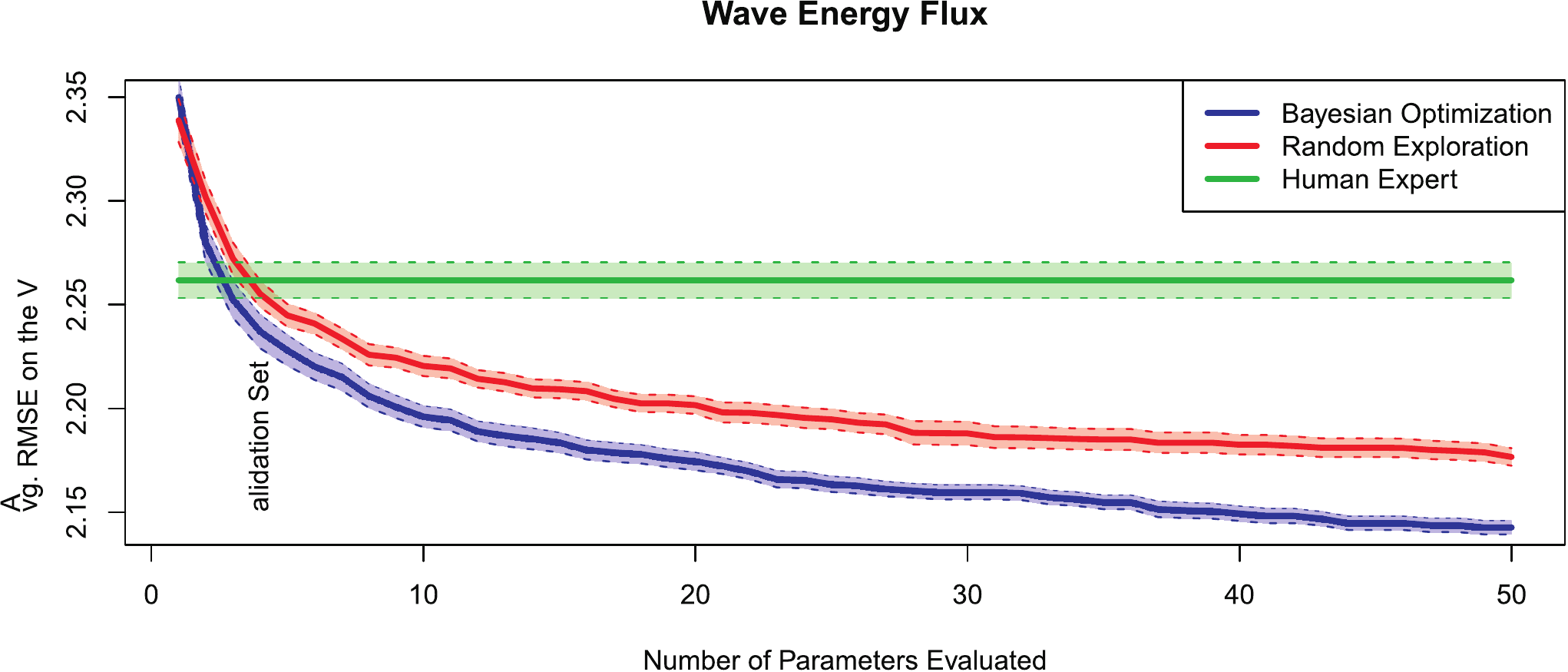}
\caption{\label{bo:experiments}
	Average results obtained for the $P$ optimization after evaluating the performance of 50 different parameters for the BO technique and a random exploration of the parameter space. The performance a configuration specified by a human expert is also shown for comparison.
}
\end{figure}

\begin{figure}[ht]
\centering
\includegraphics[width=0.75\textwidth]{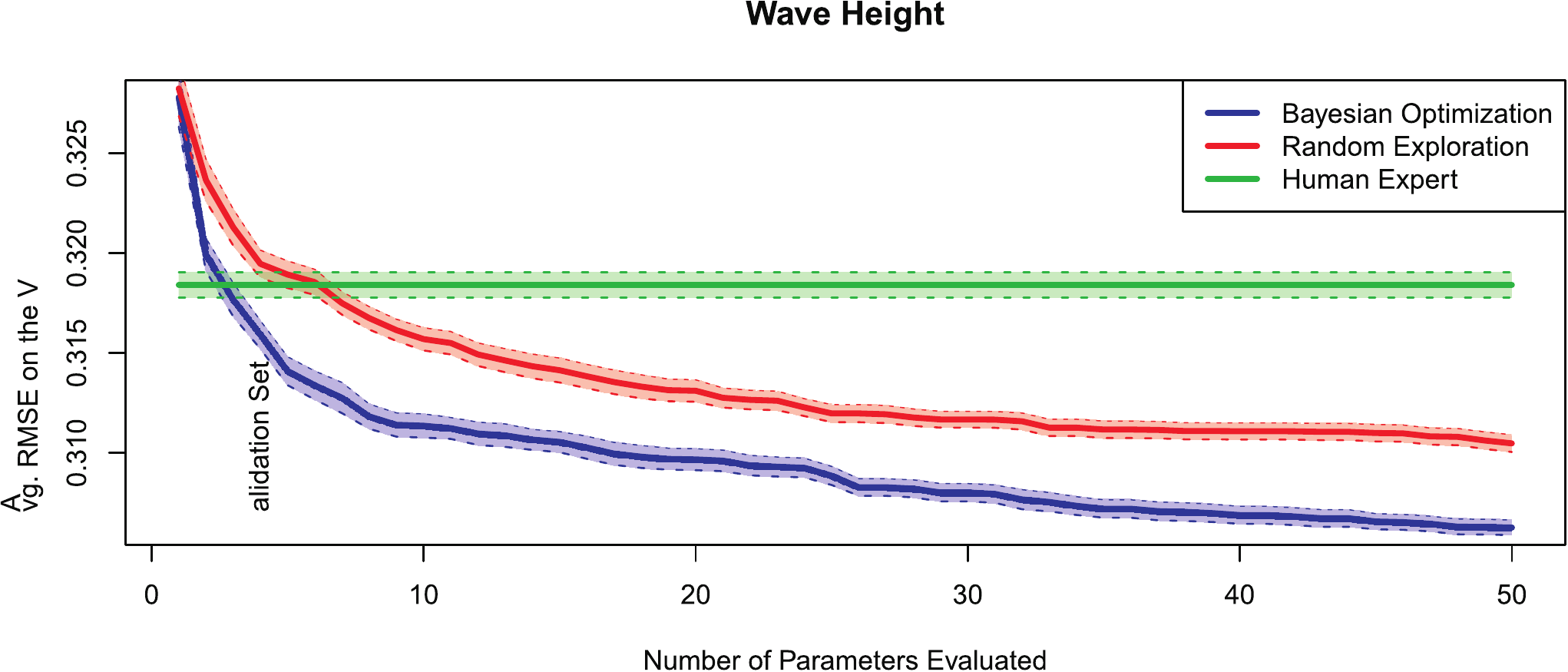}
\caption{\label{bo:hexperiments}
	Wave Height optimization average results of the performance of the 50 different parameter values
	selected by the BO technique and a random exploration of the parameter space.
	The plot also shows the performance of the parameter values selected by a human expert.
}
\end{figure}

Figures \ref{bo:experiments} and \ref{bo:hexperiments} show the average results obtained and the corresponding error bars for the task of predicting the $P$ and the task of predicting the wave height, respectively. Each figure shows the average RMSE of each method (BO and random exploration) on the validation set as a function of the number of configurations evaluated. The performance of the configuration specified by a human expert is also shown. It can be observed that the BO strategy performs best in each setting. In particular, after a few evaluations the BO method is able to outperform the results of the human expert and it provides results that are similar or better than the ones obtained by the random exploration strategy with a smaller number of evaluations.

\subsection{Results II: Estimation of the generalization performance}
In a second round of experiments, the performance of the proposed prediction system after its optimization with the BO methodology is shown. Note that, after the FS process with the GGA-ELM approach, an ELM and a SVR \cite{tutorial98,Salcedo14} to obtain the final prediction of the $P$ and the $H_s$ are used.

Table \ref{Comparativa_2010_F} shows the results obtained for the experiments carried out. It can be observed the comparison between ELM and SVR approaches in different scenarios: the prediction obtained with all the features, the prediction obtained with the hybrid algorithm GGA-ELM (without BO methodology), and finally the prediction acquired after the application of the BO process in the GGA-ELM approach. As Table \ref{Comparativa_2010_F} summarizes, it is easy to see how the hybrid GGA-ELM algorithm improves the results obtained by the ELM and SVR approaches (without FS). In fact, the SVR algorithm improves the values of the Pearson's Correlation Coefficient ($r^2$) around 75\% in the case of the FS method, against the poor 31\% when all features are used. Moreover, these results are improved by means of the BO methodology, using ELM and SVR approaches after the GGA-ELM. In the case of the ELM, values of the $r^2$ around 77\% against the 71\% achieved with the GGA-ELM algorithm without the BO improvement are obtained. The same behavior is get for the SVR algorithm: values around 78\% with the application of the BO methodology against the 75\% obtained for the GGA-ELM approach when the parameters are fixed by a human expert. In addition, the reader can comparer the results with other measurement of the accuracy, the \ac{MAE}.

\begin{table}[ht]
\begin{center}
\caption{\label{Comparativa_2010_F} Comparative results of the $P$ estimation by the ELM and SVR approaches after the FS by the GGA-ELM in 2010.} \vspace{0.3cm}
\begin{tabular}{cccc}
\hline
Experiments             & RMSE             & MAE         & $r^2$ \\
\hline
All features-ELM\       & 3.4183 kW/m\     & 2.4265 kW/m & 0.6243\  \\
All features-SVR\       & 4.4419 kW/m\     & 2.8993 kW/m & 0.3129\    \\
GGA-ELM-ELM\            & 2.8739 kW/m\     & 1.8715 kW/m & 0.7101\    \\
GGA-ELM-SVR \           & 2.6626 kW/m\     & 1.6941 kW/m & 0.7548\    \\
BO-GGA-ELM-ELM\         & \textbf{2.5672 kW/m}\     & \textbf{1.7596 kW/m} & \textbf{0.7722}\    \\
BO-GGA-ELM-SVR \        & \textbf{2.4892 kW/m}\     & \textbf{1.6589 kW/m} & \textbf{0.7823}\    \\
\hline
\end{tabular}
\end{center}
\end{table}

\begin{figure}[ht]	
  \begin{center}
	\begin{tabular}{cc}
   \subfigure[]{\includegraphics[width=0.49\textwidth]{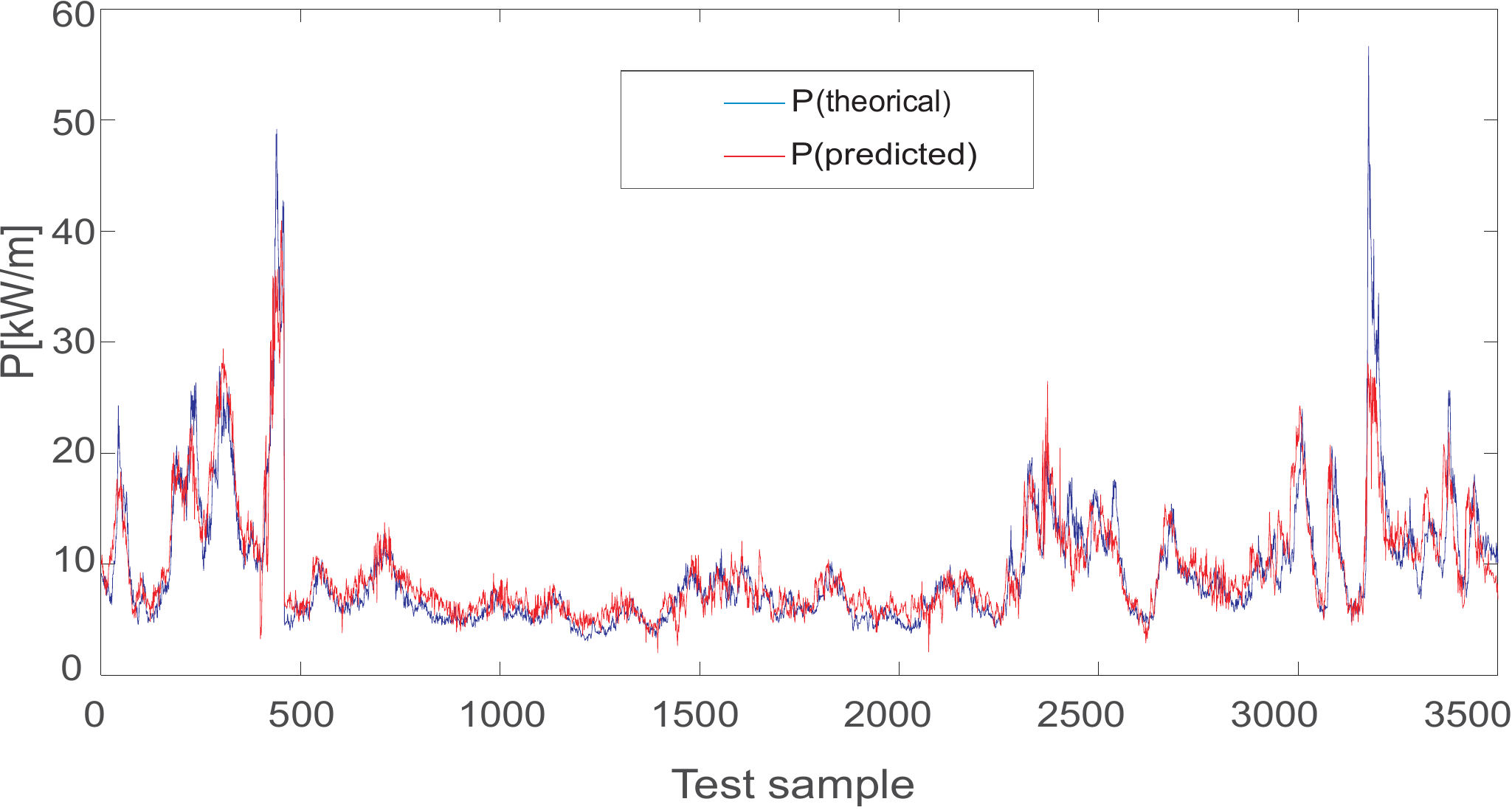}} &
    \subfigure[]{\includegraphics[width=0.49\textwidth]{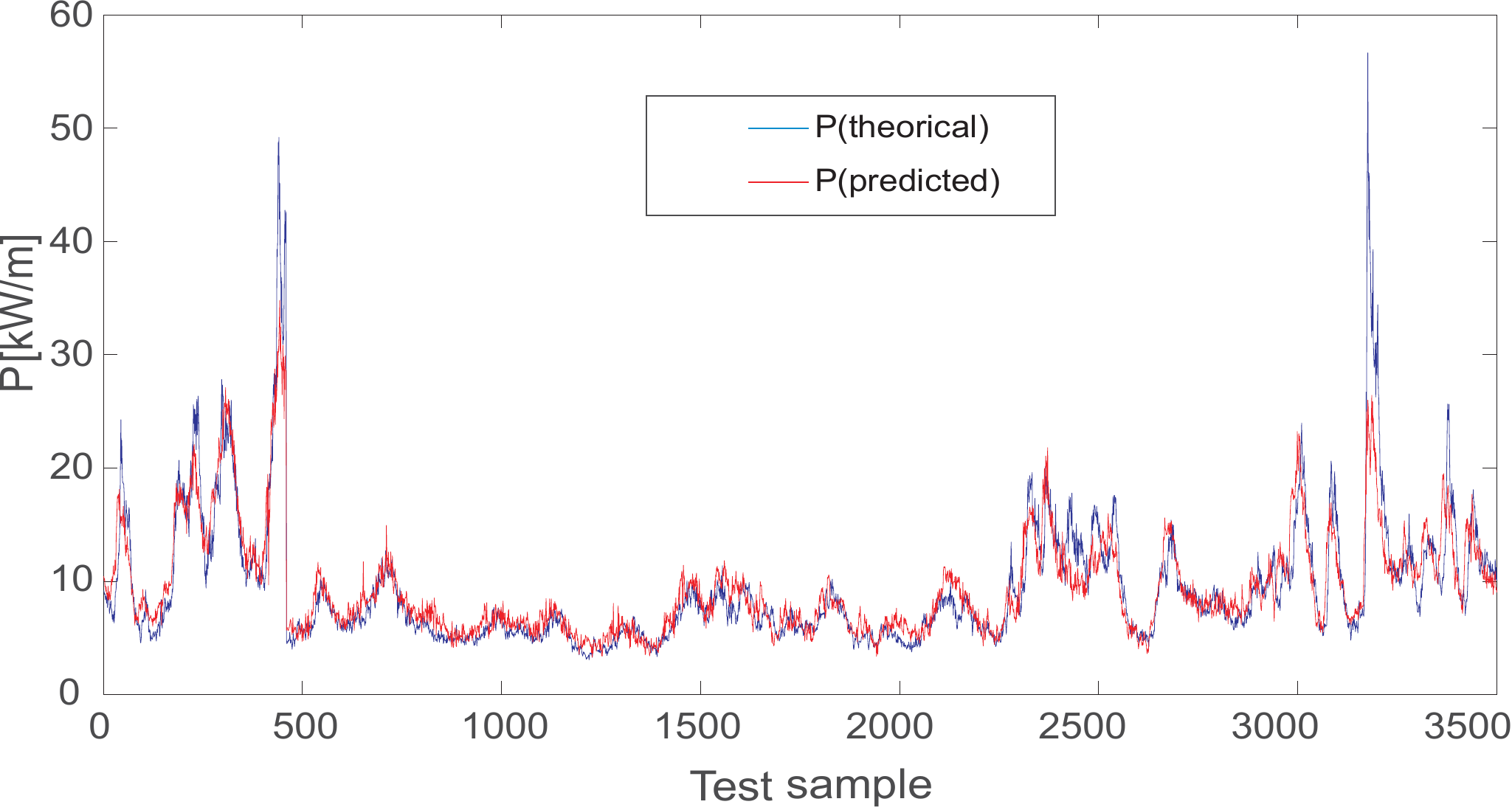}} \\
    \subfigure[]{\includegraphics[width=0.49\textwidth]{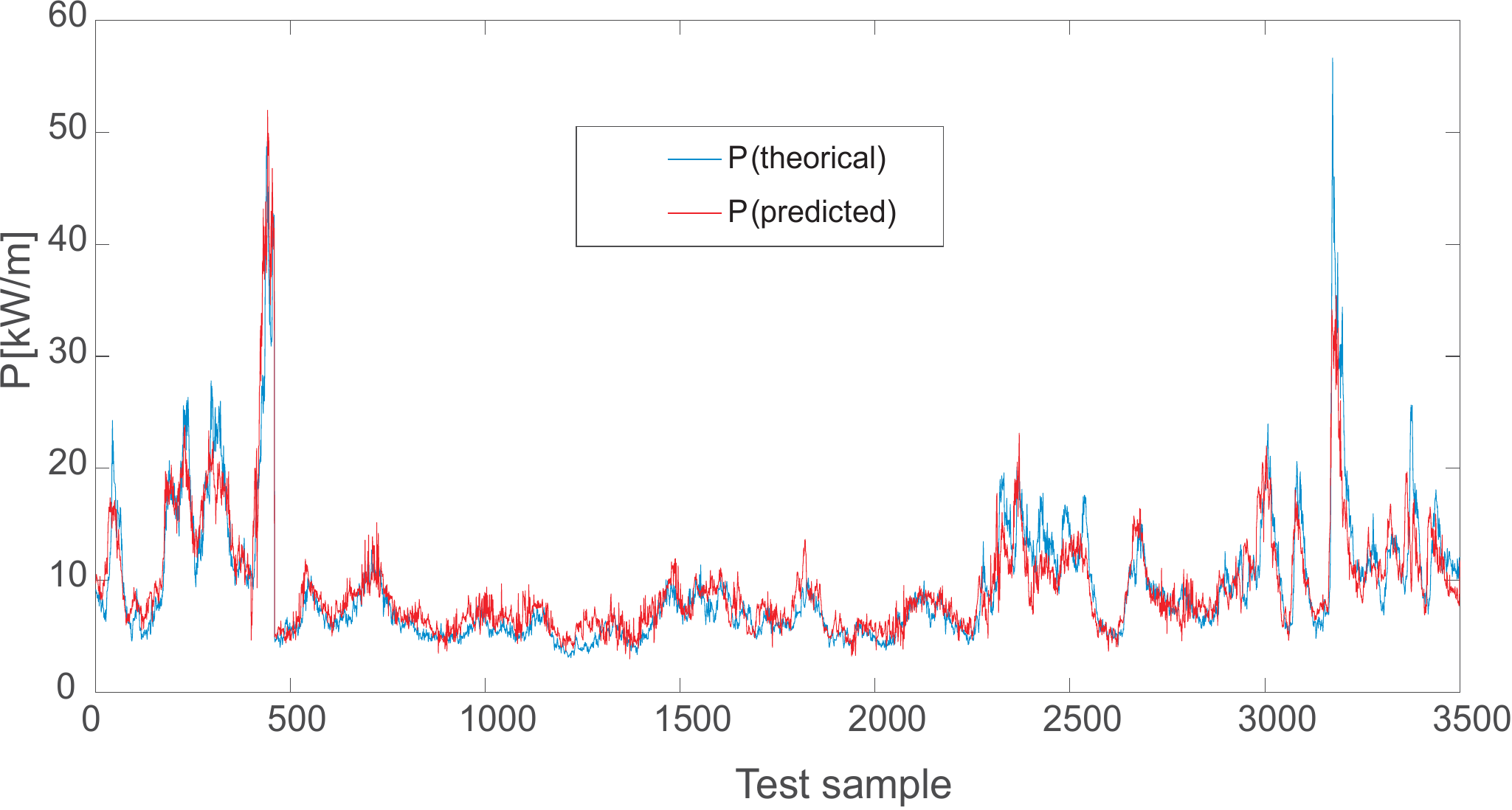}} &
    \subfigure[]{\includegraphics[width=0.49\textwidth]{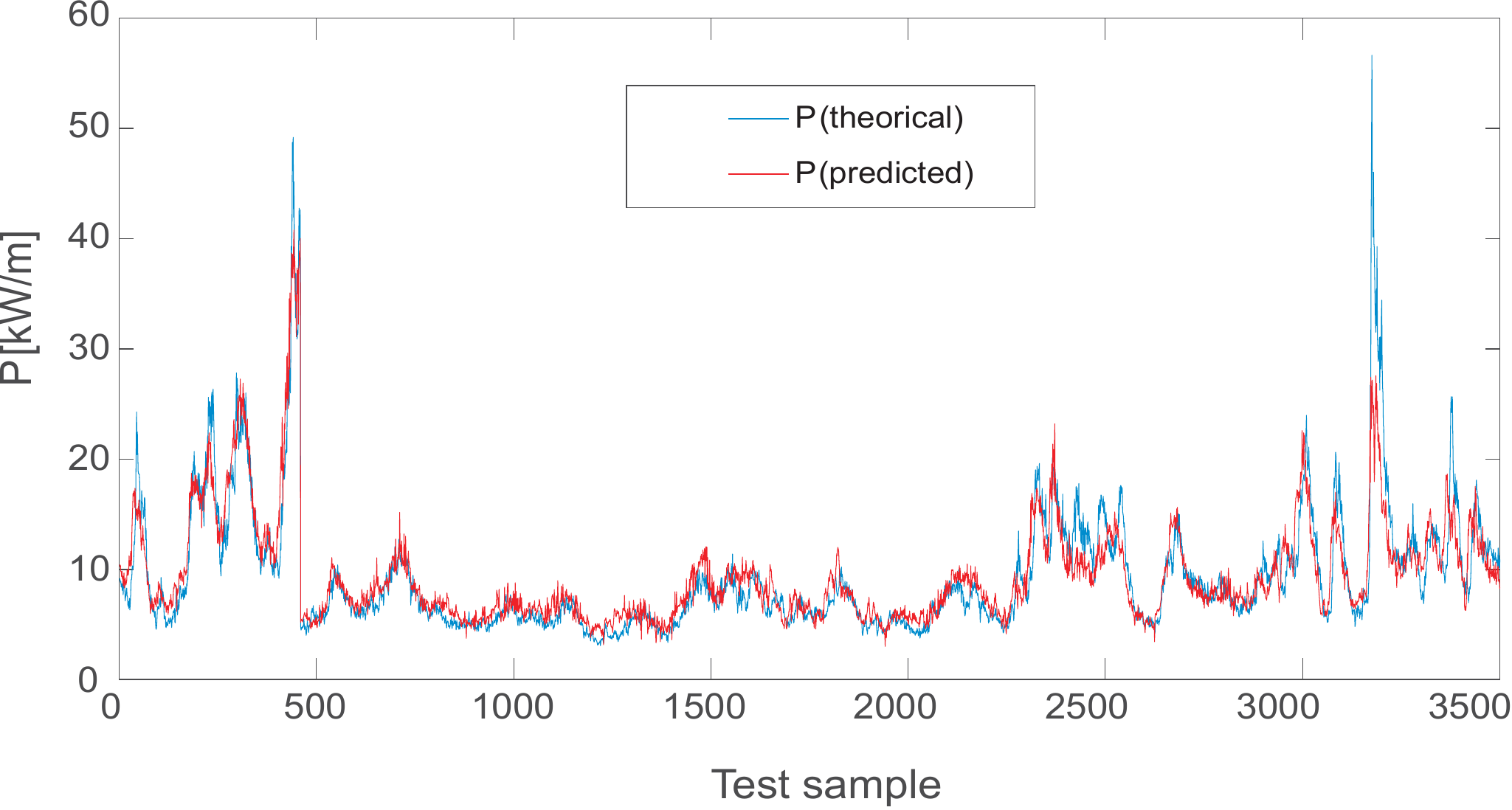}}
	\end{tabular}
  \end{center}
  \caption{$P$ prediction after the FS process with the GGA-ELM approach; (a) ELM; (b) SVR; (c) ELM with BO; (d) SVR with BO.}
\label{F_GGA}
\end{figure}

\begin{figure}[ht]	
  \begin{center}
   \subfigure[]{\includegraphics[width=0.4\textwidth]{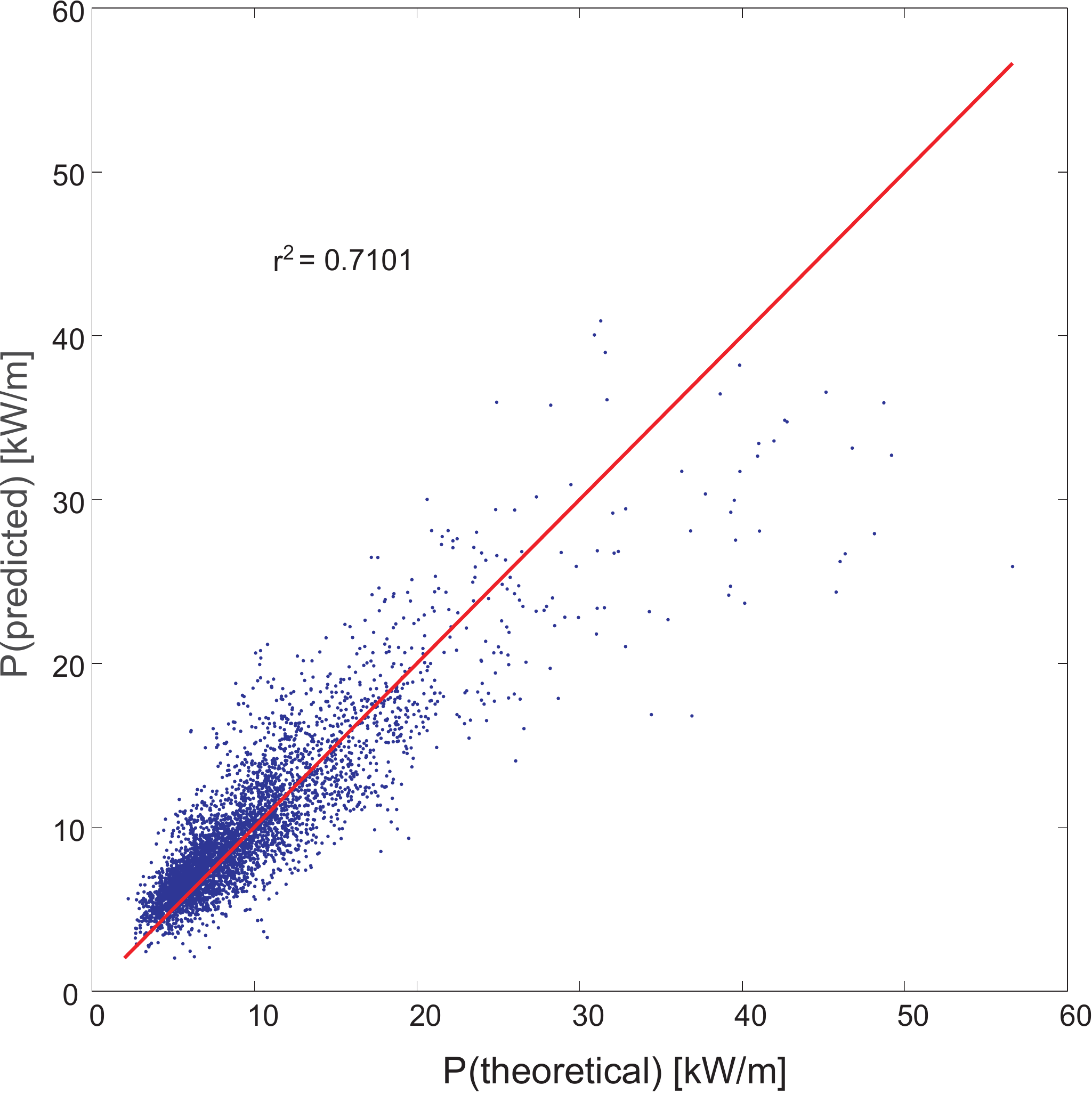}}
    \subfigure[]{\includegraphics[width=0.4\textwidth]{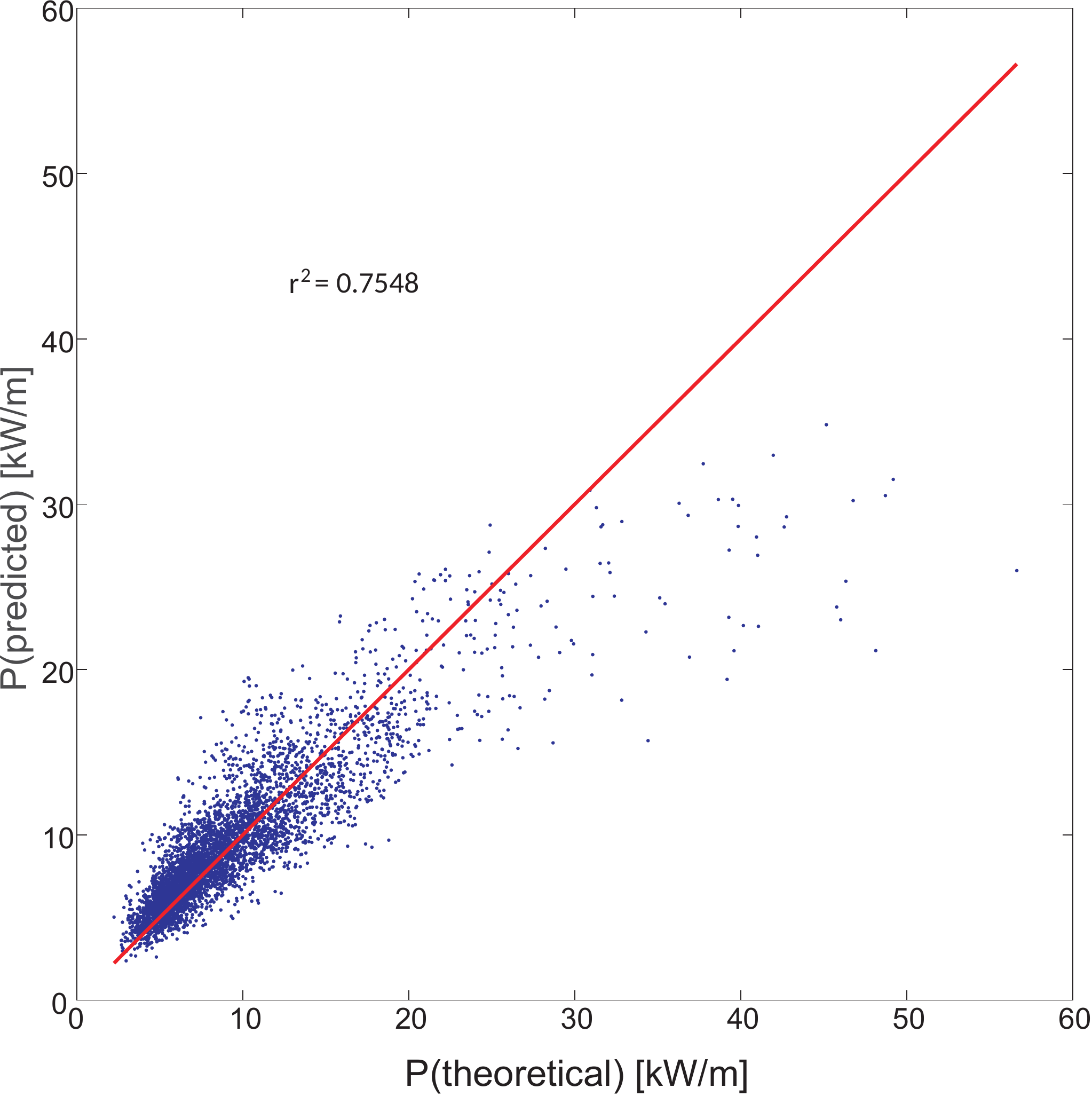}}\\
    \subfigure[]{\includegraphics[width=0.4\textwidth]{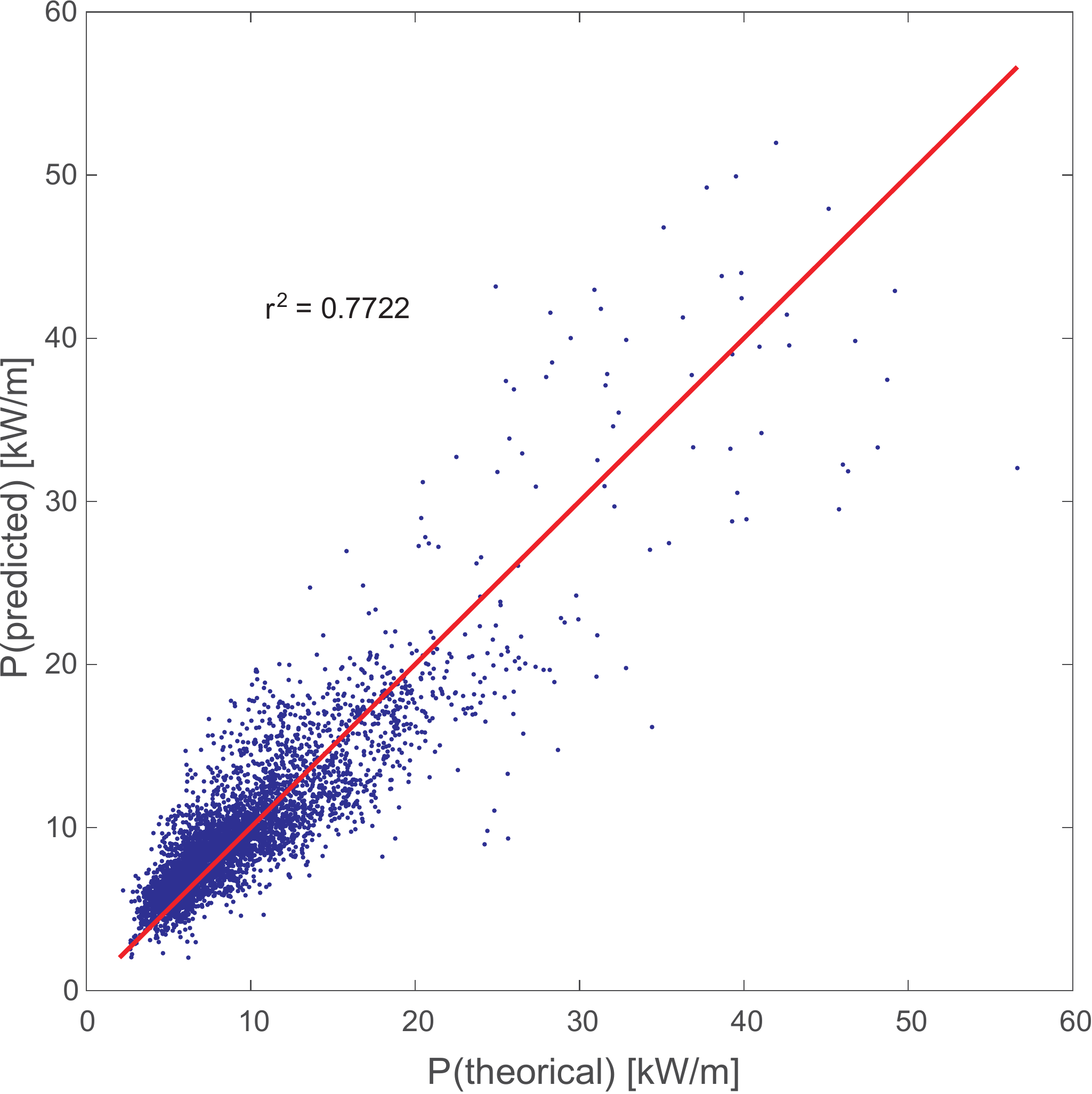}}
    \subfigure[]{\includegraphics[width=0.4\textwidth]{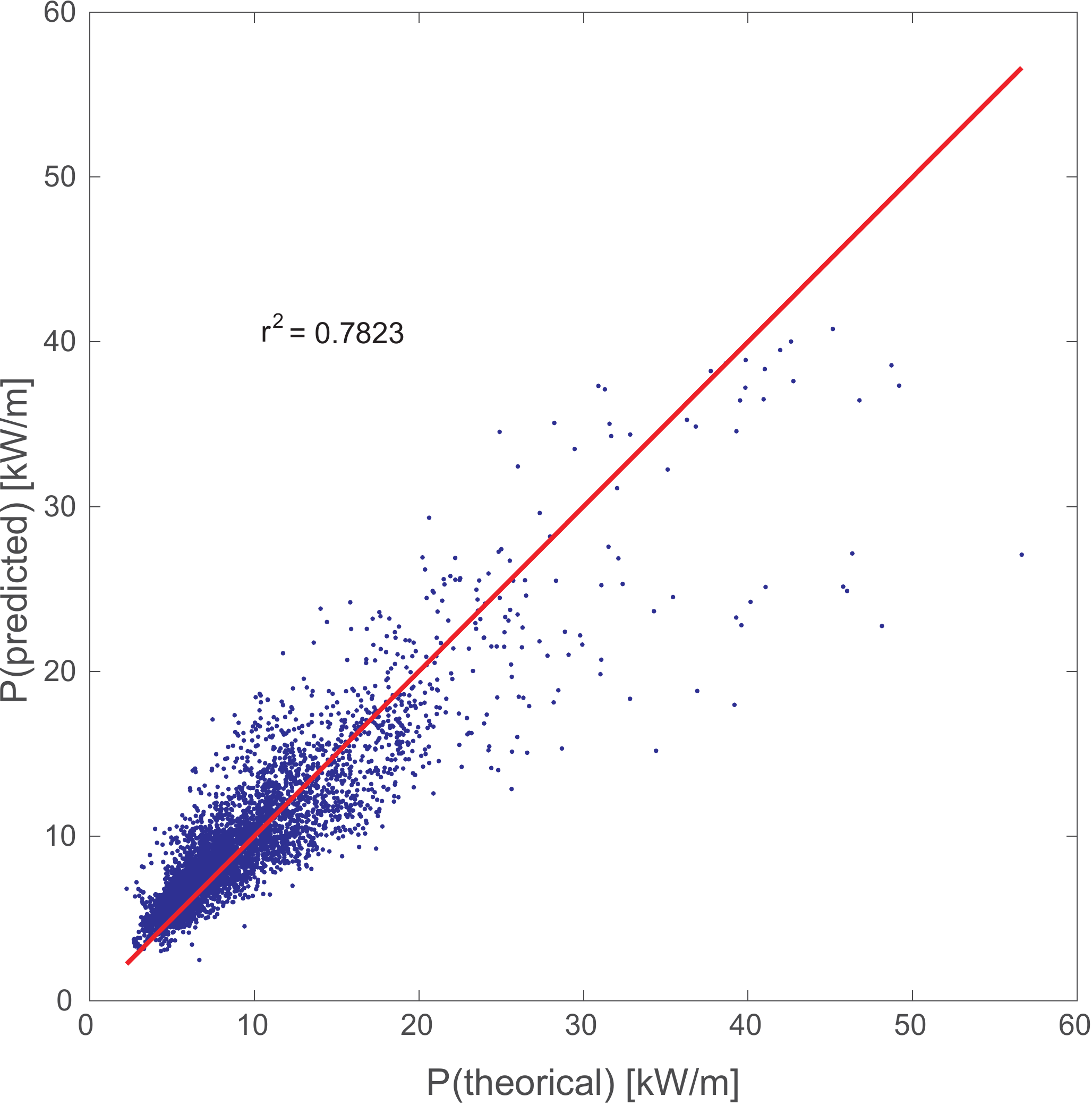}}
  \end{center}
  \caption{Scatter plots in the problem of $P$ prediction in tackled by the ELM and SVR with FS by the GGA-ELM; (a) ELM; (b) SVR; (c) ELM with BO; (d) SVR with BO.}
\label{P_scatter_GGA}
\end{figure}

The results of the previous tables can be better visualized in the following graphics. In Figure \ref{F_GGA} the temporary predictions carried out by the ELM and SVR approaches are shown. It can be seen how the cases (c) and (d) improve the approximation to the real values against the cases (a) and (b) where the BO methodology is not applied. The same situation can be seen in Figure \ref{P_scatter_GGA}, where the scatter plots are presented for the results obtained with and without the BO methodology.

\begin{figure}[ht]	
  \begin{center}
	\begin{tabular}{cc}
   \subfigure[]{\includegraphics[width=0.49\textwidth]{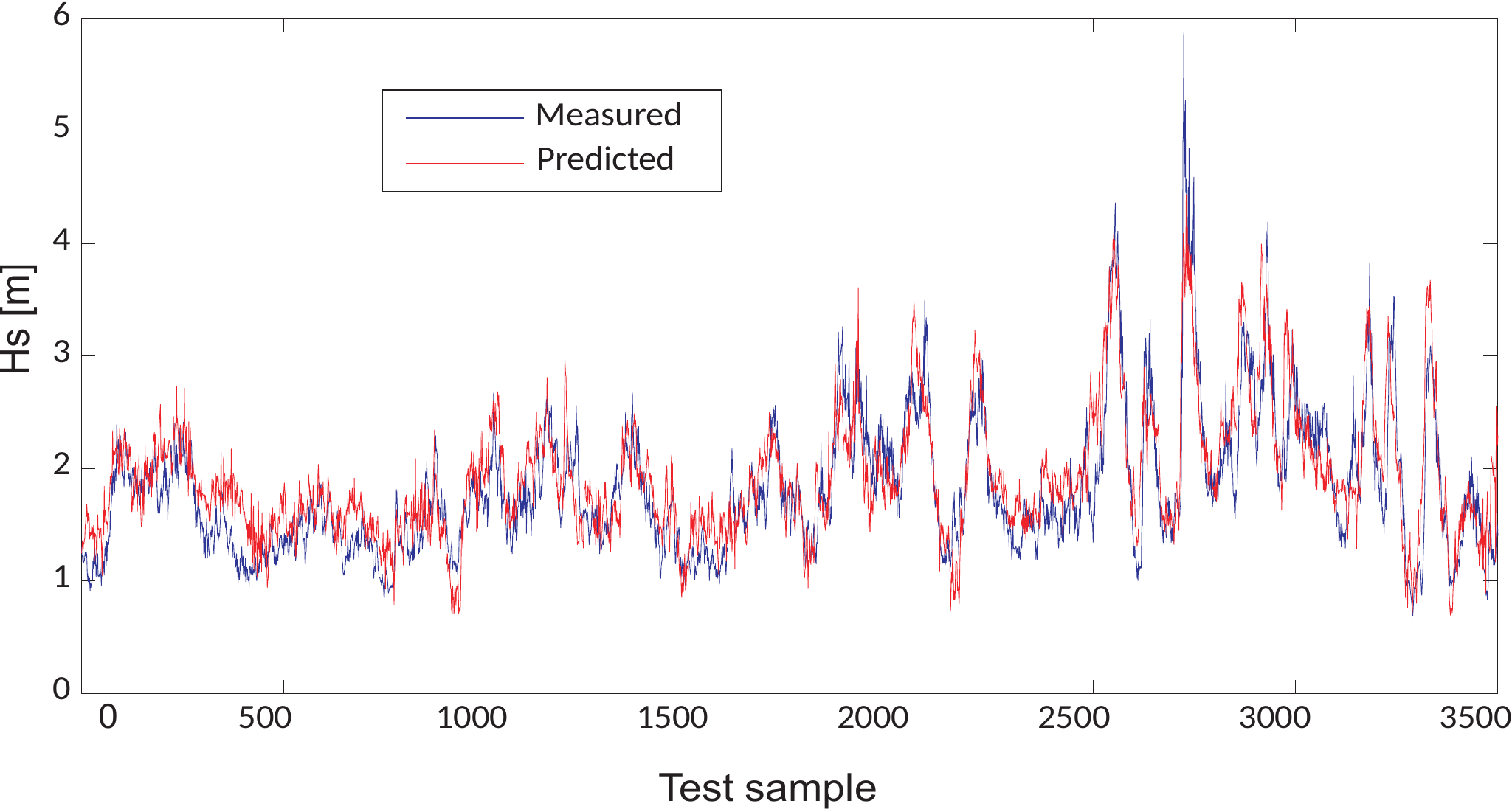}} &
    \subfigure[]{\includegraphics[width=0.49\textwidth]{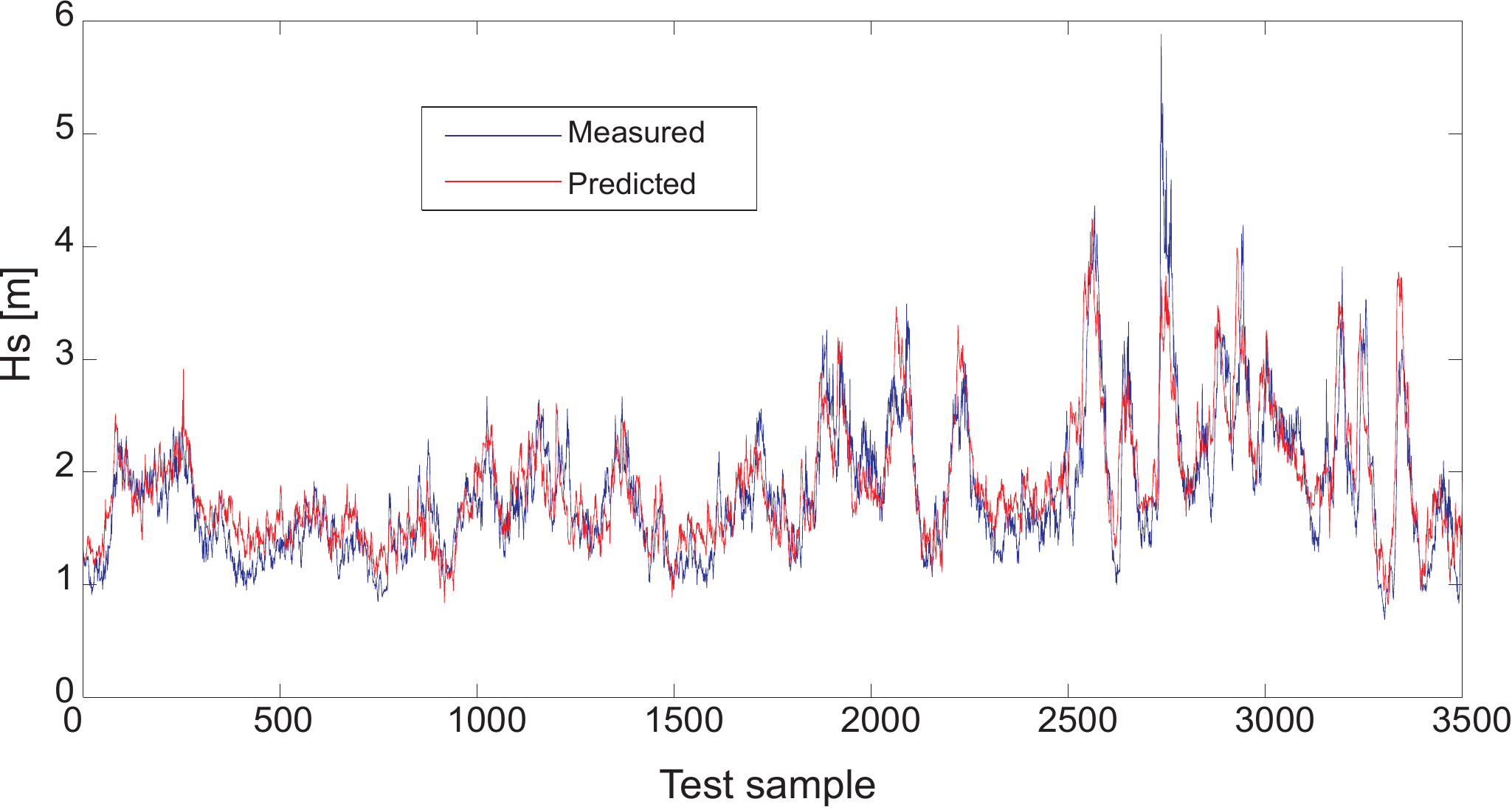}} \\
    \subfigure[]{\includegraphics[width=0.49\textwidth]{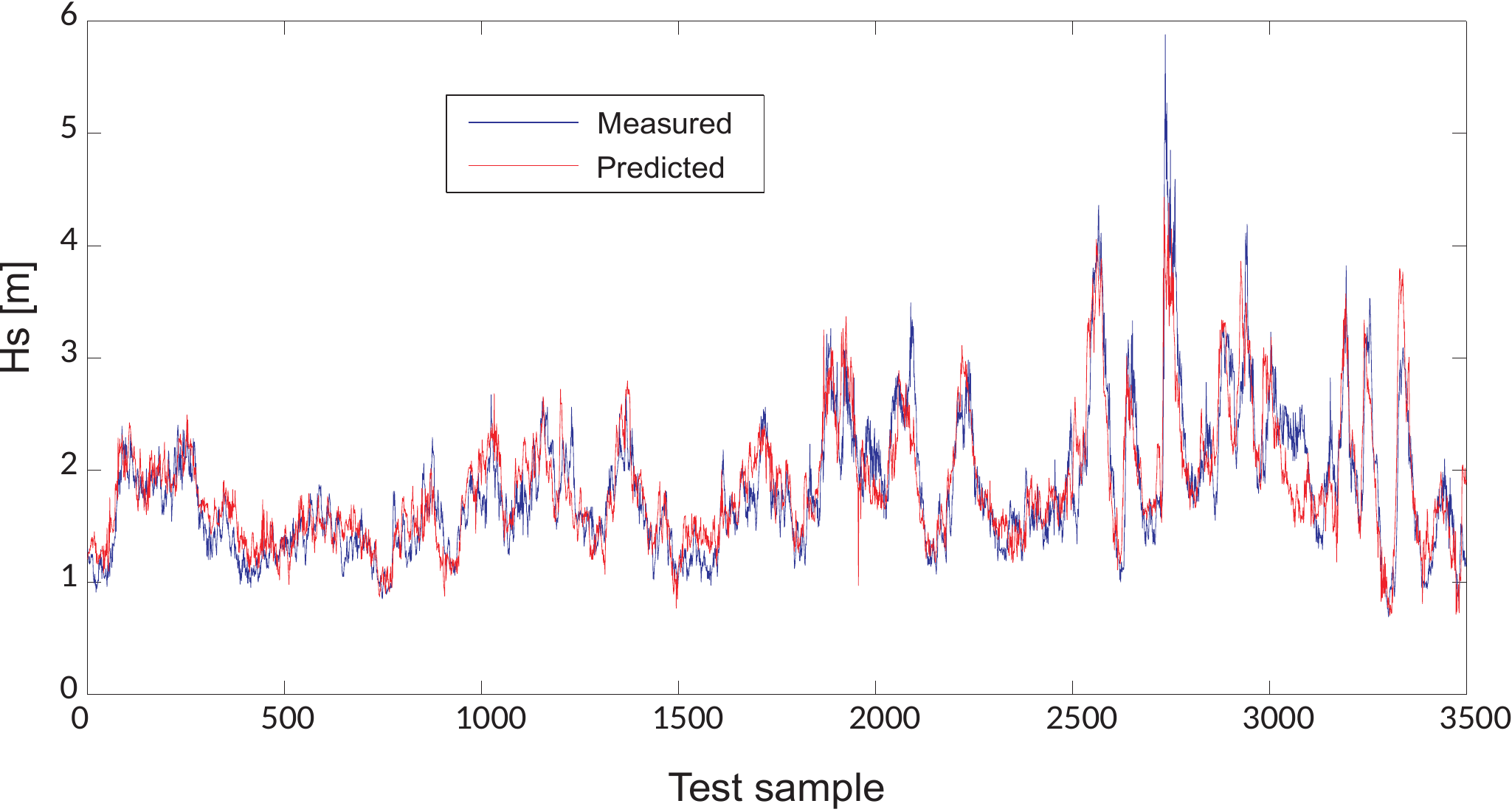}} &
    \subfigure[]{\includegraphics[width=0.49\textwidth]{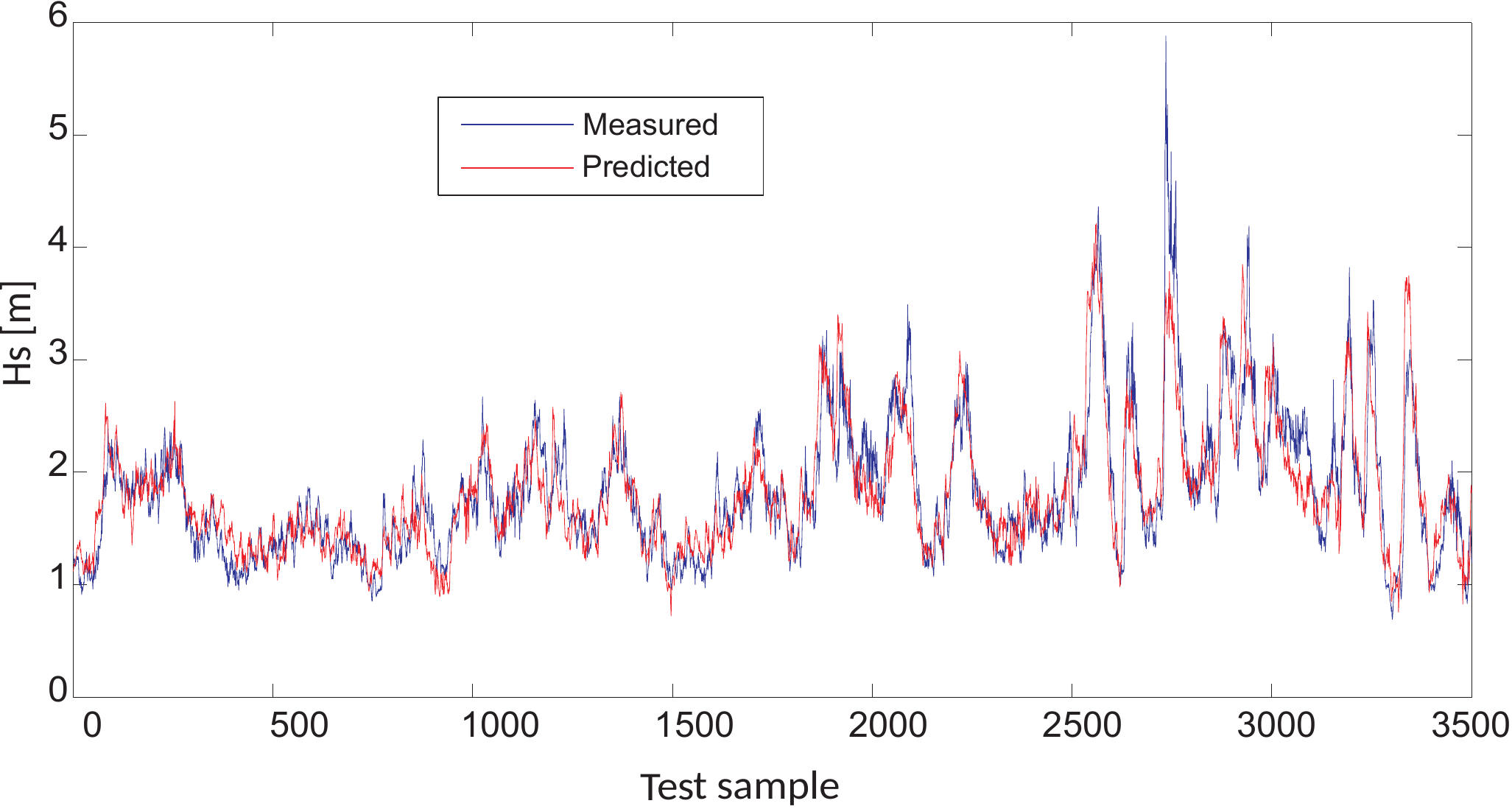}}
	\end{tabular}
  \end{center}
  \caption{$H_s$ prediction after the FS process with the GGA-ELM approach; (a) ELM; (b) SVR; (c) ELM with BO; (d) SVR with BO.}
\label{Hs_GGA}
\end{figure}

\begin{figure}[ht]	
  \begin{center}
   \subfigure[]{\includegraphics[width=0.4\textwidth]{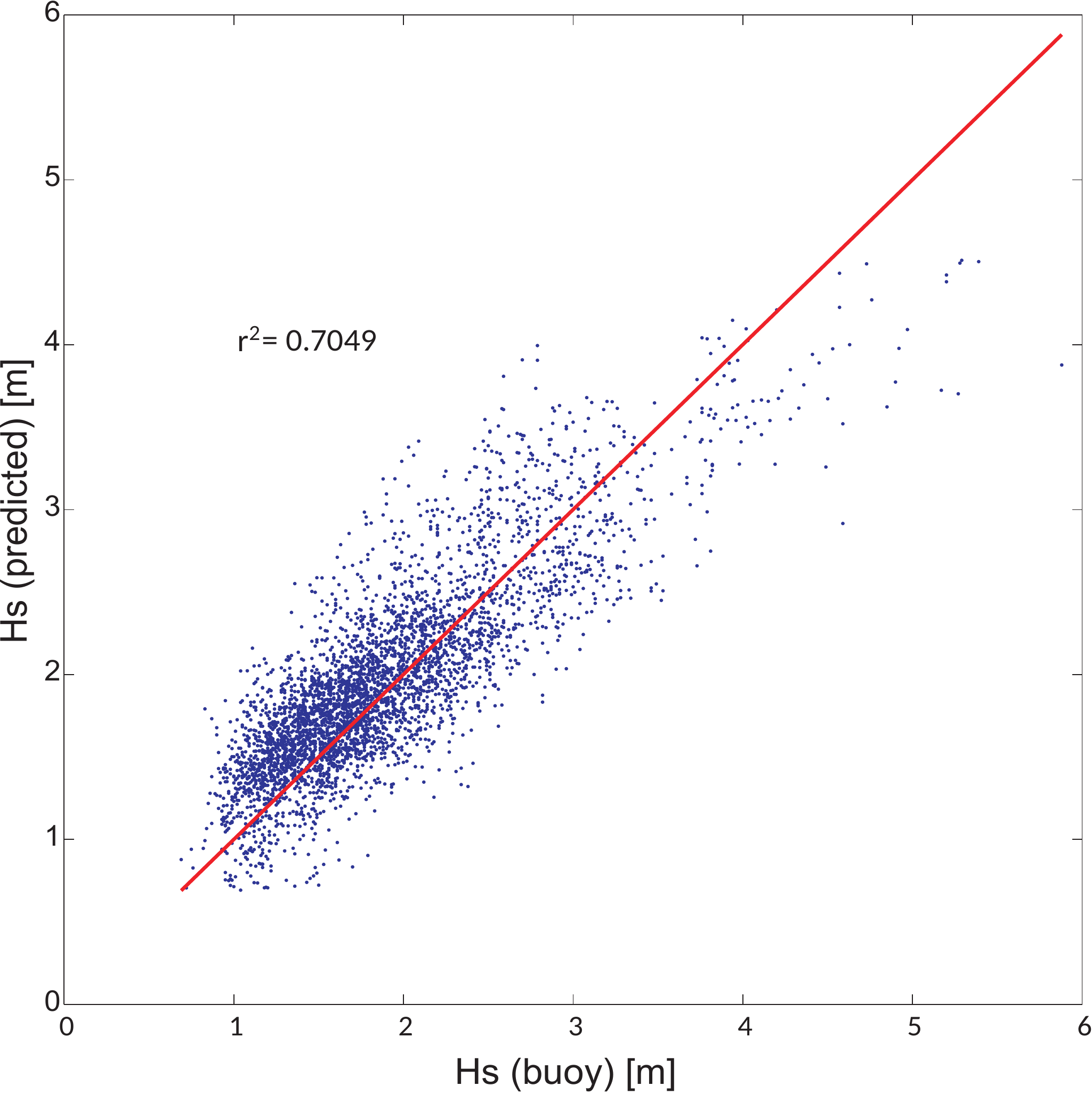}}
    \subfigure[]{\includegraphics[width=0.4\textwidth]{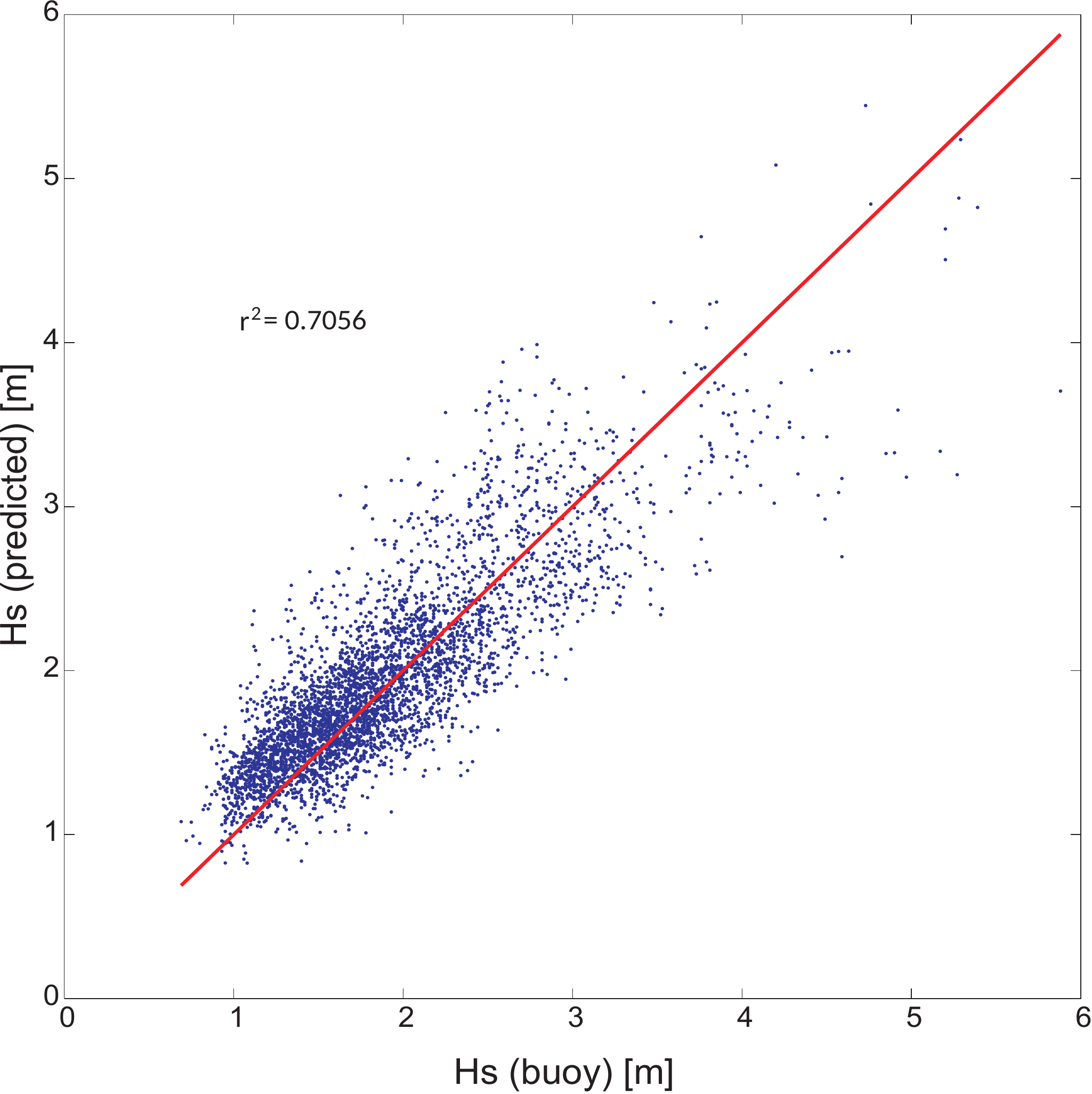}}\\
    \subfigure[]{\includegraphics[width=0.4\textwidth]{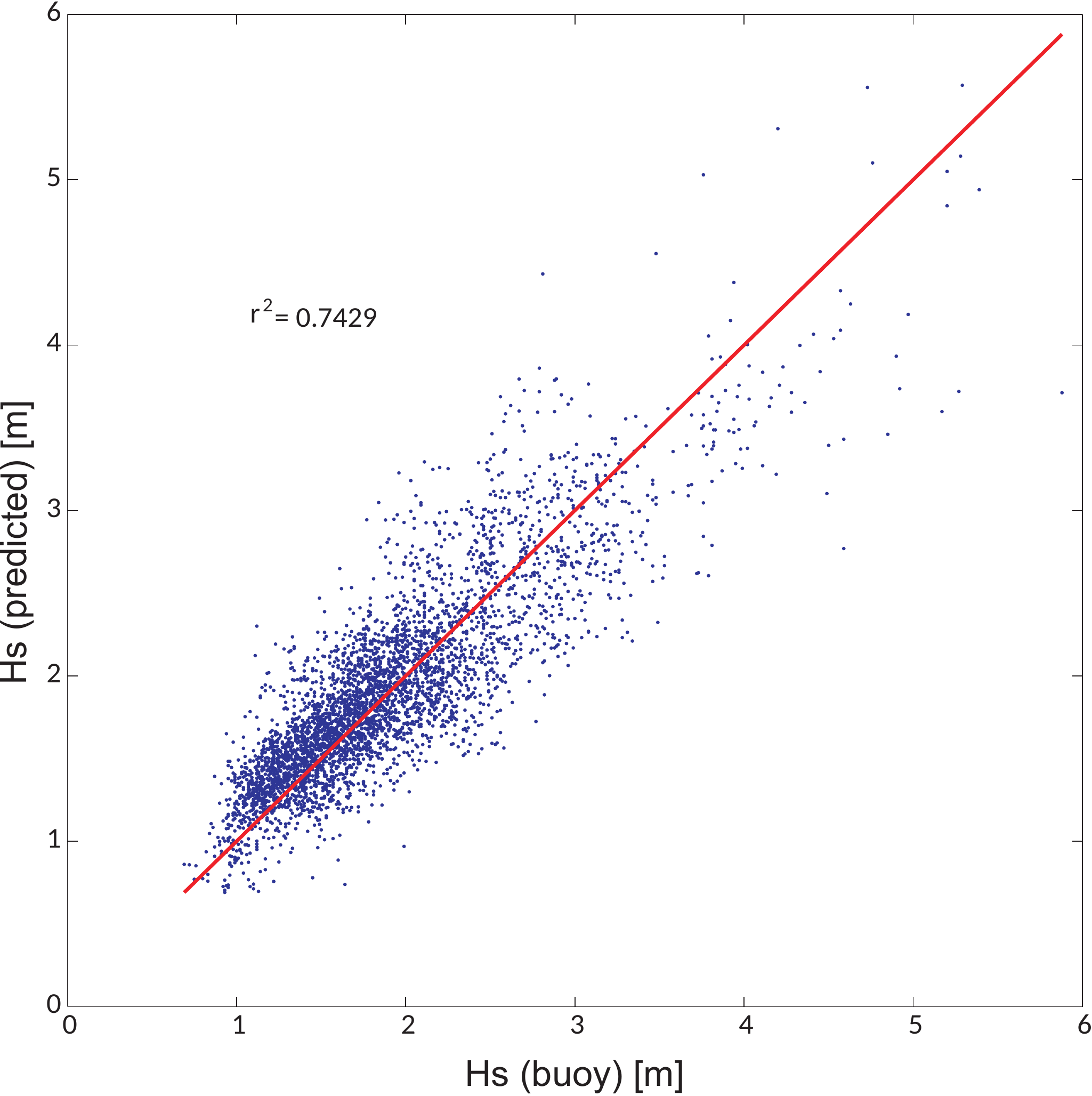}}
    \subfigure[]{\includegraphics[width=0.4\textwidth]{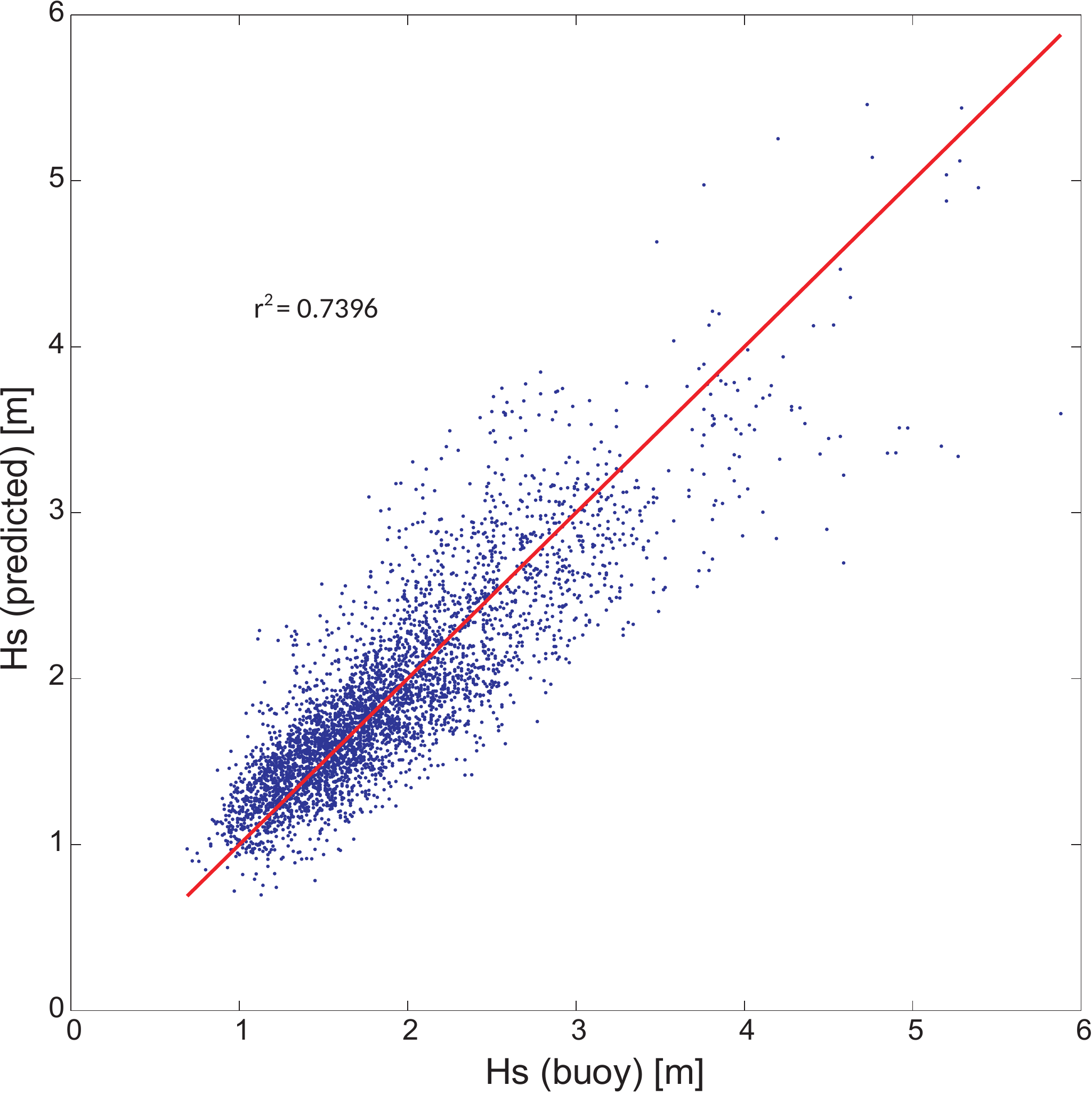}}
  \end{center}
  \caption{Scatter plots in the problem of $H_s$ prediction in tackled by the ELM and SVR with FS by the GGA-ELM; (a) ELM; (b) SVR; (c) ELM with BO; (d) SVR with BO.}
\label{Hs_scatter_GGA}
\end{figure}

The same procedure is carried out in the case of the $H_s$. Table \ref{Comparativa_2010_Hs} compares the results obtained in the different experiments. As it can be seen, the results are improved with the use of the BO methodology with values of the $r^2$ around 74\% for the ELM and SVR predictions, against the 66\% and 39\% achieved for the ELM and SVR, respectively, with all features. The GGA-ELM algorithm improves these last results, but they are not so good like when the BO methodology is used. In Figures \ref{Hs_GGA} the temporary predictions for the GGA-ELM-ELM, GGA-ELM-SVR, BO-GGA-ELM-ELM and BO-GGA-ELM-SVR are shown. The same is done for the scatter plots, whose Figures \ref{Hs_scatter_GGA}, present the results mentioned above.

\begin{table}[ht]
\begin{center}
\caption{\label{Comparativa_2010_Hs} Comparative results of the $H_s$ estimation by the ELM and SVR approaches after the FS by the GGA-ELM in 2010.} \vspace{0.3cm}
\begin{tabular}{cccc}
\hline
Experiments             & RMSE             & MAE         & $r^2$ \\
\hline
All features-ELM\       & 0.4653 m\     & 0.3582 m       & 0.6624\  \\
All features-SVR\       & 0.6519 m\     & 0.4986 m       & 0.3949\    \\
GGA-ELM-ELM\            & 0.3650 m\     & 0.2858 m       & 0.7049\    \\
GGA-ELM-SVR \           & 0.3599 m\     & 0.2727 m       & 0.7056 \    \\
BO-GGA-ELM-ELM\         & \textbf{0.3324 m}\     & \textbf{0.2519 m}       & \textbf{0.7429}\    \\
BO-GGA-ELM-SVR \        & \textbf{0.3331 m}\     & \textbf{0.2461 m}       & \textbf{0.7396} \    \\
\hline
\end{tabular}
\end{center}
\end{table}

In both predictions ($P$ and $H_s$) the BO methodology improves the results, for this reason the generality of the proposed method can be highlighted.

\section{Conclusions}\label{Conclusions1}

In this paper it has been shown how a hybrid prediction system for wave energy prediction can be improved by means of BO methodology. The prediction system is formed by a grouping GA for FS, and an ELM for effective prediction of the target variable, the $P$ and the $H_s$ in this case. After this FS process, the final prediction of the target is obtained by means of an ELM or a SVR approach. The paper describes in detail the BO methodology, and its specific application in the optimization of the GGA-ELM for a real problem of $P$ and $H_s$ prediction from buoys data in Western California USA. The results show that the BO methodology is able to improve the performance of the system, i.e., the prediction of the optimized system is significantly better than that of the system without the BO methodology applied. This improvement is related to the optimal selection of parameters carried out by the BO strategy. On the other hand, the main limitation of the proposed methodology is the increase in computation time. Nevertheless, this increase only affects the training phase and not the operation phase, in which predictions are made after training. Therefore, this limitation is not very important. Finally, note that this methodology can be extended to alternative prediction systems and other problems, specially to hybrid approaches involving ML algorithms with a high number of parameters to be tuned.

\chapter{Wind power ramps events prediction}\label{cap:wind}
\section{Introduction}
Wind Power Ramp Events (WPREs) are large fluctuations of wind power in a short time interval, which lead to strong, undesirable variations in the electric power produced by a wind farm. Its accurate prediction is important in the effort of efficiently integrating wind energy in the electric system, without affecting considerably its stability, robustness and resilience. In this study, the problem of predicting WPREs by applying ML regression techniques is tackled. The proposed approach consists of using variables from atmospheric reanalysis data as predictive inputs for the learning machine, which opens the possibility of hybridizing numerical-physical weather models with ML techniques for WPREs prediction in real systems. Specifically, the feasibility of a number of state-of-the-art ML regression techniques are explored, such as SVR, ANN (MLPs and ELMs) and GPs to solve the problem. Furthermore, the ERA-Interim reanalysis from the European Center for Medium-Range Weather Forecasts is the one used in this work because of its accuracy and high resolution (in both spatial and temporal domains). Aiming at validating the feasibility of this predicting approach, an extensive experimental work using real data from three wind farms in Spain is carried out, discussing the performance of the different ML regression tested in this wind power ramp event prediction problem.

\subsection{Motivation}
Wind power is currently one of the most important renewable energies in the world \cite{Kumar16} in terms of penetration in the electric power system \cite{Brenna17,Mohagheghi17}, economic impact and annual growth rate~\cite{Ali17}, both~onshore \cite{dai2016key} and offshore \cite{colmenar2016offshore}. Electric power generation is usually carried out in large wind \mbox{farms \cite{giebel2016overview,herbert2014review}} far from urban centers \cite{lunney2017state,jangid2016potential}, though, in the last few years, urban wind power generation is also gaining impulse \cite{simoes2016new}, including its use in smart grids \cite{kokturk2017vision}.

The counterpart of the benefits associated with the flourishing of wind energy throughout the world---mainly the reduction of CO$_2$ emission, one of the causes of global warming \cite{peters2013challenge} and climate change \cite{bauer2015co}---are problems related not only to the maintenance and management of wind farm facilities, but also to those of power grids. Regarding this, one of the most important problems yet to be solved is the efficient {integration} \cite{jones2017renewable} of an increasing number of wind energy generators in both the distribution and transmission power grids, which are becoming increasingly complex \cite{Cuadra17,Cuadra15}. Such an intrinsically complex nature of power grids is further increased because of the inherent stochastic nature of wind energy \cite{Yan15} that, depending on the weather conditions, can lead to {intermittent} generation \cite{Yan15}. This~can affect the stability, robustness and resilience \cite{Cuadra17,Cuadra15} of electric power grids. A useful discussion of the technical differences between these interrelated, but distinct concepts can be found in \cite{Cuadra15}.

Aiming at preserving grid {stability} in a scenario with a high percentage of intermittent renewable sources---not only wind energy \cite{colmenar2016offshore}, but also photovoltaic \cite{cabrera2016review} and wave \cite{cuadra2016computational} energies---power grids need to be made more flexible \cite{kroposki2017achieving}. In this effort, the emerging technologies associated with smart grids~\cite{kokturk2017vision} and micro-grids \cite{yoldacs2017enhancing} can be used to mitigate wind power intermittency. An illustrative, very recent proposal in this respect consists of increasing the penetration of \ac{V2G} technologies~\cite{gough2017vehicle} to use the batteries of idle \ac{EV} as power storage units \cite{zhao2017boosting}, absorbing peaks of intermittent overproduction.

Wind power intermittency and its influence on power grids' stability and performance are the main reasons why \ac{WPF} \cite{Taslimi16,Tasci14} is a key factor to improve its integration without unbalancing the rest of the grid components. Among the different issues in wind power prediction, one of the most significant is the existence of Wind Power Ramp Events. WPREs consist of large fluctuations of wind power in a short period of time, leading to a significant increasing or decreasing of the electric power generated in a wind farm \cite{zhang2017ramp,Gallego15a}.

The field of scientific research in WPREs' prediction (or forecasting) \cite{Ouyang13} is a relatively recent topic driven by the need for improving the management of quick and large variations in wind power output, particularly in the aforementioned context of power grids with high renewable penetration~{\cite{alizadeh2016flexibility}. A~useful review of different WPREs' definitions (in which there does not seem to be a clear consensus) and their types (increasing or decreasing, depending on the WPRE definition) can be found in \cite{ferreira2011survey}. Among them, } WPREs' severity is one of the important issues. Up and down WPREs can exhibit different fluctuating levels of severity, although down WPREs are usually more critical than up WPREs because of the availability of reserves \cite{zhang2017ramp}. WPREs are usually caused by specific meteorological processes---basically, crossing fronts \cite{Gallego15b} and fast changes in the local wind direction---and they involve at several scales (synoptic \cite{Ohba16}, mesoscale \cite{Salcedo09} and microscale). Surprisingly, it has been found recently that very large offshore farms, clustered together, can also generate large WPREs on time scales of less than \mbox{6 h \cite{drew2017importance}}. This gives an idea of the complexity of the WPRE phenomenon.

WPREs' prediction is not only important for power grid operators, but also for wind farm owners. {In fact, the occurrence of WPREs in wind farms is critical not only because of the aforementioned undesired variations of power, but also due to their potential harmful effects in wind turbines,} which~leads to an increase of management costs associated with these facilities \cite{Cui15a}. Regarding~this, the~accurate prediction of WPREs has been reported as an effective method to mitigate the economic impact of these events in wind generation power plants \cite{Gallego15a,Cui15a}.

According to \cite{Cui15a,foley2012current}, {the prediction of WPREs and their influence on electricity generation and grid stability have been recently tackled by using two major families of techniques: (1) ``physical-based'' models (or numerical approaches aiming to tackle the complexity of the physical equations, which rule the atmosphere to obtain a prediction); and (2) statistical approaches (usually data-driven models to obtain predictions).} The first group of techniques, the physical-based approaches, include~a set of equations that rules the atmospheric processes and their evolution over time and,~because~of their complexity and nonlinearity, are tackled by means of numerical methods. The second group of WRPE predicting techniques, the statistical approaches, are data-driven methods that are based on wind time series and include a variety of techniques ranging from conventional approaches---for instance, \ac{ARMA}---to \ac{CI} approaches~\cite{salcedo2016modern}. These~are physics-inspired meta-heuristics \cite{salcedo2016modern} able to find approximate solutions to complex problems that otherwise could not be solved or would require very long computational time. They include, among others, three groups of bio-inspired techniques such as \ac{EC} \cite{de2006evolutionary}, \ac{NC} \cite{ata2015artificial} and \ac{FC}~\cite{suganthi2015applications}. An introduction to the main concepts of bio-inspired CI techniques in energy applications can be found in \cite{cuadra2016computational,salcedo2015coral}.

\subsection{Purpose and Contributions}
The purpose of this work is to explore the feasibility of a novel hybrid WPRE prediction framework, which merges parts of numerical-physical models with state-of-the-art statistical approaches.
When the term ``hybrid algorithms'' is used in this work, that means that this proposal combines data from numerical-physical methods (reanalysis, in this case) with ML approaches (specifically, regressors). Regarding what the hybrid approach means in this study, there are two points to note. The first one is that it would be possible to adapt the proposed regression techniques to operate with alternative data (not~coming from numerical methods, reanalysis, in this study). The second one, which is the main novelty of this work, is that the use of data from numerical-physical methods could help achieve valuable prediction of WPREs in wind farms.

{The contributions of this work are:
\begin{enumerate}
\item The use of regression techniques in this kind of problem since, up until now, the majority of WPRE prediction frameworks have been based on classification approaches.
\item The use of reanalysis data as predictive variables of the ML regression techniques. As~will be shown, this is because the direct application of regression algorithms makes unnecessary the use of some pre-processing algorithms, which are necessary in other \mbox{approaches \cite{Dorado17,Cornejo17}}. Note~that the classification problems associated with WPREs are usually highly unbalanced, which~makes it difficult to put into practice high-performance classification techniques without having to use specific over-sampling or similar techniques \cite{Dorado17,Cornejo17}.
\item The performance of the proposed system has been tested using real data from three different wind farms in Spain.
\end{enumerate}}

The rest of this chapter is organized as follows: Section \ref{Problem_definition} states the problem definition we tackle in this work, in which the WPRE prediction is formulated as a regression task. Section \ref{data} presents the data and predictive variables involved in our proposal. In turn, Section \ref{Experiments} shows the experimental work carried out, these results being obtained by the different tested algorithms in three WPRE prediction problems located at three distinct wind farms in Spain. Sections \ref{sec:Discussion} and \ref{Conclusions} complete the study by giving some final concluding remarks on the work carried out.

\section{Problem Definition}\label{Problem_definition}
Following previous works in the literature \cite{Gallego15a,Dorado17,Cornejo17,dorado2017combining}, a WPRE can be characterized by a number of parameters:

\begin{itemize}
\item Magnitude ($\Delta P_r$): defined as the variation in power produced in the wind farm or wind turbine during the ramp event (subscript ``$r$'').
\item Duration ($\Delta t_r$): time period during which the ramp event is produced.
\end{itemize}

In addition to the magnitude and duration of a wind ramp, the derived quantity called the ramp rate ($\Delta P_r / \Delta t_r$) is used to define the intensity of the ramp.

Taking these parameters into account, in the majority of previous works in the literature, the WPRE detection problem has been defined as a classification problem \cite{Bossavy15}. {Within this framework,} let $S_t: \mathbb{R}^k \rightarrow \mathbb{R}$ be the so-called ramp function, i.e., a criterion function that is usually evaluated to decide whether or not there is a WPRE. There are several definitions of $S_t$, all~of them involving power production ($P_t$) criteria at the wind farm (or wind turbine), but the two more common ones are the following \cite{Gallego15a}:

\begin{equation}\label{Sdef1}
S_t^1=P_{t+\Delta t_r}-P_t
\end{equation}
\begin{equation}\label{Sdef2}
S_t^2=\max([P_t,P_{t+\Delta t_r}])-\min([P_t,P_{t+\Delta t_r}])
\end{equation}

Note that, in the ramp function $S_t^1$ stated by Equation (\ref{Sdef1}), the power variation is referred to a given time interval $\Delta t_r$. In the experimental work carried out throughout this work, such a time interval has been assumed to be $\Delta t_r = 6$ h (the ``reference time interval'') because of the reanalysis resolution.

Using any of these definitions of the ramp function $S_t$, the classification problem can be stated by defining a threshold value $S_0$, in the way:

\begin{equation}\label{threshold_S0}
I_{t}=\left\{
\begin{array}{l l}
1& \mathrm{,~if}~~S_t \geq S_0\\
0& \mathrm{,~otherwise}\\
\end{array}
\right.
\end{equation}
where $I_{t}$ is an ``indicator function'' to be used to label the data in the binary classification formulation of the problem.

As will be shown later on, in this approach, first of all, the threshold value $S_0$ is set, and then, a WPRE is detected if the ramp function is larger than 50\% of $S_0$. It is worth mentioning that, if there is an interest in establishing a larger number of cases (for example, five classes of WPRE), it would need at least two thresholds to do so.

The WPRE detection problem also involves a vector of predictive variables ${\bf x}$. Different types of inputs have been used as predictive variables in the literature. The key point here is that the meteorological process must be always considered, since they are physical precursors of WPREs. Different numerical weather prediction system outputs have been used to obtain these predictive variables, including reanalysis data \cite{Gallego15b}. This provides a long history record of meteorological variables to be used as predictive variables for WPRE prediction. Following these previous works, in this paper, the following version of the WPRE prediction problem is tackled:

Let ${\bf X}_t=\{x_{1}, \ldots, x_{l}\}$ (with $t=1, \ldots, l$) be time series of $l$ predictive variables and $l$ values of the ramp function $S_t$ (objective variables). The problem consists of training a regression model $\mathcal{M}$ in a subset of $({\bf X}_t, S_t)^{\mathbb{T}}$ (training set), in such a way that, when $\mathcal{M}$ is applied to a given test set $({\bf X}_t, S_t)^{\mathcal{R}}$, an~error measure $e$ is minimized.

\section{Data and Predictive Variables}\label{data}
A reanalysis project is a methodology carried out by some weather forecasting centers, which~consists of combining past observations with a modern meteorological forecast model, in~order to produce regular gridded datasets of many atmospheric and oceanic variables, with a temporal resolution of a few hours. Reanalysis projects usually extend over several decades and cover the entire planet, being a very useful tool for obtaining a comprehensive picture of the state of the Earth system, which can be used for meteorological and climatological studies. There are several reanalysis projects currently in operation, but one of the most important is the ERA-Interim reanalysis project, which~is the latest global atmospheric reanalysis produced by the ECMWF \cite{ERA_Interim}. ERA-Interim is a global atmospheric reanalysis from 1979, continuously updated in real time. The data assimilation system used to produce ERA-Interim is based on a 2006 release that includes a four-Dimensional Variational analysis (4D-Var) with a 12-h analysis window. The spatial resolution of the dataset is approximately 15 km, on 60 vertical levels from the surface up to 0.1 hPa. ERA-Interim provides six-hourly atmospheric fields on model levels, pressure levels, potential temperature and potential vorticity and three-hourly surface fields.

Aiming to tackle the WPRE prediction problem in this study, wind and temperature-related predictive variables is considered from ERA-Interim at some specific points in the neighborhood of the area under study. The variables considered as predictors (Table \ref{Variables_ERA}) are taken at different pressure levels (surface, 850 hPa and 500 hPa), in such a way that different atmospheric processes can be taken into account. A total of 12 prediction variables per ERA-Interim node and four nodes surrounding the area under study (wind farm) are considered at time $t$, i.e., in this problem, ${\bf X}_t$ is formed by $N=48$ predictive variables. The ERA-Interim time resolution for the predictive variables (6 h) sets in this case the ramp duration taken into account ($\Delta t_r=6$).

Thus, each regression model analyzed in this work ($\mathcal{M}$) must be trained with the data $({\bf X}_t,S_t^1)^{\mathbb{T}}$ or $({\bf X}_t,S_t^2)^{\mathbb{T}}$, where $S_t^1$ and $S_t^2$ are computed using Equations (\ref{Sdef1}) and (\ref{Sdef2}), respectively.

\begin{table}[ht]
\begin{center}
\caption{\label{Variables_ERA} Predictive variables considered at each node from the ERA-Interim reanalysis.}\vspace{0.3cm}
\begin{footnotesize}
\begin{tabular}{cc}
\toprule
\textbf{Variable Name} & \textbf{ERA-Interim Variable}\\
\midrule
skt & surface temperature\\
sp & surface pression\\
$u_{10}$& zonal wind component ($u$) at 10 m\\
$v_{10}$& meridional wind component ($v$) at 10 m\\
temp1& temperature at 500 hPa\\
up1& zonal wind component ($u$) at 500 hPa\\
vp1& meridional wind component ($v$) 500 hPa\\
wp1& vertical wind component ($\omega$) at 500 hPa\\
temp2& temperature at 850 hPa\\
up2& zonal wind component ($u$) at 850 hPa\\
vp2& meridional wind component ($v$) at 850 hPa\\
wp2& vertical wind component ($\omega$) at 850 hPa\\
\bottomrule
\end{tabular}
\end{footnotesize}
\end{center}
\end{table}

\section{Experimental Work}\label{Experiments}
This section presents the experimental evaluation of the proposed approach in a real problem of WPRE prediction, by exploring the different ML regressors used in this work (SVR, ELM, GP and~MLP). {Prior to describing the experiments carried out, it is worth emphasizing the practical importance of using reanalysis data to test the accuracy and feasibility of the proposed hybrid approach with ML regressors. Non-hybrid approaches (the use of regression techniques in other alternative data, from measuring stations, for example) is also possible. However, note that, from the viewpoint of the {repeatability} of the experiments, reanalysis data are very convenient since they are freely available on the Internet, so that the experimental part of this work can be easily reproduced by other researchers.}

{Starting with the detailed description of the experimental work carried out, three wind farms are considered in Spain, whose locations have been represented in Figure \ref{fig:maps}. The three wind farms chosen (labeled ``A'', ``B'' and ``C'' in Figure \ref{fig:maps}) are medium-sized facilities, with 32, 28 and 30 turbines installed, respectively. Note that the wind farms selected cover different parts of Spain, north, center and south, characterized by different wind regimes. Different numbers of data were available for each wind farm: in wind farm ``A'', data ranges  11/01/2002--29/10/2012, while in wind farm ``B'' ranges 23/11/2000--17/02/2013. In wind farm ``C'', the data used are between 02/03/2002 and 30/06/2013.

\begin{figure}[ht]
\begin{center}
\includegraphics[draft=false, angle=0,width=13cm]{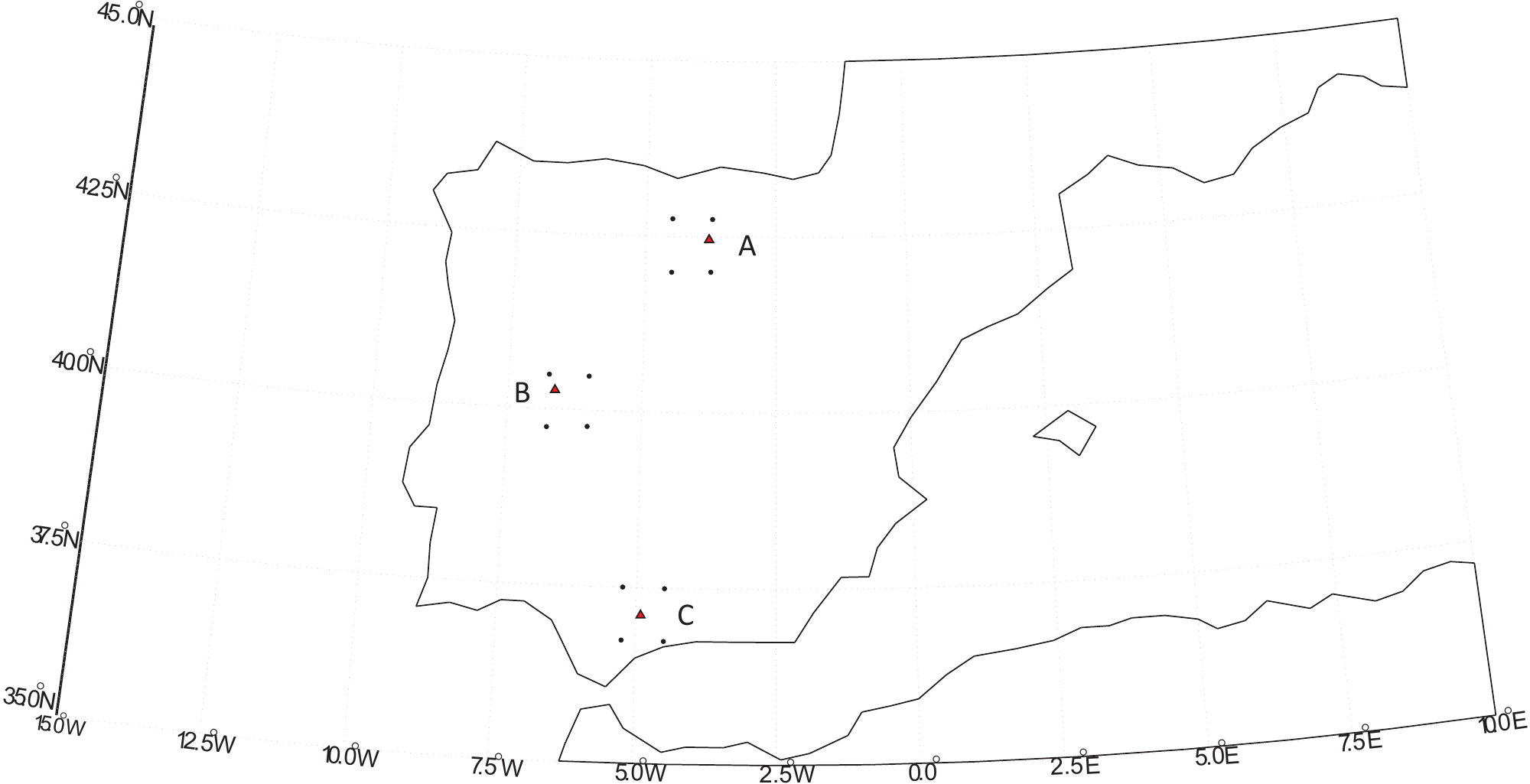}
\end{center}
\caption{ \label{fig:maps} Representation of the geographical location of the wind farms (labeled ``A'', ``B'' and~``C'') considered in the experimental work carried out in this thesis. The four closest nodes from the Era-Interim reanalysis (predictive variables) have also been represented for illustrative purposes. {The~reason why these wind farms have been selected is that they cover different parts of Spain, north, center and south, characterized by different wind regimes.}}
\end{figure}

A pre-processing step to remove missing and corrupted data was carried out. Note that data every 6 h (00 h, 06 h, 12 h and 18 h) is only kept, to match the predictive variables from the ERA-Interim to the objective variables.

The performance of the four ML regressors described in Section \ref{sec:state_of_the_art}, in WPREs prediction problems at each wind farm is shown in terms of different error measurements ($e$), such as RMSE, \ac{MAE} or ``sensitivity'', $s$, also called the true positive rate. This last measure is defined as:

\begin{equation}\label{eq:sensitivity}
s = \frac{\mathcal{NP}}{NP}\times 100,
\end{equation}

where: (1) $\mathcal{NP}$ stands for the number of positive predictions, i.e., the correct predictions of ascending ($+$), descending ($-$) and no ramps (with the $S_t^1$ definition), and ramps or no ramps (with the $S_t^2$ definition) values in the experiments; (2) $NP$ stands for the number of positive values in the test, i.e.,~the~total real values of positive ramps, negative ramps, ramps or no ramps in the database. Note that this way, the experiments are performed with the two different definitions of the ramp function ($S_t^1$ and $S_t^2$) given in Section \ref{Problem_definition}.

The following step to obtain the prediction of the WPREs is to train the considered ML regressors. A partition of the data into {training} (80\%), and {test} (20\%) sets is carried out. In the case of the SVR and MLP, a validation set from the training (5\%) set is also considered. This validation set is used to obtain the best SVR hyper-parameters $C$, $\epsilon$ and $\gamma$, by means of a {GS} \cite{tutorial98}. The validation set is also used in the training of the MLP approach, in order to {{prevent the NN from overtraining}. Both training and test sets have been randomly constructed from the available data after the cleaning pre-processing. {The concrete configurations and the values used for the parameters of the considered ML regression models, $\mathcal{M}$, are listed in Table \ref{ModelDevelop}.}

{With all these previous considerations in mind, Sections \ref{sec:Results} and \ref{sec:Discussion}, focus on showing the results obtained and on discussing them, respectively.}

\begin{table}[ht]
\centering{{
\caption{
\label{ModelDevelop}
Configuration and design parameters of the regression ML models $\mathcal{M}$ explored in the proposed approach for all the wind farms considered.}\vspace{0.3cm}
\centering\footnotesize
\begin{footnotesize}
\resizebox{16cm}{!} {
\begin{tabular}{ccc}
\toprule
\textbf{Model}
 \boldmath{$\mathcal{M}$} & {\textbf{Model Configuration}} & {\textbf{Values Used in the Design Parameters for Each Model} \boldmath{$\mathcal{M}$}} \\
\midrule
SVR & SVR with Gaussian kernel & \begin{tabular}[c]{@{}c@{}}$C=2^{c}$, $ c=-5 \cdots 12$; $\epsilon=2^{e}$, $e=-15 \cdots 0$;\\ $\gamma = (0.1-0.0001)/9 \cdot g+0.0001 $, $g=0 \cdots 9$ \end{tabular}\\
\midrule
 ELM & \begin{tabular}[c]{@{}c@{}} 3-layer NN with\\ sigmoid activation function \end{tabular} & \begin{tabular}[c]{@{}c@{}}Number of neurons in each of the\\ three layers (input-hidden-output): 48-150-1\end{tabular} \\
\midrule
 GP & RBF kernel & \begin{tabular}[c]{@{}c@{}} $\Lambda = \ln[(\max(x_i)-\min(x_i)]/2)$;\\ $\sigma_o^2=$ variance($S_{t_i}$); $\sigma^2=$ $\sigma_o^2$/4 \end{tabular} \\
\midrule
 MLP & Levenberg--Marquardt training & \begin{tabular}[c]{@{}c@{}} epoch $=1000$; gradient $ = 10^{-7}$ ;\\ $\mu = 10^{10}$; validation-checks $ = 6 $\end{tabular} \\
\bottomrule
\end{tabular}} \end{footnotesize} }
}
\end{table}

\subsection{Results}\label{sec:Results}
As mentioned in the description of the problem at hand, among the several definitions of ramp functions, $S_t$, the most common ones are considered \cite{Gallego15a}, stated, respectively, by~\mbox{Equations (\ref{Sdef1}) and (\ref{Sdef2})}, because both include power production criteria ($P_t$) at the wind farm. The~variation of power caused by a wind ramp, $P_{t+\Delta t_r}-P_t$, has been studied in the experiments below in the three wind farms (Figure \ref{fig:maps}) within a {time interval} $\Delta t_r = 6$ h, which is determined by the resolution of the reanalysis data.

In addition, in order to properly understand the analysis of the results obtained, it is convenient to point out that, by using the indicator function $I_{t}$ stated by Equation (\ref{threshold_S0}), the proposed methodology is able to successfully detect those WPREs that surpass the thresholds ($S_0$ or $-S_0$), when using the $S_t^1$ ramp function definition, or the single threshold ($S_0$), when using the $S_t^2$ definition. As will be shown later on, this~is due to the fact that, with the first ramp definition ($S_t^1$), it can be detected three types of events: ascending ramps (which are those whose power exceeds $S_0$), descending ramps (those surpassing $-S_0$) and the existence of ``no ramps'' (when the generated electric power is in between the two thresholds). Conversely, in the case of using the $S_t^2$ ramp function definition, it is only necessary to determine whether or not there is a ramp, so that only a threshold is necessary.

Taking these considerations into account and aiming at better explaining the results, the discussion is organized according to the objective function used, either $S_t^1$ or $S_t^2$, leading~to \mbox{Sections \ref{sec:S1_results} and \ref{sec:S2_results}}, respectively.

\subsubsection{Results using $S_t^1$ as the Ramp Function Definition }\label{sec:S1_results}
Table \ref{ResultsS1} shows the results obtained in this problem of WPRE prediction when considering $S_t^1$ as the objective function, in the three aforementioned wind farms in Spain (labeled ``A'', ``B'' and ``C'' in Figure \ref{fig:maps}). For each wind farm, the performance of any of the ML regressors explored (SVR, ELM, GP and MLP) has been measured using the metrics RMSE, MAE and sensitivity ($s$ ($+$ramp), $s$ ($-$ramp), \mbox{$s$ (no  ramp)}).

\begin{table}[ht]
\begin{center}
\caption{\label{ResultsS1} {Results (in terms of RMSE, MAE and sensitivity) corresponding to the estimation of the ramp function $S_t^1$ (Equation (\ref{Sdef1})) obtained when using the proposed approach, as a function of the ML regressors explored (SVR, ELM, GP and MLP), in the tree study cases: the wind farms ``A'', ``B'' and ``C'', whose locations have been represented in Figure \ref{fig:maps}}.}\vspace{0.3cm}
\begin{small}
\begin{tabular}{cccccc}
\toprule
\multicolumn{6}{c} { \textbf{Wind Farm A} }\\ 
\midrule
\textbf{ML regressor} &\textbf{RMSE} &\textbf{MAE} & \boldmath{$s$} \textbf{(+ramp)} & \boldmath{$s$} \textbf{(}\boldmath{$-$}\textbf{ramp)} & \boldmath{$s$} \textbf{(no ramp)} \\
 & \textbf{ (MW)} & \textbf{(MW)} &\textbf{(\%)} &\textbf{(\%)} &\textbf{(\%)}\\
\midrule
SVR & 7.0085 & 5.2673 & 26.93 & 24.20 & 96.59 \\
ELM & 5.6779 & 4.2499 & 40.54 & 42.59 & 95.51 \\
\textbf{GP} & \textbf{5.3066} & \textbf{3.9519} & \textbf{54.93} & \textbf{51.95} & \textbf{93.96} \\
MLP & 5.4538 & 4.0021 & 12.13 & 5.72 & 99.41 \\
\midrule
\multicolumn{6}{c} { \textbf{Wind Farm B} } \\
\midrule
 \textbf{ML regressor}  &\textbf{RMSE} &\textbf{MAE} & \boldmath{$s$} \textbf{(+ramp)} & \boldmath{$s$} \textbf{(}\boldmath{$-$}\textbf{ramp)} & \boldmath{$s$} \textbf{(no ramp)} \\
 & \textbf{(MW)} & \textbf{(MW)} &\textbf{(\%)} &\textbf{(\%)} &\textbf{(\%)} \\
\midrule
SVR & 8.0025 & 5.9773 & 35.53 & 34.10 & 86.66 \\
ELM & 7.4539 & 5.9768 & 32.93 & 33.14 & 92.13 \\
\textbf{GP} & \textbf{5.9856} & \textbf{4.4298} & \textbf{52.10} & \textbf{58.25} & \textbf{91.71}  \\
MLP & 5.9009 & 4.3429 & 15.11 & 13.14 & 97.25 \\
\midrule
\multicolumn{6}{c} { \textbf{Wind Farm C} } \\
\midrule
\textbf{ML regressor}  &\textbf{RMSE} &\textbf{MAE} & \boldmath{$s$} \textbf{(+ramp)} & \boldmath{$s$} \textbf{(}\boldmath{$-$}\textbf{ramp)} & \boldmath{$s$} \textbf{(no ramp)}\\
 & \textbf{(MW)} & \textbf{(MW)} &\textbf{(\%)} &\textbf{(\%)} &\textbf{(\%)} \\
\midrule
SVR & 7.1370 & 5.3406 & 45.38 & 44.20 & 91.33 \\
ELM & 5.8367 & 4.4462 & 50.32 & 47.64 & 94.01 \\
\textbf{GP} & \textbf{4.7515} & \textbf{3.4771} & \textbf{57.14} & \textbf{61.05} & \textbf{93.99} \\
MLP & 5.0727 & 3.6827 & 14.21 & 10.26 & 98.52 \\
\bottomrule
\end{tabular}
\end{small}
\end{center}
\end{table}

Regarding the reasons why the mentioned metrics are used to the detriment of others, it is convenient to stress some aspects related to what, in fact, are two conceptually distinct groups of measures: metrics that measure errors (RMSE and MAE), on the one hand, and metrics that quantify success prediction rates (sensitivity), on the other. These facets to be highlighted are:

\begin{itemize}
\item With respect to the ``conventional'' metrics that measure errors, there are two reason that have compelled us to include the RMSE and MAE metrics. The first one is that they are the most commonly used in the literature. Examples of relevant papers in which these metrics are used for WPRE forecasting are \cite{Gallego15a,Cutler07,Gallego11,gallego2013statistical}. Please see \cite{Gallego15a} for a useful discussion on this issue. The~second cause is, as will be shown, that the utility of these error measures can be complemented by using the sensitivity metric, the other class of metrics that are chosen.
\item The second couple of points that are important to be emphasized here are just those related to the aforementioned sensibility in Equation (\ref{eq:sensitivity}), one with respect to its meaning and the other regarding its application. On the one hand, the physical meaning of sensitivity is just the percentage of correct ramp predictions with respect to actual measured data. Despite its apparent simplicity, this is, however, an excellent measure of the extent to which the regressor algorithm under test is efficient in detecting wind ramps. On the other hand, regarding its application step in the proposed methodology, the key point is that sensitivity is only used after having predicted the ramp function with a regression technique and a threshold has been defined. After applying the threshold, the number of real WPREs is thus obtained and compared to the predicted number. This way, the fact that the problem is highly unbalanced is not an issue any longer; or, in other words, the regression techniques are applied to the ramp function, and then, a threshold to classify events is established. In this case, the percentage of correct WPRE identifications is obtained. Note that the work's objective is to deal with a regression problem, it is enough to show the good percentage of correct classification after the threshold setting in the predicted ramp function.
\end{itemize}

The analysis of Table \ref{ResultsS1} allows for elucidating some interesting conclusions:

\begin{enumerate}
\item {The performance of the ML regressors is, in general, good in terms of RMSE, MAE and sensitivity $s$, although, as shown, there are some ML regressors that work better than others}.
 \item {Regarding the performance of one regressor with respect to that of another, the results of Table \ref{ResultsS1} clearly indicate that the GP model reaches the best results of all the regressors tested, with an excellent reconstruction of the ramp function $S_t^1$ from the ERA-Interim variables. Note in Table \ref{ResultsS1} that the values of the metrics obtained by the GP regressor are marked in bold. Its RMSE and MAE values are much lower (better) than those of the other ML regressors explored. In terms of sensitivity, its performance is even better. Specifically, its sensitivity $s$ (or percentage of correct predictions (with respect to the real, measured data) stated by Equation (\ref{eq:sensitivity})) is much higher (better) than those of the other regressors: $s$ (+ramp)$_{\mathrm{GP}} \gg $ $s$ (+ramp)$_{\mathrm{others}} $ (for ascending ramps) and $s$ ($-$ramp)$_{\mathrm{GP}} \gg $ $s$ ($-$ramp)$_{\mathrm{others}}$ (for descending ramps). This confirms the validity of the results measured with the error metrics and proves the feasibility of the proposed methodology for predicting wind ramps, both ascending and descending ramps.}
\item The worst result corresponds to the MLP, with a poorer detection of positive WPREs, when~compared to the other ML regressors.
\item The SVR and ELM work well in between both GP and MLP, with acceptable values of detection in positive WPREs.
\end{enumerate}

With this analysis in mind, Figures \ref{temp_mnon}--\ref{temp_jara_gpr} show the estimation of $S_t^1$ obtained by the GP and ELM algorithms (the two best approaches tested in the experiments), when using $S_t^1$ as the objective function, for the wind farms A, B and C, respectively. Some aspects to correctly interpret these figures~are:

\begin{itemize}
\item Aiming at clearly showing the algorithms' performance, only the 300 first samples of the test set have been represented in these figures.
\item Furthermore, a threshold value $S_0$ (and the corresponding $-S_0$) has been marked in these figures, so it can be used to decide whether or not the event is a ramp power event (see Equation \eqref{threshold_S0}). When a ramp occurs, it is possible to decide whether the ramp event is ascending or descending.
\end{itemize}

\begin{figure}[H]
\begin{center}
\includegraphics[draft=false,angle=0,width=13cm]{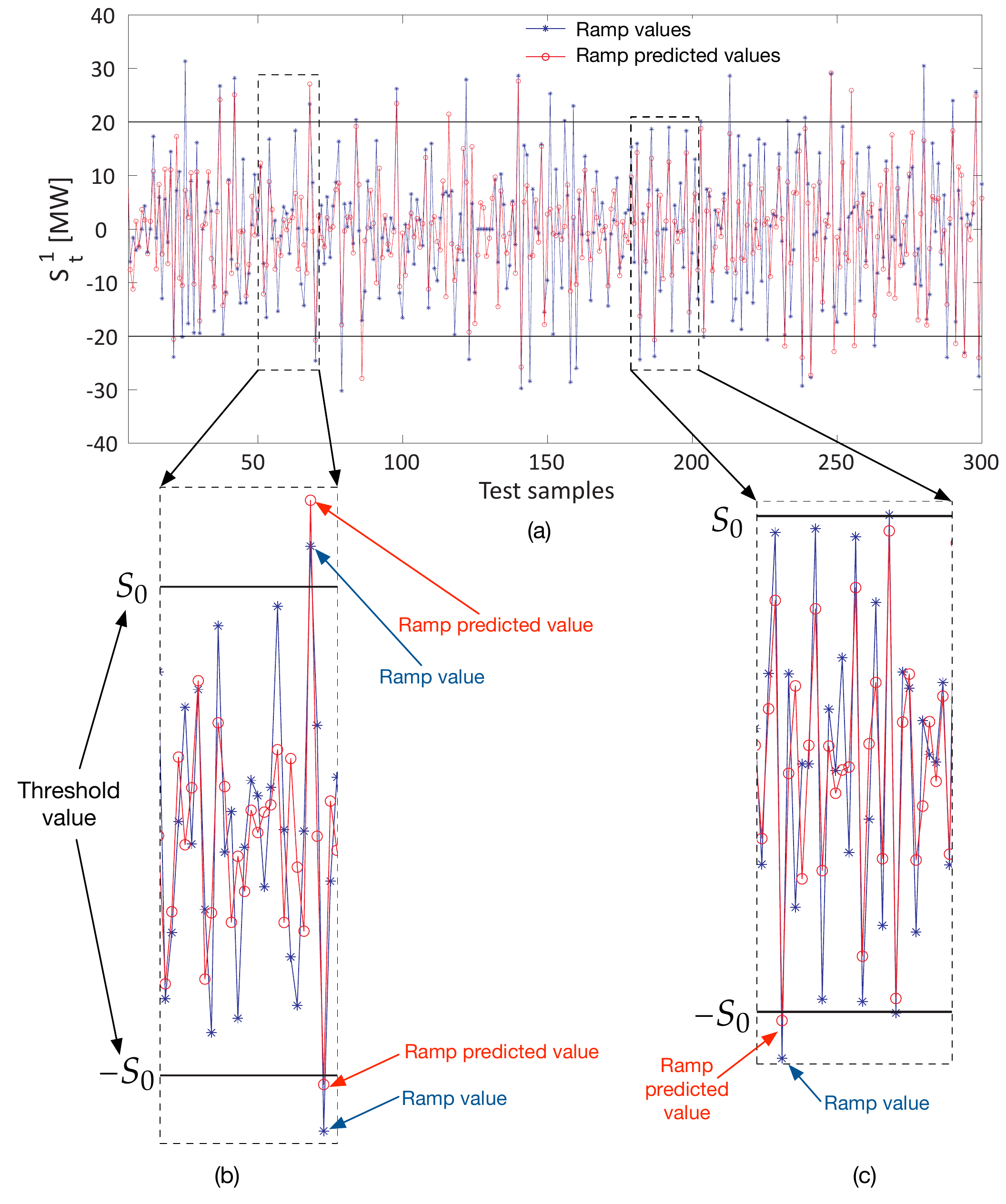}
\vspace{-12pt}
\caption{\label{temp_mnon} {(\textbf{a}) Estimation of the ramp function $S_t^1$  (Equation  (\ref{Sdef1}))  obtained by using the proposed approach in the particular case in which the ML is an ELM regressor. This figure corresponds to Wind Farm A, whose location has been represented in Figure \ref{fig:maps}. (\textbf{b},\textbf{c}) represent two shorter excerpts in which the predicted WPREs that exceed the thresholds ($S_0$ or $-S_0$) are shown to be correctly detected. A WPRE is detected if $S_t^1 > 0.5 S_0$. The predicted series exhibits RMSE $ \approx 5.68$ MW, \mbox{MAE $ \approx 4.25$ MW}, \mbox{$s$ (+ramp) $= 40.54\%$}, $s$ ($-$ramp) $= 42.59\%$ and $s$ (no ramp) $= 95.51\%$.}}
\end{center}
\end{figure}

The results illustrated in Figure \ref{temp_mnon} (a) show two data series: the series of real measured WPRE (red $\circ$) and the series of {predicted} WPRE (blue $\ast$) values computed by using the proposed hybrid methodology. In the effort to better explain the results and the applicability of this proposal, Figure~\ref{temp_mnon} is drawn in a more detailed way than the others, zooming into two shorter time excerpts, b and c. The insets b and c show how there are some WPREs that surpass any of the thresholds $S_0$ and $-S_0$. Specifically, and as mentioned before, a WPRE is detected in this approach if the ramp function is larger than 50\% of $S_0$. Note that Figure \ref{temp_mnon}b,c show how the predicted WPREs (blue $\ast$) exceeding any thresholds ($S_0$ or $-S_0$) are correctly predicted when compared to the real, measured WPRE (red $\circ$).

Regarding such a threshold value, it is worth mentioning that $S_0$ is not used until the very end of the experiments, once the ramp function has been predicted with the ML regression algorithms. In this respect, it is also convenient to remark that, in the proposed approach, it does not look to optimize $S_0$. Only $S_0$ is displayed as an indication (example) that the ML regression model $\mathcal{M}$ applied can be turned into a classification for WPRE. Note, however, that the purpose of this study is to deal with it as a regression problem.

The good performance observed in Figure \ref{temp_mnon} for the ELM is common (and even better) to those illustrated in Figures \ref{temp_pena_gpr} and \ref{temp_jara_gpr}.

\begin{figure}[ht]
\begin{center}
\includegraphics[draft=false,angle=0,width=13cm]{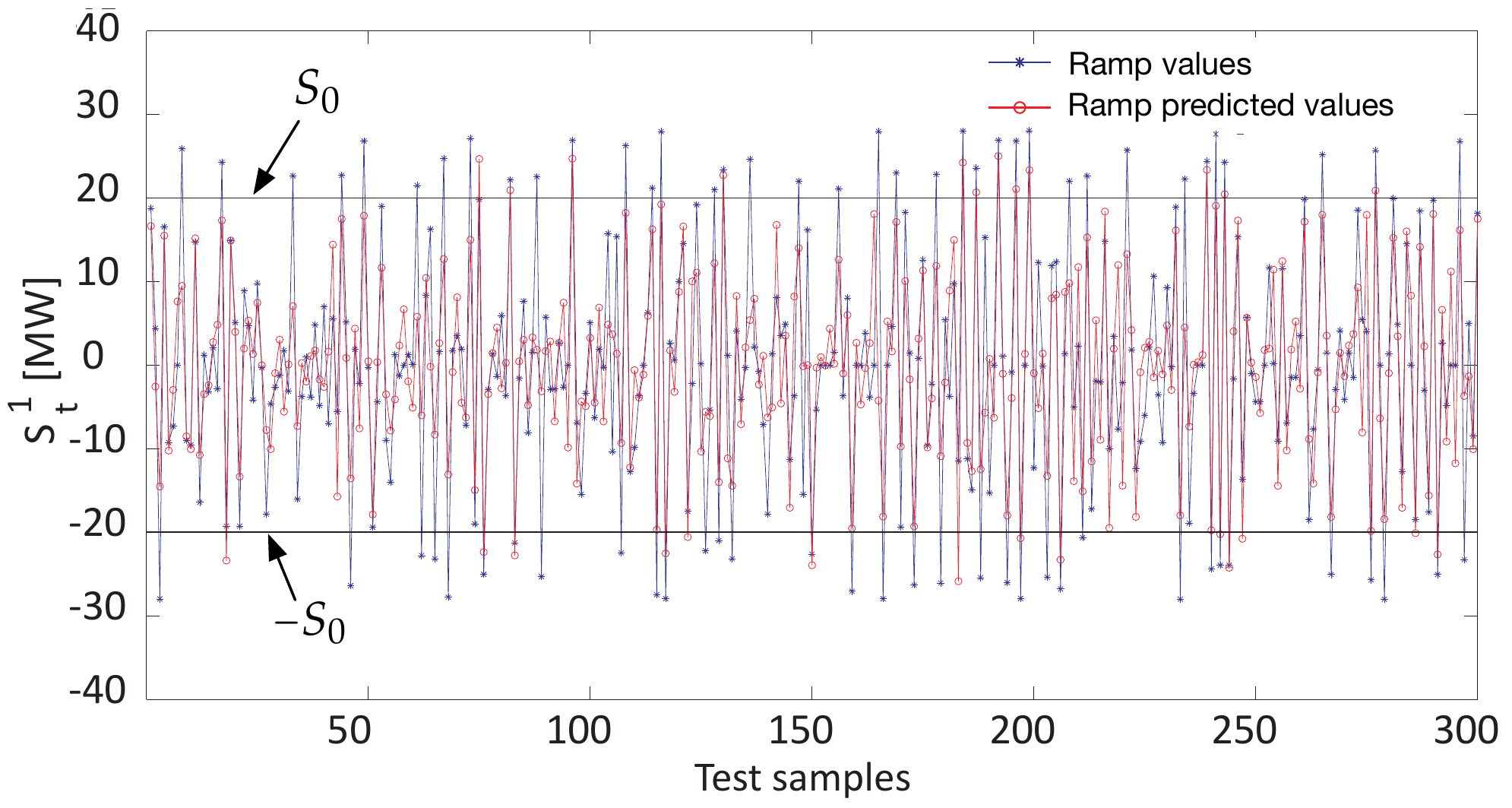}
\vspace{-12pt}
\caption{\label{temp_pena_gpr} {Estimation of the ramp function $S_t^1$  (Equation (\ref{Sdef1}))  obtained by this proposed hybrid approach when using the GP as the ML regressor in Wind Farm B. The predicted series exhibits \mbox{RMSE $ \approx 5.98$ MW}, MAE $ \approx 4.43$ MW, $s$ (+ramp) $= 52.10\%$, $s$ ($-$ramp) $= 58.25\%$ and $s$ (no ramp) $= 91.71\%$ (\mbox{see Table \ref{ResultsS1}})}.}
\end{center}
\end{figure}

\begin{figure}[ht]
\begin{center}
\includegraphics[draft=false,angle=0,width=13cm]{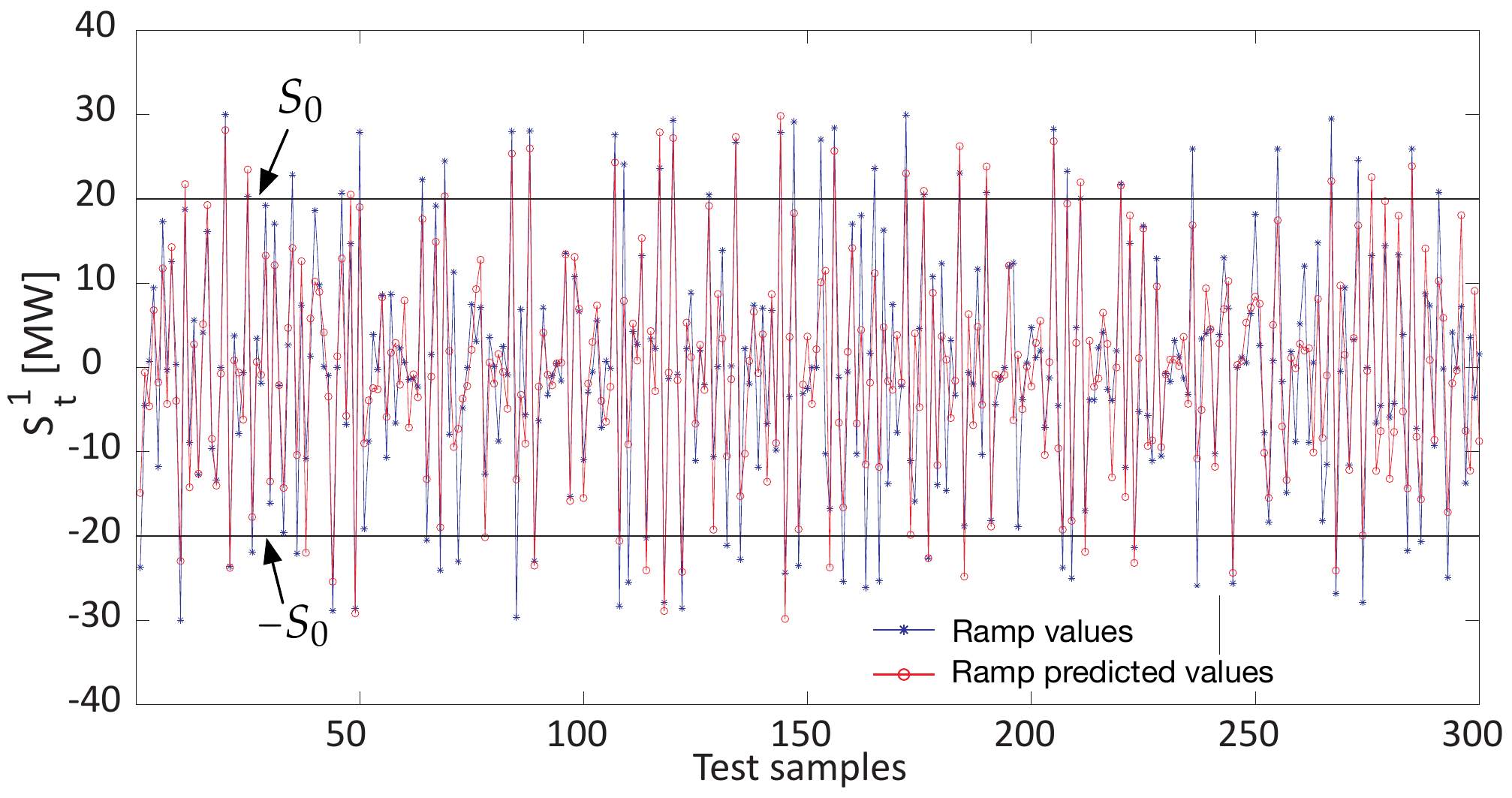}
\vspace{-12pt}
\caption{\label{temp_jara_gpr} {Prediction of the ramp function $S_t^1$  (Equation (\ref{Sdef1}))  when using the GP in \mbox{Wind Farm C}. The predicted ramps series exhibits RMSE $ \approx 4.75$ MW, MAE $ \approx 3.48$ MW, $s$ (+ramp) $= 57.14\%$, \mbox{$s$ ($-$ramp) $= 61.05\%$}, and $s$ (no ramp) $= 93.99\%$ (see Table \ref{ResultsS1})}.}
\end{center}
\end{figure}

The joint analysis of both Figures \ref{temp_mnon} and \ref{temp_jara_gpr} and Table \ref{ResultsS1} reveals the suitable throughput of the ML regression techniques (mainly the GP model), which hybridized with the ERA-Interim predictive values, assist in obtaining a robust decision system in terms of the existence or not of a power ramp, depending, of course, on the definition of the threshold $S_0$.

\subsubsection{Results using $S_t^2$ as the Ramp Function Definition }\label{sec:S2_results}
On the other hand, Table \ref{ResultsS2} and Figures \ref{temp_mnon_gpr}--\ref{temp_jara} will assist us to explain the results when $S_t^2$ is the ramp function to be predicted.

\begin{table}[ht]
\begin{center}
\caption{\label{ResultsS2} Results (in terms of RMSE, MAE and sensitivity) corresponding to the estimation of the ramp function $S_t^2$  (Equation (\ref{Sdef2}))  obtained by the proposed approach as a function of the ML regressors explored (SVR, ELM, GP, and MLP), for Wind Farms ``A'', ``B'' and ``C'', respectively.} \vspace{0.3cm}
\scalebox{0.95}[0.95]{
\begin{small}
\begin{tabular}{ccccc}
\toprule
\multicolumn{5}{c} { \textbf{Wind Farm A}} \\
\midrule
\textbf{ML regressor}  &\textbf{RMSE} &\textbf{MAE} & \boldmath{$s$} \textbf{(ramp)} & \boldmath{$s$} \textbf{(no ramp)} \\
 & \textbf{(MW)} & \textbf{(MW)} &\textbf{(\%)} &\textbf{(\%)} \\
\midrule
SVR & 6.8847 & 5.1876 & 31.33 & 96.27 \\
ELM & 5.7037 & 4.2925 & 41.99 & 95.01 \\
\textbf{GP} & \textbf{5.2048} & \textbf{3.7897} & \textbf{49.66} & \textbf{96.36} \\
MLP & 5.4351 & 3.9861 & \textbf{8.71} & 99.44 \\
\midrule
\multicolumn{5}{c} { \textbf{Wind Farm B}} \\
\midrule
\textbf{ML regressor}  &\textbf{RMSE} &\textbf{MAE} & \boldmath{$s$} \textbf{(ramp)} & \boldmath{$s$} \textbf{(no ramp)} \\
 & \textbf{(MW)} & \textbf{(MW)} &\textbf{(\%)} &\textbf{(\%)} \\
\midrule
SVR & 7.9439 & 5.8853 & 44.16 & 85.55 \\
ELM & 7.3148 & 5.8675 & 34.67 & 93.23 \\
\textbf{GP} & \textbf{5.9223} & \textbf{4.4037} & \textbf{65.32} & \textbf{84.12} \\
MLP & 5.9051 & 4.3475 & \textbf{14.76} & 97.17 \\
\midrule
\multicolumn{5}{c} { \textbf{Wind Farm C} }\\
\midrule
\textbf{ML regressor}  &\textbf{RMSE} &\textbf{MAE} & \boldmath{$s$} \textbf{(ramp)} & \boldmath{$s$} \textbf{(no ramp)} \\
 & \textbf{(MW)} & \textbf{(MW)} &\textbf{(\%)} &\textbf{(\%)} \\
\midrule
SVR & 7.1525 & 5.4677 & 37.88 & 93.83 \\
ELM & 5.8624 & 4.4368 & 58.16 & 92.10 \\
\textbf{GP} & \textbf{5.1030} & \textbf{3.6991} & \textbf{57.26} & \textbf{94.42} \\
MLP & 5.0605 & 3.6670 & \textbf{11.22} & 98.56  \\
\bottomrule
\end{tabular}
\end{small}
}
\end{center}
\end{table}

\begin{figure}[ht]
\begin{center}
\includegraphics[draft=false,angle=0,width=13cm]{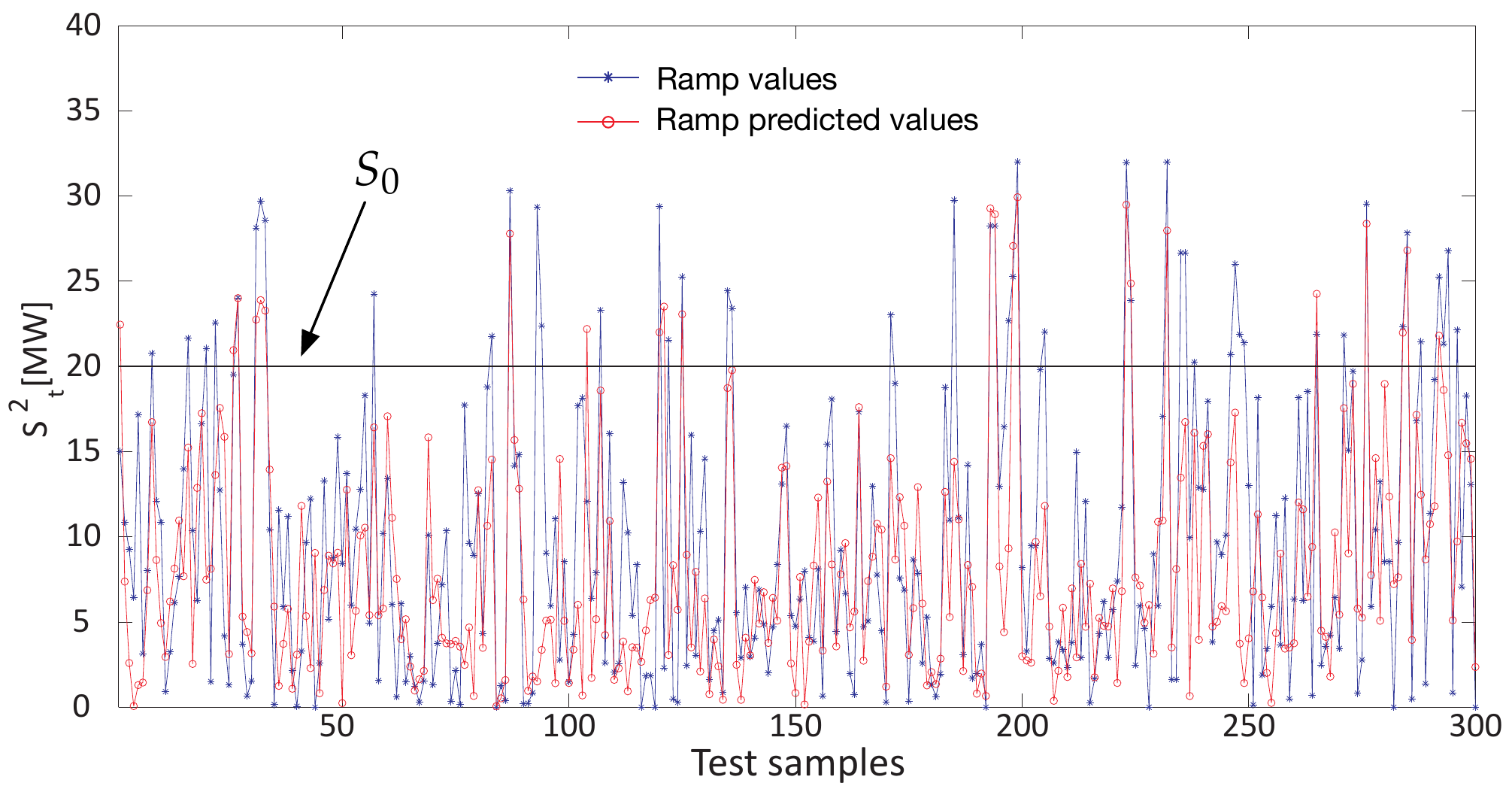}
\vspace{-12pt}
\caption{\label{temp_mnon_gpr} {Estimation of the ramp function $S_t^2$  (Equation  (\ref{Sdef2}))  obtained by the proposed approach using the GP regressor, in Wind Farm A. The ramp predicted values resemble the ramp measured ones with RMSE $ \approx 5.20$ MW and MAE $ \approx 3.79$ MW, $s$ (ramp) $=49.66\%$ and $s$ (no ramp) $= 96.36\%$ (\mbox{see Table \ref{ResultsS2}})}.}
\end{center}
\end{figure}

\begin{figure}[ht]
\begin{center}
\includegraphics[draft=false,angle=0,width=13cm]{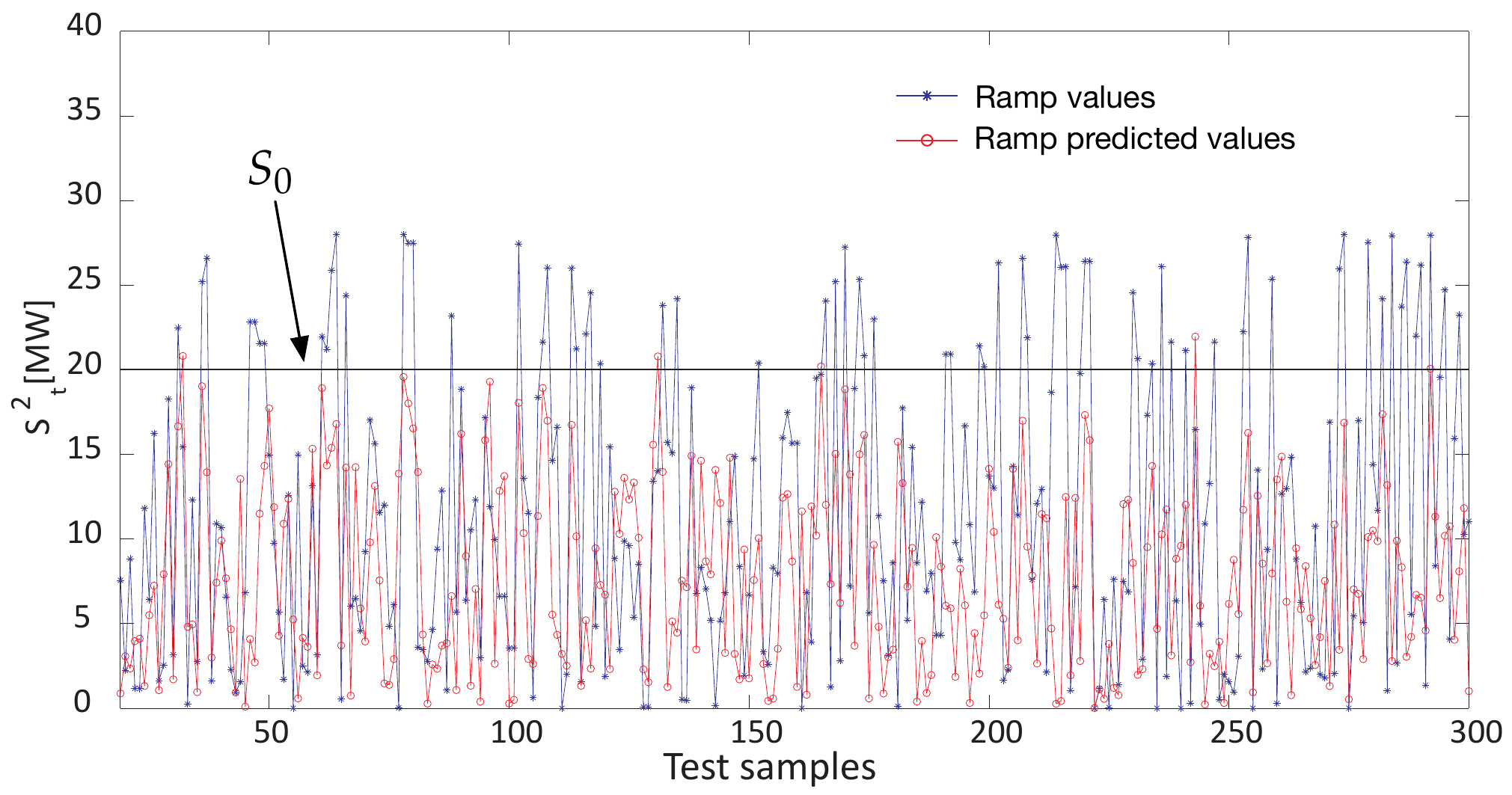}
\vspace{-12pt}
\caption{\label{temp_pena} {Estimation of the ramp function $S_t^2$ (Expression (\ref{Sdef2})) obtained by the proposed method when using the ELM regressor, in Wind Farm B. The predicted series follows the measured series with RMSE $ \approx 5.90$ MW and MAE $ \approx 4.40$ MW, $s$ (ramp) $=65.32\%$ and $s$ (no ramp) $= 84.12\%$ (see Table \ref{ResultsS2})}.}
\end{center}
\end{figure}

\begin{figure}[ht]
\begin{center}
\includegraphics[draft=false,angle=0,width=13cm]{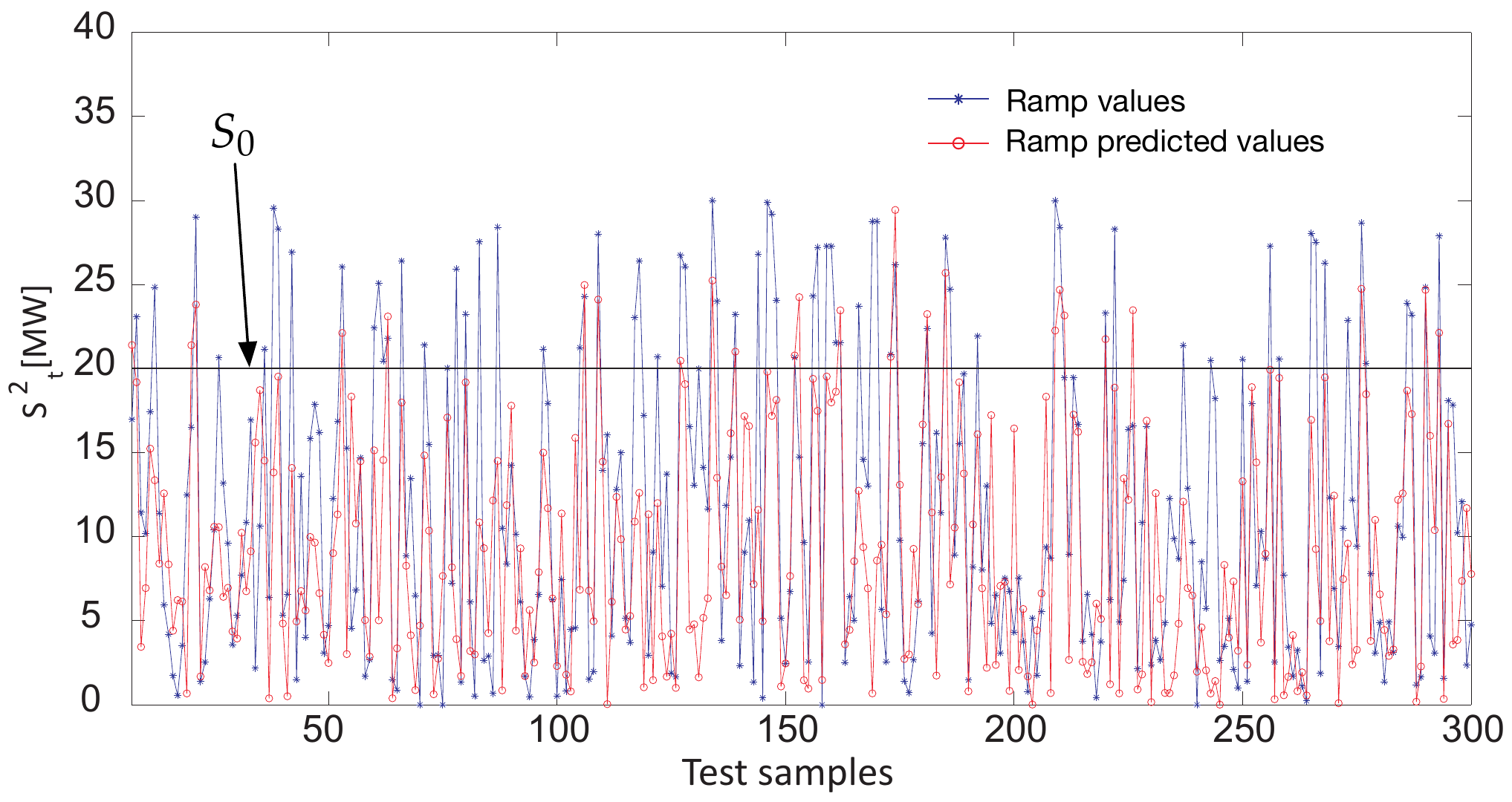}
\vspace{-12pt}
\caption{\label{temp_jara} {Estimation of the ramp function $S_t^2$ (Expression (\ref{Sdef2})) obtained by the proposed method when using the ELM regressor, in Wind Farm C. The predicted series follows the measured series with RMSE $ \approx 5.86$ MW and MAE $ \approx 4.43$ MW, $s$ (ramp) $=58.16\%$ and $s$ (no ramp) $= 92.10\%$ (see Table \ref{ResultsS2})}.}
\end{center}
\end{figure}

Table \ref{ResultsS2} presents the results (in terms of RMSE, MAE and sensitivity) corresponding to the estimation of the ramp function $S_t^2$ (Expression (\ref{Sdef2})) achieved by using the proposed approach as a function of the ML regressors explored (SVR, ELM, GP and MLP).

A first aspect that stands out of Table \ref{ResultsS2} is that it has fewer columns related to sensitivity than those of Table \ref{ResultsS1}. This is an interesting points that arises from the different definitions of the ramp function $S_t^2$, either $S_t^1$ or $S_t^2$. Note that, for definition $S_t^2$, the sensitivity is the percentage of correctly predicted results (either ramp or no ramp) with respect to the actual measured data. This is the reason why $s$ has only two columns in Table \ref{ResultsS2}, $s$ (ramp) and $s$ (no ramp), whereas Table \ref{ResultsS1} exhibits three $s$-related columns. This is because, in the case of the $S_t^1$ ramp definition, there are three events to be detected: ascending ramp ($+$), descending ramp ($-$) and no ramps.

In the same way as Table \ref{ResultsS1}, Table \ref{ResultsS2} also reveals that, for $S_t = S_t^2$, the GP approach exhibits the best results, outperforming clearly the rest of the ML regressors tested, except the MLP. This~has similar values {only} in its error metric, RMSE and MAE, but {not} in its $s$ (ramp) value, which is considerably worse than that of the GP. This is clear, for instance, in Wind Farm A, in which RMSE$_{\mathrm{GP}} \approx 5.20$~MW, less~than that of the other regressors. Note that $s$(ramp)$_{\mathrm{GP}} = 49.66 \gg $ $s$ (ramp)$_{\mathrm{MLP}} = 8.71$. \mbox{In Wind Farm B}, the~performance of the GP (RMSE$_{\mathrm{GP}} \approx 5.90$ MW) is similar to that of the MLP and much better than that of SVR (RMSE$_{\mathrm{SVR}} \approx 7.94 $ MW) and SVR (RMSE$_{\mathrm{SVR}} \approx 7.32 $ MW). Note again that, although the GP model is similar to the MLP in error metrics, however, the GP exhibits much better sensitivity than the MLP, $s$ (ramp)$_{\mathrm{GP}} \gg $ $s$ (ramp)$_{\mathrm{MLP}} $. This is true not only for the MLP (which has similar errors), but also for the rest of the ML, which are long surpassed by the GP model in the aim of detecting wind ramps. For clarity, this is marked in bold in Table \ref{ResultsS2}. This means that the GP is more efficient in predicting wind ramps (the very core of this approach) than the others, and this is the reason why the sensitivity helps supplement the information provided by the error metrics.

Once the results shown in Table \ref{ResultsS2} have already been analyzed, it is convenient to have a look at its associated figures showing the data series, which involve both the estimated (predicted) and the measured values of the ramp function $S_t^2$. Regarding this, Figures \ref{temp_mnon_gpr}--\ref{temp_jara} show the estimation of $S_t^2$ obtained by the GP (in Wind Farm A) and ELM algorithms, for the wind farms B and C, respectively.

In Figures \ref{temp_mnon_gpr}--\ref{temp_jara} a threshold value $S_0$ to mark the presence (or not) of a WPRE is also represented. As in the first objective function, the good performance of the ML regressors allows a significant detection of WPRE in wind farms.

\subsection{Discussion}\label{sec:Discussion}
The results obtained show that the proposed hybrid WPREs prediction approach---which combines data from numerical-physical models (reanalysis) with state-of-the-art statistical ML approaches (regressors)---is a feasible option to tackle this problem in wind farms. Regarding the proposed fusion of reanalysis data and ML regressors, the results have pointed out that:

\begin{itemize}
\item The use of reanalysis data as predictive variables for WPRE forecast has the following beneficial~properties:

\begin{enumerate}
\item Reanalysis makes the training of the ML regressors easier if there are enough measures of the objective variables. This is just the case in this approach because reanalysis data provide robust meteorological variable estimation back to 1979 in the case of the ERA-Interim reanalysis, with high spatial and enough temporal resolution to tackle this problem.
\item The variables from reanalysis projects are similar to those by any weather numerical forecast system, even meso-scale ones, so it is straightforward to tackle the WPRE prediction by using alternative models, such as the well-known Weather Research and Forecasting (WRF) meso-scale model \cite{Skaramrock05}, to predict future values of the predictive variables and, then, the corresponding WPRE prediction for a given wind farm.
\item The use of reanalysis data allows the repeatability of the described experiments by other researchers since such data are freely available on the Internet.
\end{enumerate}
\item The performance studies of the state-of-the-art ML regressors, the other pillar this approach is based on, have shown that the GP reaches the best results in both definitions of the wind power ramp function considered:
\begin{enumerate}
\item When using the $S_t^1$ definition, the results clearly show that the GP model achieves the best results of all the regressors tested, with an accurate reconstruction of the ramp function from the ERA-Interim variables. Its RMSE and MAE vales are much lower than those of the other ML regressors explored. Furthermore, its sensitivity $s$---or percentage of correct predictions (with respect to the real, measured data)---is much higher than those provided by the other regressors: $s$ (+ramp)$_{\mathrm{GP}} \gg $ $s$ (+ramp)$_{\mathrm{others}} $ (for ascending ramps) and $s$ ($-$ramp)$_{\mathrm{GP}} \gg $ $s$($-$ramp)$_{\mathrm{others}}$ (for descending ramps). This demonstrates the feasibility of the proposed methodology for predicting wind ramps, both ascending and descending ones.
\item Similarly, when using the $S_t^2$ ramp definition, the GP approach also exhibits the best results, outperforming clearly the rest of the ML regressors tested, except the MLP, which has similar values {only} in its error metric, RMSE and MAE, but {not} in its $s$(ramp) value, which~is considerably worse than that of the GP.
These sensitivity results point out that the GP is more efficient in predicting wind ramps (the very core of this approach) than the other regressors, this being the reason why the sensitivity metric helps complement the information provided by the error measures.
\end{enumerate}
\end{itemize}

Finally, the results show how the proposed approach allows the use of threshold values to detect whether or not a wind power ramp occurs. The method is also flexible enough to choose a ramp function definition in the aim of considering a multi-class problem. Although in the experiments carried out, the multi-class problem contains three classes (ascending, descending or not ramp, in the $S_t^1$ definition), more classes could be defined. The optimal selection of the threshold values is an open question in the literature that has not been considered in this case.

\section{Conclusions}\label{Conclusions}
In this work, the feasibility of a novel hybrid approach that---by combining data from numerical-physical models (reanalysis) and state-of-the-art statistical ML regressors---aims at predicting WPREs has been explored. The accurate prediction of WPREs---caused by large fluctuations of wind power in a short time interval lead---is of practical interest not only for utility companies and independent system operators (in the effort of efficiently integrating wind energy without affecting power grid stability), but also for wind power farm owners (to reduce damage in turbines).

Specifically, several state-of-the-art statistical ML regressors---ranging from a MLP neural network to an ELM, a GP Regression or a SVR algorithm---have been applied to solve this problem in three different wind farms in Spain.

This has been the first contribution of this proposal since the use of regressors has {not} been previously applied directly to this WPRE prediction problem. The second contribution has been the use of direct reanalysis data as input (predictive) variables of the ML regression techniques. In~this regard, the use of data from the ERA-Interim reanalysis are proposed because it ensures a high resolution of the inputs, both spatial (grid of 0.125 $\times$ 0.125 at global level) and temporal (6-h time horizon). {Two~other reasons why reanalysis is used are: (a) the use of reanalysis data allows the repeatability of the experiments by other researchers since such data are available on the Internet; (b) the variables from reanalysis are similar to those from weather numerical forecast systems, even~mesoscale ones, so~that it would be straightforward to tackle the WPRE prediction problem by using other alternative models. Note however that it would be possible to adapt the proposed regression techniques to operate with alternative data not coming from numerical methods (or reanalysis), but other types of input~variables.

This purpose has been modeling the wind ramp function as accurately as possible in terms of several input variables. This way of tackling the problem overcomes some problems associated with the {WPRE} defined as a binary classification task \cite{Dorado17,Cornejo17}, or even ordinal classification \cite{dorado2017combining}, such as the appearance of highly imbalanced problems.

Two different definitions of the ramp function have been considered, those that are used the most in the literature. The experimental work has been carried out using data corresponding to three wind farms, located in different zones of Spain and having different atmospheric conditions, in the effort to obtain results as generalizable as possible. The experimental work carried out basically points out that:

\begin{enumerate}
\item The results show a good performance of the explored ML regression techniques hybridized with the ERA-Interim reanalysis data, especially those corresponding to the ELM and the GP ML regressors. In particular, the GP has been found to exhibit the best results, outperforming clearly the rest of the ML regressors tested. This has been shown especially evident in terms of its sensitivity (or percentage of correct predictions (with respect to the real, measured data)), which is much higher than those provided by the other regressors, showing the feasibility of the proposed methodology for predicting WPREs.
\item The experimental work has also revealed that the use of reanalysis data as predictive variables for WPRE forecast is beneficial: reanalysis has been found to make the training of the ML regressors easier since the ERA-Interim reanalysis provides robust meteorological variable estimation back to 1979, with high spatial and enough temporal resolution to tackle this problem.
\end{enumerate}

As a general conclusion, the results achieved by the proposed approach show that the hybrid method proposed is a feasible alternative to deal with the important problems that WPREs can cause in both the management of wind farms and in the balanced operation of power grids.

\part{Proposed contributions with numerical results in facilities management}\label{part:aplicaciones2}
\chapter{Accurate estimation of $H_s$ with SVR and marine radar images}\label{cap:ocean2}
\section{Introduction}
The availability and accuracy of wave data play a crucial role in the better understanding of numerical \cite{Wamdig1988,Tolman2009} and statistical wave models \cite{Durrant2013,Casas2014}, wave forecasting for safe ship navigation, design and operation of WECs \cite{Lopez2013}, and the design of vessels and marine structures: oil platforms, breakwaters \cite{Comola2014,Kim2014}, wave overtopping volumes \cite{Norgaard2014}, ports and harbours, etc. Thus, the topic has a clear impact on human safety, economics and clean energy production. One of the most important parameters to define the severity of a given ocean wave field is the $H_s$. $H_s$ is usually estimated using in-situ sensors, such as buoys, recording time series of wave elevation information. Buoys provide reliable sea state information that characterizes wave field in a fixed position (i.e. the mooring point). In addition, as buoys are anchored in a hostile media (the ocean), the probability that measuring problems (and therefore missing data) occur in situations of severe weather is very high \cite{Rao05}.

Complementary to the punctual information that buoys' measurements represent, an alternative way to estimate $H_s$ (and therefore an useful tool to reconstruct missing data from ocean buoys) consists of using remote sensing imaging methods, such as air and space borne \ac{SAR} images \cite{Alpers82}, on- and off-shore coherent radars \cite{Plant08,Nwogu10,Seemann13} or conventional X-band marine radars \cite{Hessner01,Reichert05,Izquierdo05}, which are broadly installed in every moving ship, and off- and on-shore platforms.

The analysis of the marine radar images of the sea surface is capable of estimating wave field and surface current information in real time for oceanographic monitoring purposes \cite{Young85,Nieto-Borge00,Senet01,Reichert05,Izquierdo05,Chen12}. Radar images of the ocean surface are produced by the backscattering phenomenon of the electromagnetic waves due to the roughness of the sea surface \cite{Alpers82,Plant08}. These radar images are then analyzed to obtain estimations of wave spectra in different spectral domains \cite{Reichert05,Izquierdo05}, which allow calculating typical sea state parameters, such as characteristic wave periods, wave lengths, wave propagation directions, etc. \cite{Hessner01,Hessner14}. Estimating $H_s$ from the wave spectrum derived from the X-band marine radar analysis is not straightforward, since the physics of the imaging mechanisms has complex dependencies on environmental conditions, included both wave conditions and other environmental factors such as wind. The wave spectral estimations derived from the radar images are not properly scaled in the sense that their integral cannot provide values of the standard deviation of the wave elevation field, and therefore, a direct estimation of $H_s$ is not possible.

Some approaches to estimate $H_s$ from marine radars take into account the geometrical shadowing effect of the lower waves by the higher waves to the radar antenna illumination \cite{Buckleyetal94,Buckleyetal98,SalcedoCarro15}. An alternative approach to estimate $H_s$ from X-band marine radar considers that $H_s$ depends linearly with the squared root of the signal-to-noise ratio $SNR$, where the signal is the spectral energy of the un-scaled wave spectrum, and the noise is related to the spectral energy of the speckle noise within the radar image \cite{Nieto-Borge08}.
This technique is an extension of the methodology initially proposed by \cite{Alpers82} to derive $H_s$ from SAR images of the sea surface. The $SNR$-based method is more robust, from the operational point of view, than the shadowing-based method and it is widely used for the standard applications of wave monitoring activities using conventional X-band marine radars \cite{Hessner01,Chen12}. Thus, the $SNR$-based method is used as an standard technique for $H_s$ estimation. Note that the $SNR$-based method needs a calibration campaign with an in-situ sensor, such as a buoy, to calibrate the marine radar. This calibration is not necessary in the method that analyzes the shadowing effect \cite{SalcedoCarro15}.
Although the $SNR$-based method to estimate $H_s$ is used all over the world, there are some limitations where this technique does not provide reliable values for $H_s$, giving some indications that the $H_s$ estimation depends on more parameters than only $SNR$ \cite{Vicen12}.

In this work an extension to the $SNR$-based method is proposed. This proposed extension uses SVR to estimate $H_s$. The method takes into account additional sea state parameters than only $SNR$. All those parameters are derived from the standard analysis of wave fields by using X-band marine radars. The work analyzes the relevant sea state parameters to estimate $H_s$ and compare the obtained results with the results derived from the \ac{SM}, based only on the estimation of $SNR$. For that purpose, a set of marine radar data in combination with $H_s$ values measured by buoys have been used. The data were recorded in three different geographical locations under different oceanographic conditions: the German basin and the Norwegian sector, both in the North Sea, and the Sable Field in South Africa.

The rest of the chapter is structured as follows: Section~\ref{sc:X-Radar} deals with the basics of the wave field analysis by using X-band marine radar data sets, including the $H_s$ estimation by the SM, and its limitations.
Section~\ref{sc:UsedData} describes the geographical locations and the oceanographic conditions of the X-band radar and buoy data used in this work. Section~\ref{Experiments} shows the achieved results after applying the SVR algorithms to the used data. Finally, Section~\ref{sc:Conclusions} summarizes the conclusions of the work.

\section{Analysis of the sea surface from X-band radar}\label{sc:X-Radar}
As mentioned before, the analysis of wave fields from X-band marine radars is based on the acquisition of consecutive radar images of the sea surface. Hence, the data sets are time series of radar images where the spatio-temporal $(x, y, t)$ evolution of the sea surface can be analyzed. From these data sets, applying a three-dimensional Fourier decomposition the so-called image spectrum $\mathcal{I}({\bf k}, \omega)$ is obtained,
where ${\bf k} = (k_x, k_y)$ is the wave number vector and $\omega$ is the angular frequency.
In practice, $\mathcal{I}({\bf k}, \omega)$ is estimated by using a three-dimensional FFT-based algorithm, therefore the $({\bf k}, \omega)$ values are defined in a discrete domain, where the sampling wave numbers $(\Delta k_x, \Delta k_y)$ depend on the spatial size of the radar images and their spatial resolutions given by the range and azimuthal resolutions of the radar system. The angular frequency resolution $\Delta \omega$ depends on the number of images in the radar image time series and its sampling time (i.e. the radar antenna rotation period). Hence, the spectral components are located within the spectra domain $\Omega_{{\bf k}, \omega}$ defined as

\begin{equation}\label{eq:3DSpectralDomain}
\Omega_{{\bf k}, \omega} \myeq [-k_{x_c}, \, k_{x_c}) \times [-k_{y_c}, \, k_{y_c}) \times [0, \, \omega_c]
\mbox{,}
\end{equation}

where $k_{x_c}$, $k_{y_c}$, and $\omega_c$ are the respective Nyquist limits in wave numbers and angular frequency given by the spatio-temporal resolution of the radar image time series. For the estimation of $H_s$, the relevant spectral components $({\bf k}, \omega) \in \Omega_{{\bf k}, \omega}$ of the three-dimensional image spectrum $\mathcal{I}({\bf k}, \omega)$ are classified in the following contributions (see the example illustrated in Figure~\ref{fig:Fimag3D_kwPlus}):

\begin{figure}[ht]	
  \begin{center}
    \includegraphics[width=0.80\textwidth]{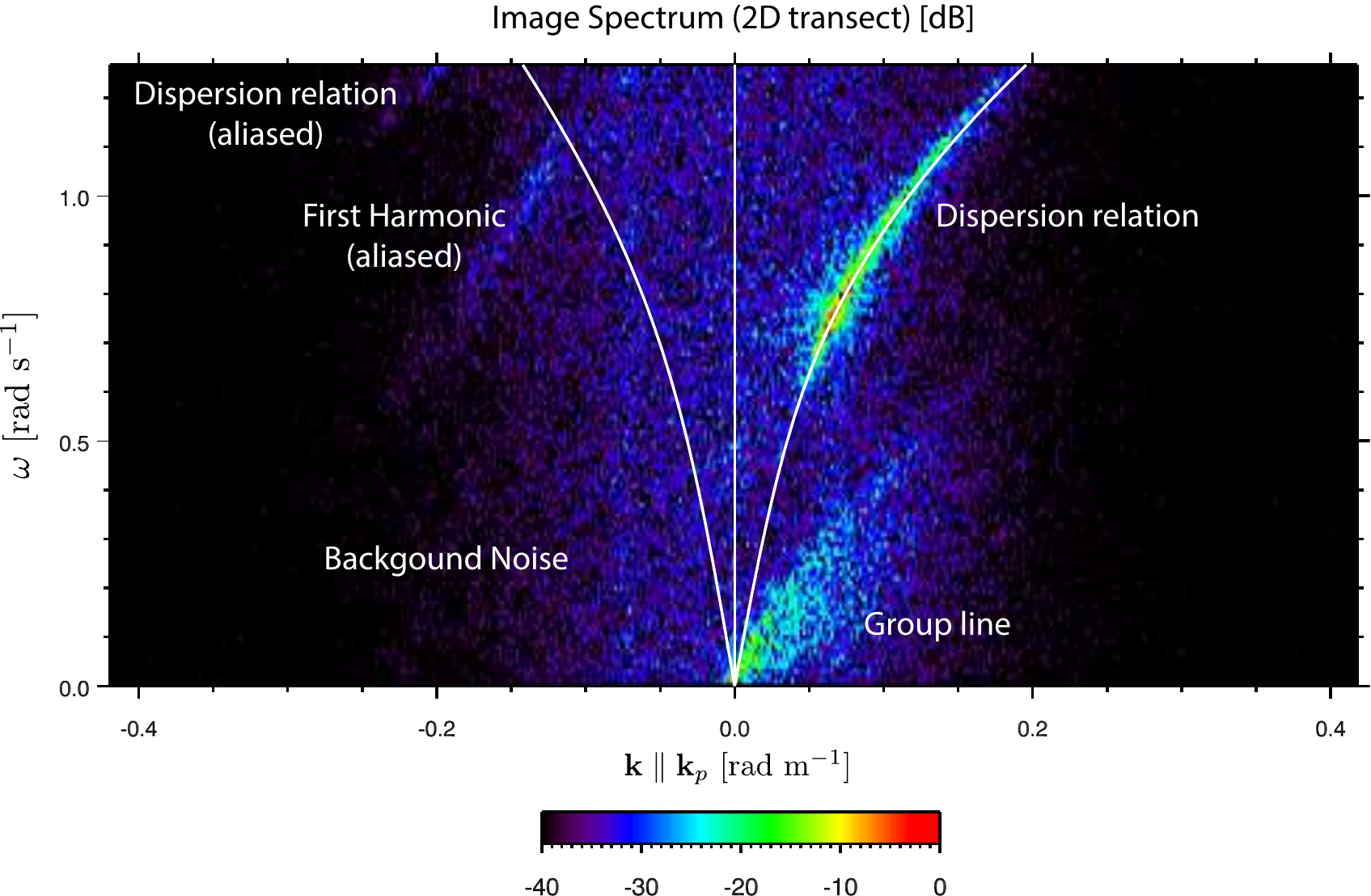}
  \end{center}
  \caption{Estimation of the image spectrum $\mathcal{I}({\bf k}, \omega)$ of a radar image time series.
           The plot corresponds to a transect in the spectral domain $\Omega_{{\bf k}, \omega}$ along the peak wave direction, ${\bf k}\parallel{\bf k}_p$, where ${\bf k}_p$ denotes the peak wave wave number vector.}
  \label{fig:Fimag3D_kwPlus}
\end{figure}

\begin{itemize}
\item[-] Static patterns caused by the long range dependence of the radar backscatter intensity due to the radar equation \cite{Skolnik02}. As this dependence is not on the time domain, the spectral components of this contribution of the image spectrum $\mathcal{I}({\bf k}, \omega)$ correspond to those wave numbers ${\bf k}$, where $\omega \approx 0$ \cite{Young85}. To avoid the static pattern components, the spectral domain $\Omega_{{\bf k}, \omega}$ defined in Expression (\ref{eq:3DSpectralDomain}) includes only those frequencies higher than a threshold value, $\omega \ge \omega_{\rm th}$ \cite{Nieto-Borge04}. For practical applications \cite{Nieto-Borge08}, typical value of the threshold frequency is $f_{\rm th} = 0.04$~Hz (i.e. $\omega_{\rm th} = 2 \pi f_{\rm th}$).

\item[-] Wave components that hold the dispersion relation of linear gravity waves. These spectral $({\bf k}, \omega)$-components are located in the surface $\Lambda_{\omega({\bf k})} \subset \Omega_{{\bf k}, \omega}$ defined by the dispersion relation

\begin{equation}\label{eq:DisShell}
\Lambda_{\omega({\bf k})} \myeq \left\{ ({\bf k}, \omega) \in \Omega_{{\bf k}, \omega} \, | \; \omega = \sqrt{g k \tanh (kd)} + {\bf k} \cdot {\bf U} \right\}
\mbox{,}
\end{equation}

where $k = \left\| {\bf k} \right\|$, $g$ is the acceleration of the gravity, $d$ is the water depth and ${\bf U} = (U_x, U_y)$ is the so-called current of encounter \cite{Senet01} responsible of the Doppler shift in frequency given by the dot product ${\bf k} \cdot {\bf U}$. As in the case of the domain $\Omega_{{\bf k}, \omega}$, $\Lambda_{\omega({\bf k})}$ includes the frequencies that holds the condition $\omega \ge \omega_{\rm th}$. In practice, the domain $\Lambda_{\omega({\bf k})}$ is sampled with the spectral resolutions $(\Delta k_x, \Delta k_y, \Delta \omega)$ given by the FFT algorithm.
This sampled $\Lambda_{\omega({\bf k})}$ domain is commonly known in the analysis of ocean waves by using marine radars as dispersion shell \cite{Young85}.

\item[-] Background noise: This spectral noise is caused by speckle noise due to the roughness of the sea surface induce by the local wind.
The spectral noise appears in the image spectra of different radar systems under different polarization and incidence conditions, such as SAR \cite{Alpers82}, or, like in this case, in X-band marine radar images acquired at grazing incidence conditions \cite{Nieto-Borge08}.
\end{itemize}

Taking into account these different spectral contributions to $\mathcal{I}({\bf k}, \omega)$, it is possible to retrieve sea state information by applying inversion modeling techniques \cite{Young85,Seemann97,Nieto-Borge00,Nieto-Borge04}. The sea state information provided by the inversion modeling techniques are the current of encounter ${\bf U}$ \cite{Senet01,Hessner14}, the water depth $d$ \cite{Bell99,Bell08,Serafino10,Bell11}, as well as the directional and scalar wave spectra and their related sea state parameters, such as peak and mean wave directions, periods, and wave lengths, or directional spreading \cite{Hessner01,Reichert05,Izquierdo05}, among others. One of those parameters is the $H_s$. The method to estimate $H_s$ is described in the following section.

\subsection{Standard method to estimate $H_s$ from X-band radar image time series}\label{ssc:HsEstimation}
As mentioned before, the inversion modeling techniques need to be complemented with an additional algorithm that allows estimating $H_s$ from the analysis of the image spectrum $\mathcal{I}({\bf k}, \omega)$.

The $H_s$ estimation method that is operationally in use considers the existence of the background noise spectral components.
Hence, in a similar way that is carried out for space borne SAR images of the sea surface \cite{Alpers82}, $H_s$ is assumed to be proportional to the squared root of the signal-to-noise ratio $SNR$ \cite{Nieto-Borge08,Chen12}. Taking into account the characteristics of the X-band marine radar (i.e. it is possible to acquire time series of radar images to define the image spectrum in the three-dimensional domain $\Omega_{{\bf k}, \omega}$, rather than in a two-dimensional wave number domain like in the SAR case), $SNR$ is defined as the ratio of the spectral energy of the $({\bf k}, \omega)$-components within the dispersion shell and the spectral energy of the background noise components. For marine radar $SNR$ is defined as \cite{Nieto-Borge08}

\begin{equation}\label{eq:SNRDef}
SNR \myeq \frac{\displaystyle \int_{\Lambda_{\omega({\bf k})}} \mathcal{M}(k) \, \mathcal{I}({\bf k}, \omega) \, dk_x dk_y d \omega}{\displaystyle \int_{\Omega_{{\bf k}, \omega} \backslash \Lambda_{\omega({\bf k})}} \mathcal{I}({\bf k}, \omega) \, dk_x dk_y d \omega}
\mbox{ ,}
\end{equation}

where $\mathcal{M}(k)$ is an empirical modulation transfer function that takes into account the radar imaging mechanisms at grazing incidence for different wave numbers \cite{Ziemer94}. Comparing the spectra derived from radar images with in-situ data, the dependence found for the modulation transfer function was $\mathcal{M}(k) \approx k^{-1.2}$ \cite{Nieto-Borge04}.
In Equation (\ref{eq:SNRDef}), the integration domain $\Lambda_{\omega({\bf k})} \subset \Omega_{{\bf k}, \omega}$ is the dispersion shell defined in Equation (\ref{eq:DisShell}), and $\Omega_{{\bf k}, \omega} \backslash \Lambda_{\omega({\bf k})} \myeq \left\{ ({\bf k}, \omega) \in \Omega_{{\bf k}, \omega} \, | \; ({\bf k}, \omega) \not \in \Lambda_{\omega({\bf k})} \right\}$ denotes all those $({\bf k}, \omega)$-components outside the dispersion shell.
Note that the integration domains used in Equation (\ref{eq:SNRDef}) depend on the estimation of the parameters that affect the dispersion relation, such as ${\bf U}$ and $d$, which are usually derived from the inversion modeling scheme as it was mentioned above.
From Equation (\ref{eq:SNRDef}), $H_s$ is estimated as

\begin{equation}\label{eq:HsfromSNR}
H_s = c_0 + c_1 \sqrt{SNR}
\mbox{ ,}
\end{equation}

\noindent
where $c_0$ and $c_1$ are calibration constants that are determined empirically by using in-situ sensor data, for example $H_s$ values acquired by a wave buoy. The values of $c_0$ and $c_1$  depend on the different installation conditions (i.e. angle of incidence, range of measurement, used radar system, etc. \cite{Hessner01}).

\subsubsection{Limitations of the SM estimation}\label{ssc:HsLimitStand}
Expression (\ref{eq:HsfromSNR}) provides reliable results for operational purposes \cite{Hessner99}, permitting the estimation of $H_s$ in real time for sea state monitoring purposes from X-band radar data sets \cite{Hessner01,Reichert05}.
However, a proper estimation of $H_s$ depends on a correct determination of $SNR$ by using Equation (\ref{eq:SNRDef}).
Under some circumstances, the $SNR$ estimation does not suite the range applicability of Equation (\ref{eq:HsfromSNR}) \cite{Nieto-Borge98a}.
For example, under the presence of low wind conditions, the spectral energy of the background noise takes small values, and the denominator in Equation (\ref{eq:SNRDef}) is too small as well.

Consequently, this effect leads to high values of $SNR$, which causes that Equation (\ref{eq:HsfromSNR}) overestimates the value of $H_s$.
In addition, another effect occurs for low amplitude swell, which induces low backscatter modulation \cite{Schmidt95,Rozenberg96}. Consequently,  the numerator in Equation (\ref{eq:SNRDef}) yields too small values of $SNR$, which leads Equation (\ref{eq:HsfromSNR}) to underestimate $H_s$.

It should be noted that the backscattering phenomenon at grazing incidence (i.e. the marine radar operational conditions) is not fully explained yet \cite{Plant08} and the empirical modulation transfer function $\mathcal{M}(k)$ in Equation (\ref{eq:SNRDef}) does not take into account all the microwave backscattering imaging mechanisms present for the marine radar measuring conditions. Due to the reasons above mentioned, the estimation of $H_s$ should include more parameters than only $SNR$ given by Equation (\ref{eq:SNRDef}).
Hence, Equation (\ref{eq:HsfromSNR}) needs to be improved to include additional parameters. One possible parameter could be the wind speed, but that would need an additional sensor.
In this work only the sea state parameters delivered by the standard wave analysis of X-band data sets have been considered \cite{Izquierdo05,Reichert05}.
These parameters are delivered from the un-scaled estimation of the wave spectra, such as the peak wave number $k_p$, the peak frequency $f_p$, or the different estimations of the mean powers of the frequency derived from ratios of the spectral moments $m_j/m_0$.
They are related to different estimators of the mean period giving more weight to different regions of the frequency domain.
Hence, the normalized spectral moment of $j^{\rm th}$-order is defined from the frequency spectrum $S(f)$ as

\begin{equation}\label{eq:mnm0Ratio}
	\overline{m}_j \myeq \frac{m_j}{m_0} = \frac{\displaystyle \int_{f_{\rm th}}^{f_c} f^j S(f) d f}{\displaystyle \int_{f_{\rm th}}^{f_c} S(f) d f}
	\mbox{ ,}
\end{equation}

Where the frequency $f = \omega/ (2 \pi)$. Note that the ratio given by Equation (\ref{eq:mnm0Ratio}) does not depend on the scale of the spectrum $S(f)$ because it is normalized by its own area.

\section{Description of the data used}\label{sc:UsedData}
The marine radar image time series used in this work were acquired by WaMoS-II systems \cite{Hessner01,Hessner08}. WaMoS-II is an operational Wave Monitoring System built up for the specific purpose of wave and current measurement by X-band marine radars, which was originally developed at the German research institute HZG (Helmholtz-Zentrum Geesthacht). The measuring system consists of a conventional X-band marine radar, and a high-speed video digitizing and storage device connected to a computer. Hence, the analogue radar video signal is read out and digitized into a scale of grey levels.
This information is transferred and stored on the computer where the wave analysis software carries out the estimation of the sea state parameters. For WaMoS-II measurements, radar raw data signals are needed.
Hence, preprocessing filters, such as rain filter, anti clutter filter, image intensity amplification, etc., must be switched off.
The marine radar image time series used in this work have been measured in different geographic locations, the North Sea and the Sable Field in South Africa:

\begin{itemize}
\item[-] North Sea: WaMoS-II data from two stations located at the North Sea have been used. During the measurement period of each station, different sea state cases were recorded. These two North Sea locations are:

\begin{enumerate}
\item Fino~1 Research Platform (\cite{FINO1}): This platform is located at the German basin of the North Sea (54$^\circ$00'53.5''\,N, 06$^\circ$35'15,5''\,E) at 45~km to the north of Borkum island. The local water depth is about 30~m. The period of data used for the analysis is from July 1$^{\rm st}$, 2004 to August, 25$^{\rm th}$, 2009. The WaMoS-II system at Fino~1 measured a radar image time series every 3 minutes.

\item Ekofisk Oil Field Complex: This complex is located in Norwegian sector of the North Sea (56$^\circ$32'57.11''\,N, 03$^\circ$12'35.95''\,E). The local water depth in the area is about 75~m. The period of data used for the analysis is from October 10$^{\rm th}$, 2004 to November, 11$^{\rm th}$, 2009. The WaMoS-II system at Ekofisk measured a radar image time series every 4 minutes.
\end{enumerate}

\item[-] Sable Field: In this case, WaMoS-II data from only one station is available. This location is:
\begin{enumerate}
\item Glas Dowr: This area is located at the Bredasdorp basin about 150~km Southwest of Mossel Bay off South Africa (35$^\circ$12'25.7''\,S,   21$^\circ$19'18.4''\,E). The local water depth in the area is about 100~m. The data were acquired by a WaMoS-II system installed on board of FPSO Glas Dowr. The measurement period cover the dates from March 1$^{\rm st}$, 2008 to August 31$^{\rm st}$, 2008. During this measurement period, several cases of long swell were acquired. The WaMoS-II system at Glas Dowr measured a radar image time series every 3 minutes.
\end{enumerate}
\end{itemize}

For each location, the WaMoS-II systems were set-up in the standard way for operational wave spectral estimation to derive the related sea state parameters \cite{Hessner99,Nieto-Borge00,Hessner01,Reichert05}. Hence, each radar measurement is composed of a time series of 32 consecutive radar images, where the sampling time of those time series is the antenna rotation period ($\approx 2.5$~s) and the sampling spatial resolution is given by the range and azimuthal resolutions of the radar system. These raw data defined in polar coordinates (range and azimuth) are interpolated onto a Cartesian grid to enable the proper computation of the image spectra by using FFT-based algorithms. The spatial resolution of the interpolated Cartesian grid used in this work is $\approx 7.5 \times 7.5$~m$^2$.
As reference in-situ data, $H_s$ estimations measured from a buoy deployed in the vicinity of each radar location were used.

\subsection{Predictive variables}\label{ssc:InputPar}
There is a clear need to improve the robustness of $\hat{H}_s$. To do this, it is basic that the predictive variables contain as much information as possible about $H_s$. Thus, it is considered to use not only the signal-to-noise ratio as predictive variables for the SVR (as in the SM), which is the proposed method to do the prediction, but also to include additional parameters related with the wave length and periods of the wave field and the normalized spectral moments $\overline{m}_j$ given by Equation~(\ref{eq:mnm0Ratio}). The list of predictive variables considers that, as it was mentioned above, the radar imagery mechanisms depend on the modulation of the backscattering of the electromagnetic fields by the long waves (swell and/or wind sea wave fields) \cite{Schmidt95,Rozenberg96,Nieto-Borge04,Plant08}. As this modulation is produced in the spatial domain, it depends on wave lengths, or, alternatively, on wave periods. Hence, using the standard output parameter list derived from the operational WaMoS-II analysis, the natural choice was to consider parameters related to peak or mean periods, or wave numbers. The list is completed with the normalized third-order moment $\overline{m}_3$ because this spectral moment is calculated giving more weight to higher frequencies than the other moments considered $\overline{m}_1$ and $\overline{m}_2$, which give respectively the mean wave period estimator $T_{m01}$, and $T_{m02}$. Taking that into account, the following predictive variables have been used for the proposed SVR method for the prediction:

\begin{itemize}
\item $SNR$: signal-to-noise ratio defined from Equation~(\ref{eq:SNRDef}).
\item $k_p$: peak wave number derived from the wave number spectrum.
\item $f_p$: peak frequency derived from the frequency spectrum $S(f)$.
\item $\overline{m}_1$: this normalized spectral moment is an estimator of the mean frequency using the spectrum $S(f)$ as weighting function. This parameter is related to the $T_{m01}$ estimator of the mean wave period (i.e $T_{m01} = 1 / \overline{m}_1$).
\item $\overline{m}_2$: this parameter is the estimation of the mean value of $f^2$. $\overline{m}_2$ is related to the estimator of the mean wave period $T_{m02} = 1 / \sqrt{\overline{m}_2}$.
\item $\overline{m}_3$: this parameter is the estimation of the mean value of $f^3$. $\overline{m}_3$ is the normalized spectral moment used in this work that gives more weight to the high frequency tail of the spectrum $S(f)$.
\end{itemize}

As mentioned before, in addition of these parameters derived from the analysis of the radar data, values of $H_s$ from buoys moored in the vicinity of the sea surface area illuminated by the radar antenna were used to obtain $\hat{H}_s$ using the SVR algorithm (Section \ref{sec:state_of_the_art}.

\section{Experiments and results}\label{Experiments}
This section presents the $H_s$ estimations obtained by the proposed SVR method for three platforms (Fino~1, Ekofisk and Glas Dowr) considered in the study. The SVR objective is obtained by means of in-situ sensors (buoy). To validate the proposed method, these results are compared with those by the SM described in Section \ref{ssc:HsLimitStand}. First of all, it is detailed how the databases obtained from the considered platforms are processed to train the SVR. After this step, the results obtained and the SVR and SM performances on this problem are described.

\subsection{Pre-processing of the databases}
In order to proceed to the training of the SVR model, the values of the SVR hyper-parameters $C$, $\epsilon$ and $\gamma$ must be chosen. For this purpose, a GS guided by the performance measured by cross validation on a subset of the database, so-called {\em validation set}, will be used. The size of the validation set has been selected depending on the total number of samples available for each one of the platforms, making it large enough to prevent over-fitting but in such a way that the required computation time to perform the evaluation is not excessive.
With this in mind, the size of the validation set for each one of the platforms is: Ekofisk (10\% of the samples), Glas Dowr (8\% of the samples) and Fino~1 (2.5\% of the samples).

Once the hyper-parameters have been set, the remaining data samples are divided into two subsets: Training set with 80\% of the samples, and Test set with the remaining 20\%. The SVR model obtained after the optimization of the hyper-parameters is then trained using the training set and its performance evaluated using the data from the Test set.  The complete process to train the SVR is outlined in Figure \ref{fig:SVRtraining}, whereas Table \ref{SVR_training} shows how the specific databases obtained from the different platforms considered where divided to train the SVR.

\begin{figure}[ht]	
  \begin{center}
    \includegraphics[width=0.65\textwidth]{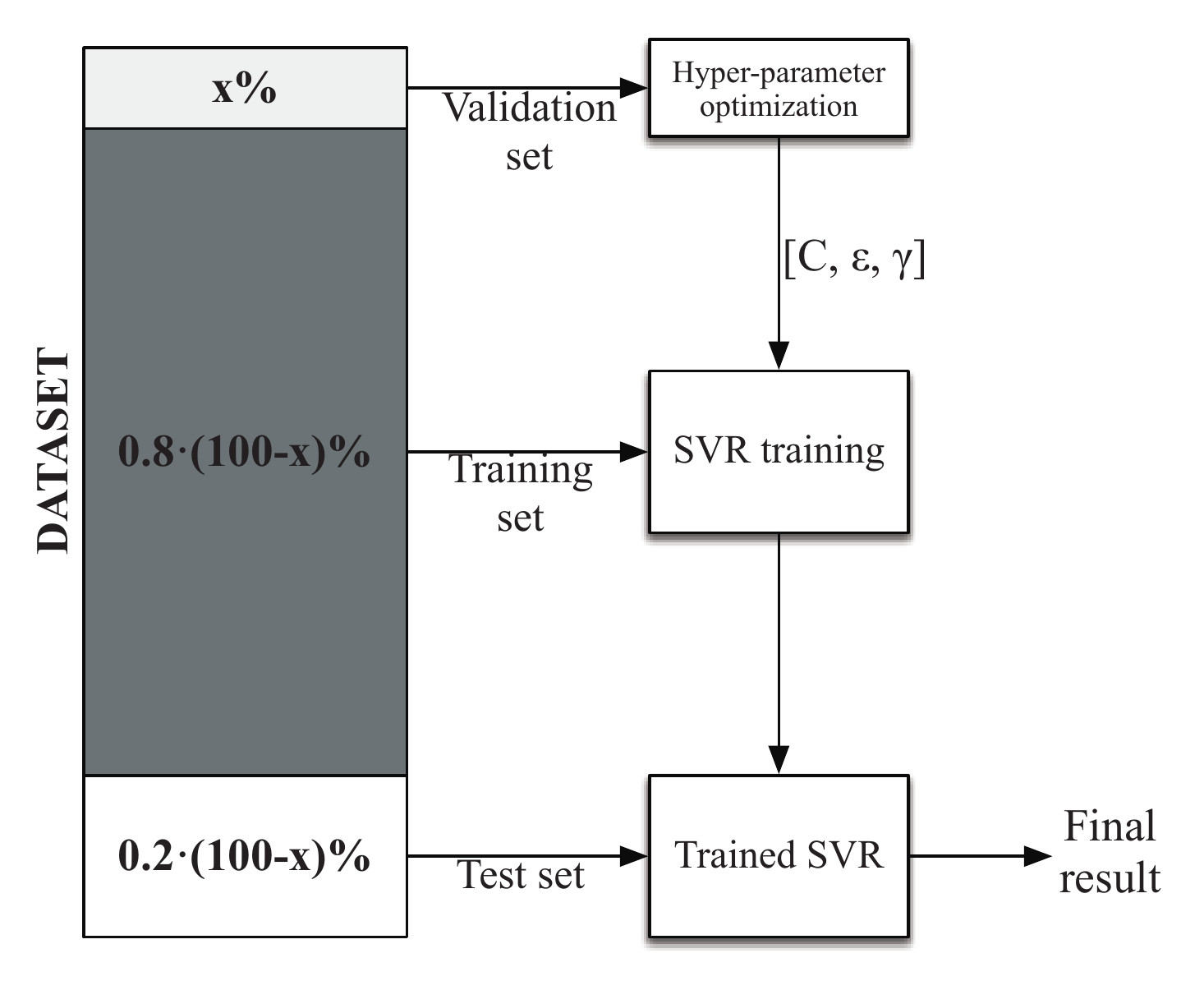}
  \end{center}
  \caption{Illustration of the SVR training and testing process.}
  \label{fig:SVRtraining}
\end{figure}

\begin{table}[ht]
\begin{center}
\caption{\label{SVR_training} Division of the databases into different sets for the experiments. }
\vspace{0.3cm}
\begin{tabular}{|c|c|c|c|}
       \hline
Platform		& Validation		& Training		& Testing \\
\hline
FINO~1		& 2.5\%			& 78\%		& 19.5\% \\
Ekofisk		& 10\%			& 72\%		& 18\%\\
Glas Dowr		& 8\%			& 73.6\%		& 18.4\% \\
\hline
\end{tabular}
\end{center}
\end{table}

\subsection{Results obtained}

Figures \ref{fig:FINO}, \ref{fig:Ekofisk} and \ref{fig:Glas} show the scatter plots ($H_s$ estimated with the predictive method versus the real $H_s$ measured at buoy), for Fino~1, Ekofisk and Glas Dowr, respectively. In each figure the comparison of the SVR with the SM is carried out.

\begin{figure}[ht]	
  \begin{center}
   \subfigure[]{\includegraphics[width=0.7\textwidth]{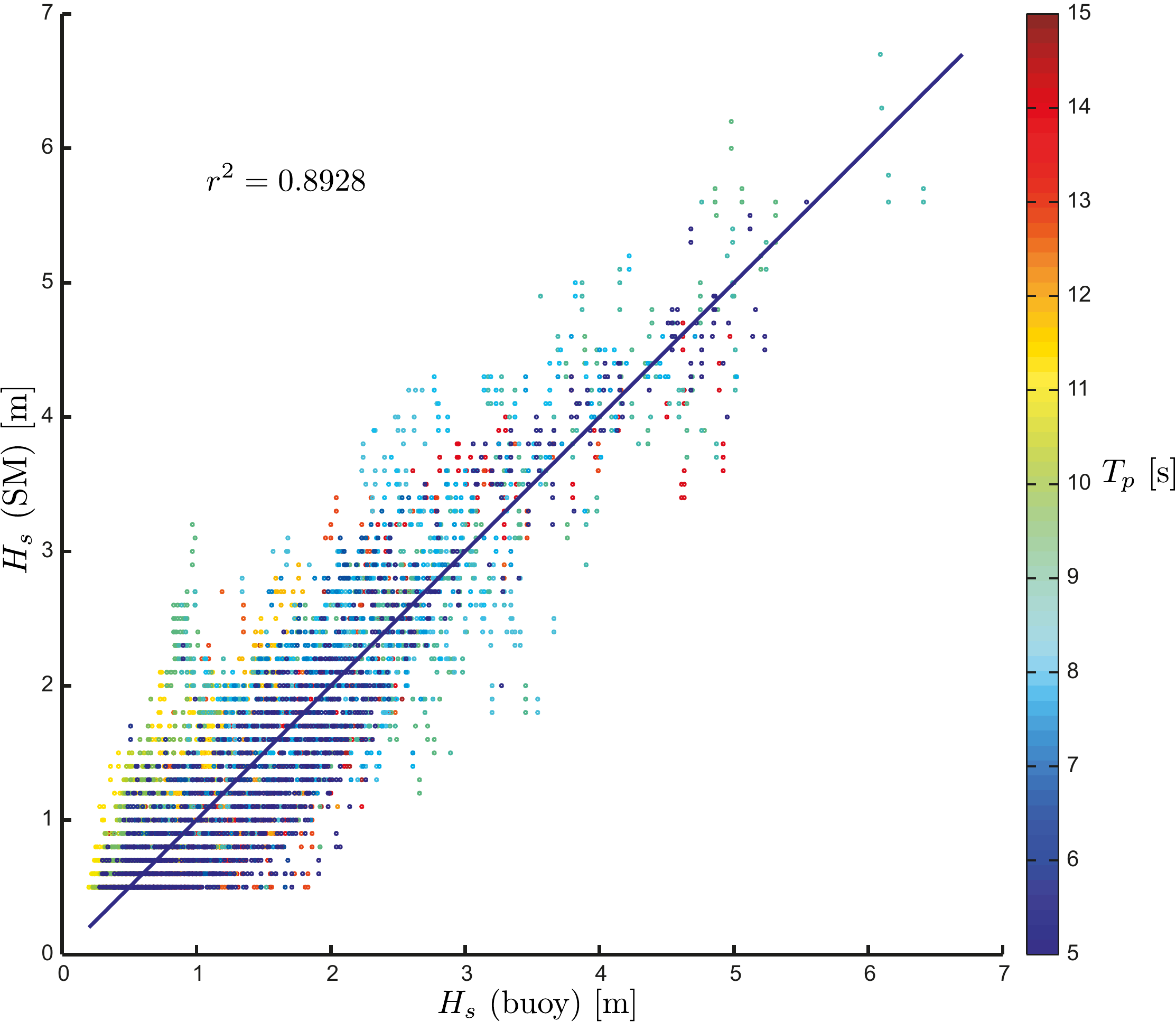}}
    \subfigure[]{\includegraphics[width=0.7\textwidth]{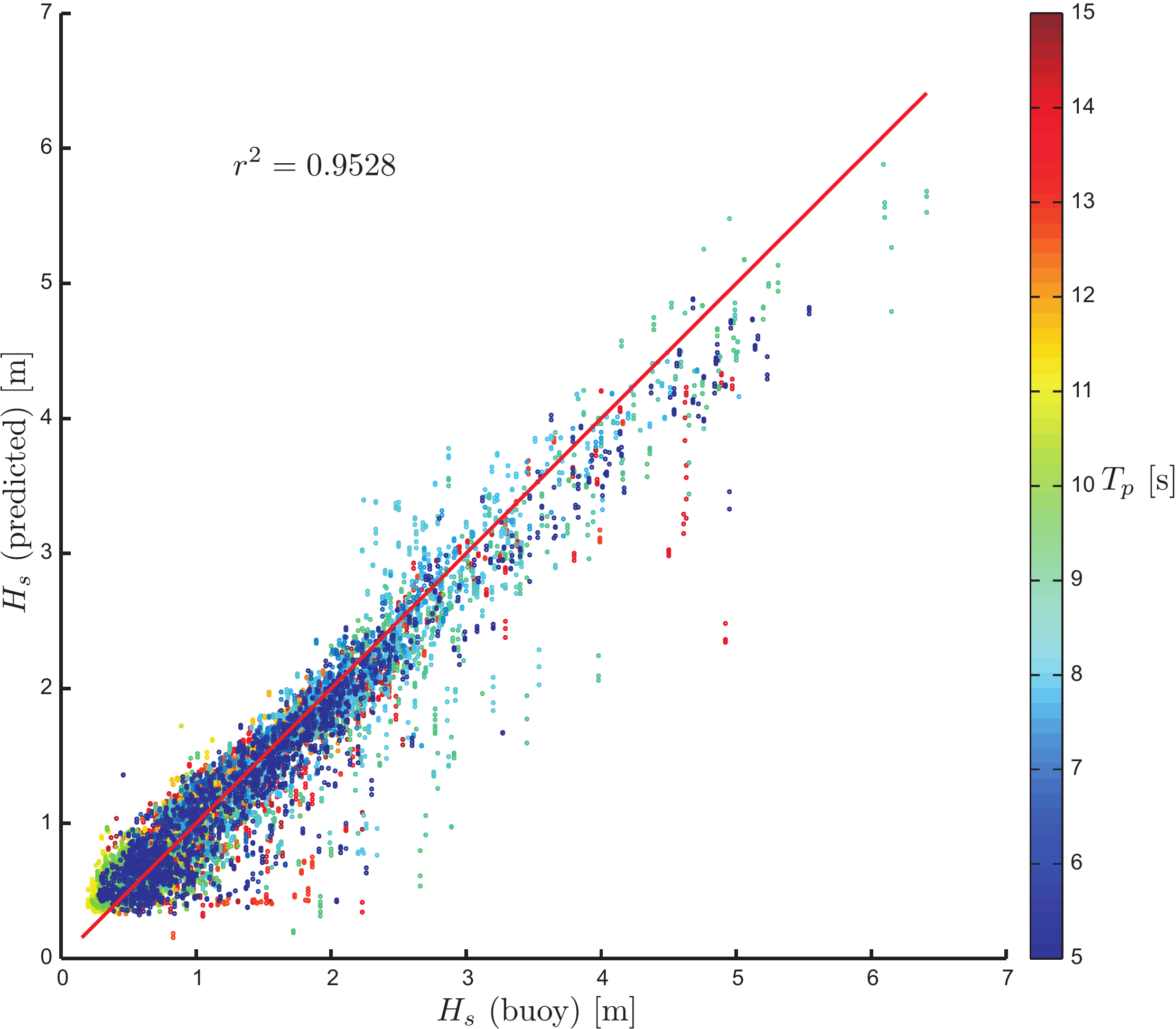}}
  \end{center}
  \caption{Scatter plots colored by $T_p$ of the $H_s$ measured by the buoy and estimated by the SM and the SVR approach for the test data set at FINO~1; (a) SM; (b) SVR approach. The solid line indicates the best fit.}
\label{fig:FINO}
\end{figure}

\begin{figure}[ht]	
  \begin{center}
     \subfigure[]{\includegraphics[width=0.7\textwidth]{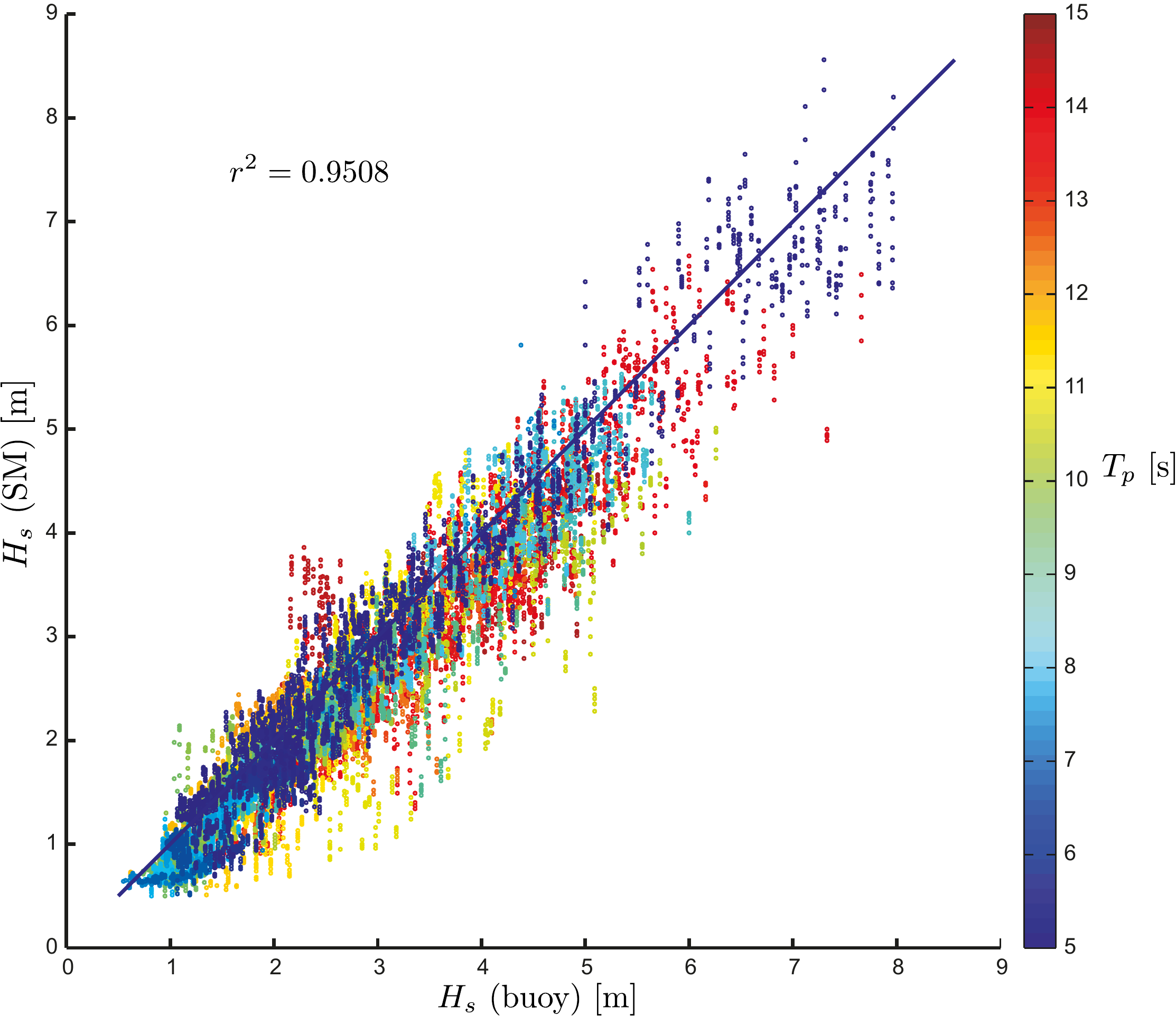}}
   \subfigure[]{\includegraphics[width=0.7\textwidth]{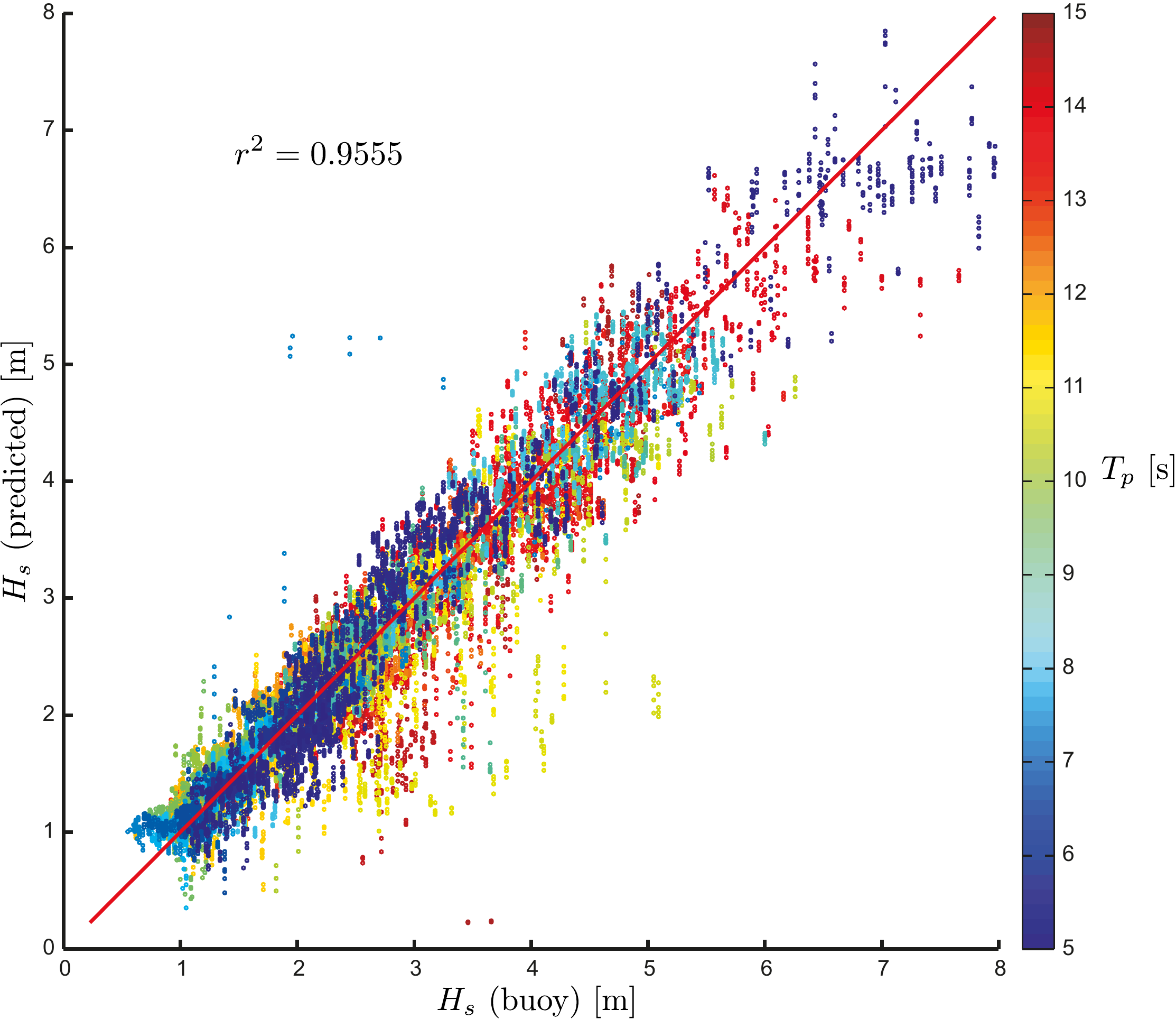}}
  \end{center}
  \caption{Scatter plots colored by $T_p$ of the $H_s$ measured by the buoy and estimated by the SM and the SVR approach for the test data set at Ekofisk; (a) SM; (b) SVR approach. The solid line indicates the best fit.}
\label{fig:Ekofisk}
\end{figure}

\begin{figure}[ht]	
  \begin{center}
      \subfigure[]{\includegraphics[width=0.7\textwidth]{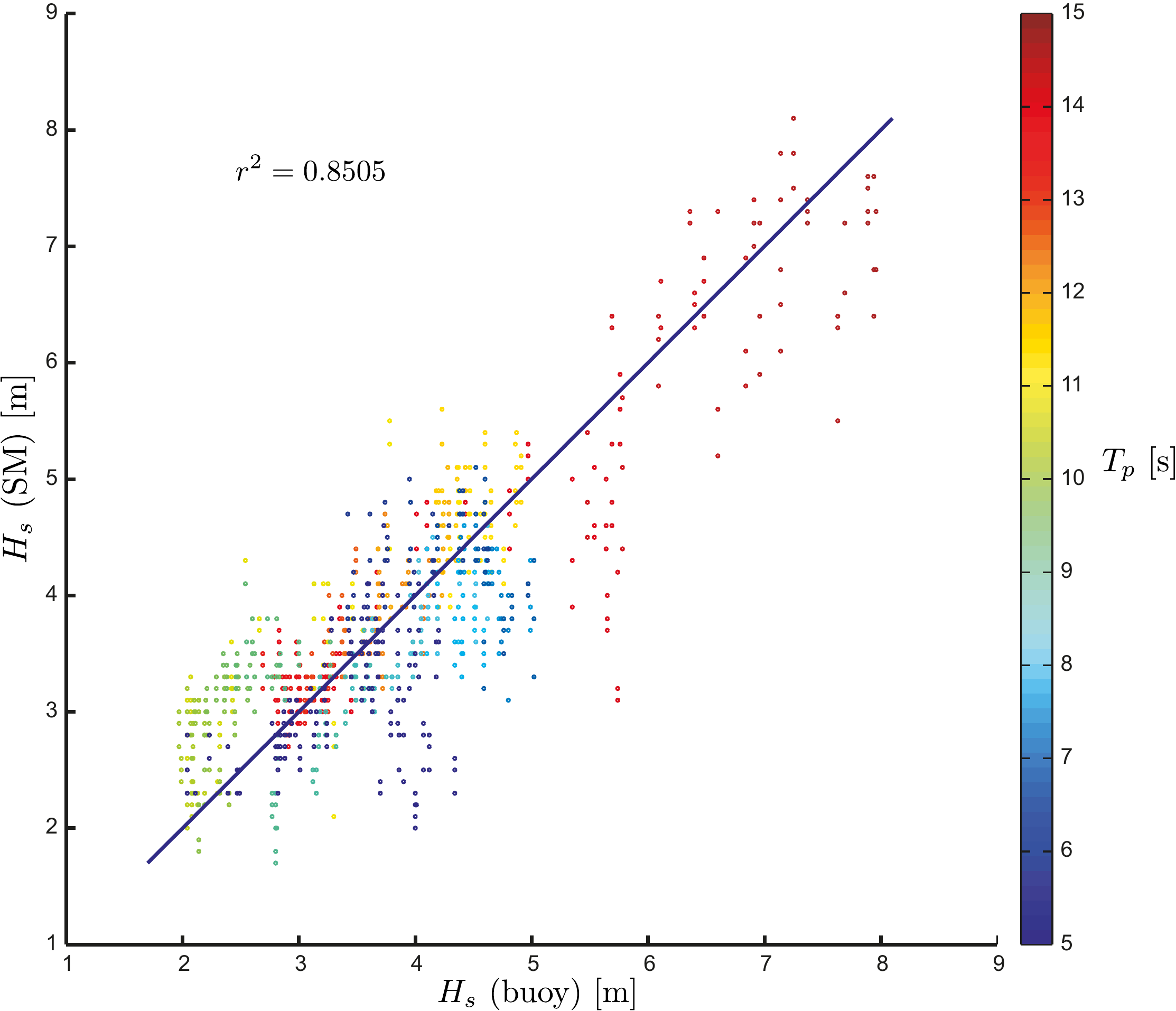}}
    \subfigure[]{\includegraphics[width=0.7\textwidth]{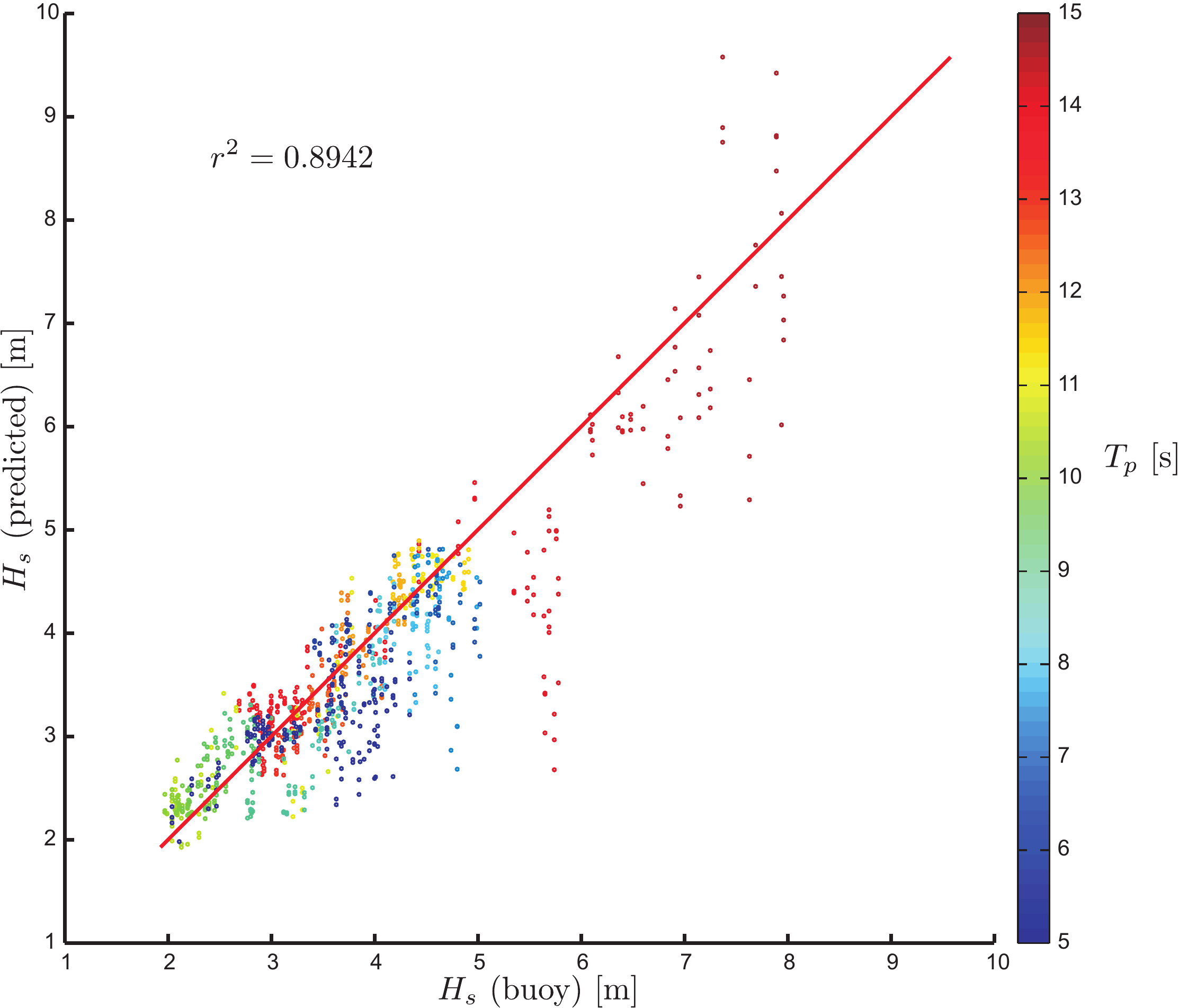}}
  \end{center}
  \caption{Scatter plots colored by $T_p$ of the $H_s$ measured by the buoy and estimated by the SM and the SVR approach for the test data set at Glas Dowr; (a) SM; (b) SVR approach. The solid line indicates the best fit.}
\label{fig:Glas}
\end{figure}

Note that the plots have been depicted in color scale by $T_p$, so the performance at different $T_p$ values can be observed.
Table~\ref{Comparativa} shows a summary of the  obtained results, including values for $r^2$ and \ac{MSE} in all the platforms considered.

\begin{table}[ht]
\begin{center}
\caption{\label{Comparativa} Comparative results of the $H_s$ estimation by the SM and the SVR approaches.} \vspace{0.3cm}
\begin{tabular}{lrrcrrcrrcrr}
\hline
& \multicolumn{2}{c}{FINO~1}  & &\multicolumn{2}{c}{Ekofisk}   & &\multicolumn{2}{c}{Glas Dowr} \\
\cline{2-3}   \cline{5-6}  \cline{8-9}
 &MSE  & $r^2$ && MSE  & $r^2$ && MSE  & $r^2$ \\
\hline
SM   & 0.18 m & 0.89 &&  0.22 m & 0.95 && 0.38 m & 0.85 \\
SVR & \textbf{0.08 m} & \textbf{0.95} &&  \textbf{0.16 m} & \textbf{0.96} && \textbf{0.30 m} & \textbf{0.89} \\
\hline
\end{tabular}
\end{center}
\end{table}

It is easy to see how the SVR approach outperforms SM in all the platforms considered, with values of $r^2$ significantly better for Fino~1 (0.95 vs. 0.89), slightly better at Ekofisk (0.96 vs. 0.95) and also better at Glas Dowr (0.89 vs. 0.85).
The MSE values for the three locations are 0.18~m vs. 0.08~m in Fino~1, 0.22~m vs. 0.16~m in Ekofisk, and 0.38~m vs. 0.30~m in Glas Dowr.
In this latter case, the poorer performance of both approaches respect to the other platforms requires a deeper analysis.
A first hypothesis is that the algorithms' performance is affected by the number of training/test samples available.
In order to clarify this point, some more experiments in Fino~1 platform data have been carried out, where different size for training/test partitions for the SVR have been used (60\% train, 40\% test, 40/60 and 20/80).
Note that in the last case, the number of training samples is very reduced, as in the Glas Dowr case.
Table~\ref{SVR_partition} shows the results obtained in these experiments, where it can be seen that the performance of the SVR is affected somehow by the number of training samples.

\begin{table}[ht]
\begin{center}
\caption{\label{SVR_partition} SVR performance with different train/test partitions at Fino 1 measuring station. }
\vspace{0.3cm}
\begin{tabular}{|c|c|c|}
       \hline
Partition (train/test)	& $r^2$  &MSE\\
\hline
78/19.5 (paper)   & 0.95      & 0.08\\
60/40             & 0.90      & 0.12\\
40/60             &0.90       & 0.10\\
20/80             &0.87       & 0.12\\
\hline
\end{tabular}
\end{center}
\end{table}

This indicates that there must be a different cause for the poor performance of the algorithms in this platform.
A possible reason for this poor algorithms' performance might be found in the sea state conditions. Therefore, as it was discussed before, the model used in the SM, which uses Equations (\ref{eq:SNRDef}) and (\ref{eq:HsfromSNR}) tends to provide not so accurate $H_s$ estimations under some circumstances.
In order to obtain additional information of the sea state conditions, the value of the significant wave steepness have been calculated ($\epsilon_s = 2 \pi H_s / g T_p^2$) from the buoy data (i.e. the reference sensor) in all the locations considered (Table~\ref{Stepness}).

\begin{table}[ht]
\begin{center}
\caption{\label{Stepness} Averaged significant wave steepness $\epsilon_s$ derived from the buoy data at the dates when the measurements were obtained in the different platforms considered. }
\vspace{0.3cm}
\begin{tabular}{|l|c|c|}
       \hline
Platform		& Mean value      & Standard deviation        \\
\hline
FINO~1		       & 0.0219\			& 0.0111\\
Ekofisk		       & 0.0245\			& 0.0091\\
Glas Dowr		   & 0.0182\			& 0.0060\\
\hline
\end{tabular}
\end{center}
\end{table}

As can be seen, the significant steepness in Glas Dowr is significantly smaller (with averaged values of $\epsilon_s$ of swell sea state conditions \cite{Goda10}) than in the other two platforms. Then, this may indicate that Glas Dowr is mainly dealing with situations in which swell is the dominant sea state, reducing the performance of the algorithms.

The analysis of the SVR performance can be extended by showing the $H_s$ estimation obtained with this technique in the test set, in terms of the temporal variation of $H_s$. Figures \ref{Temporal_evolution} (a), (b) and (c) show this temporal SVR performance in Fino~1, Ekofisk and Glas Dowr platforms, respectively.

\begin{figure}[ht]	
  \begin{center}
      \subfigure[]{\includegraphics[width=0.7\textwidth]{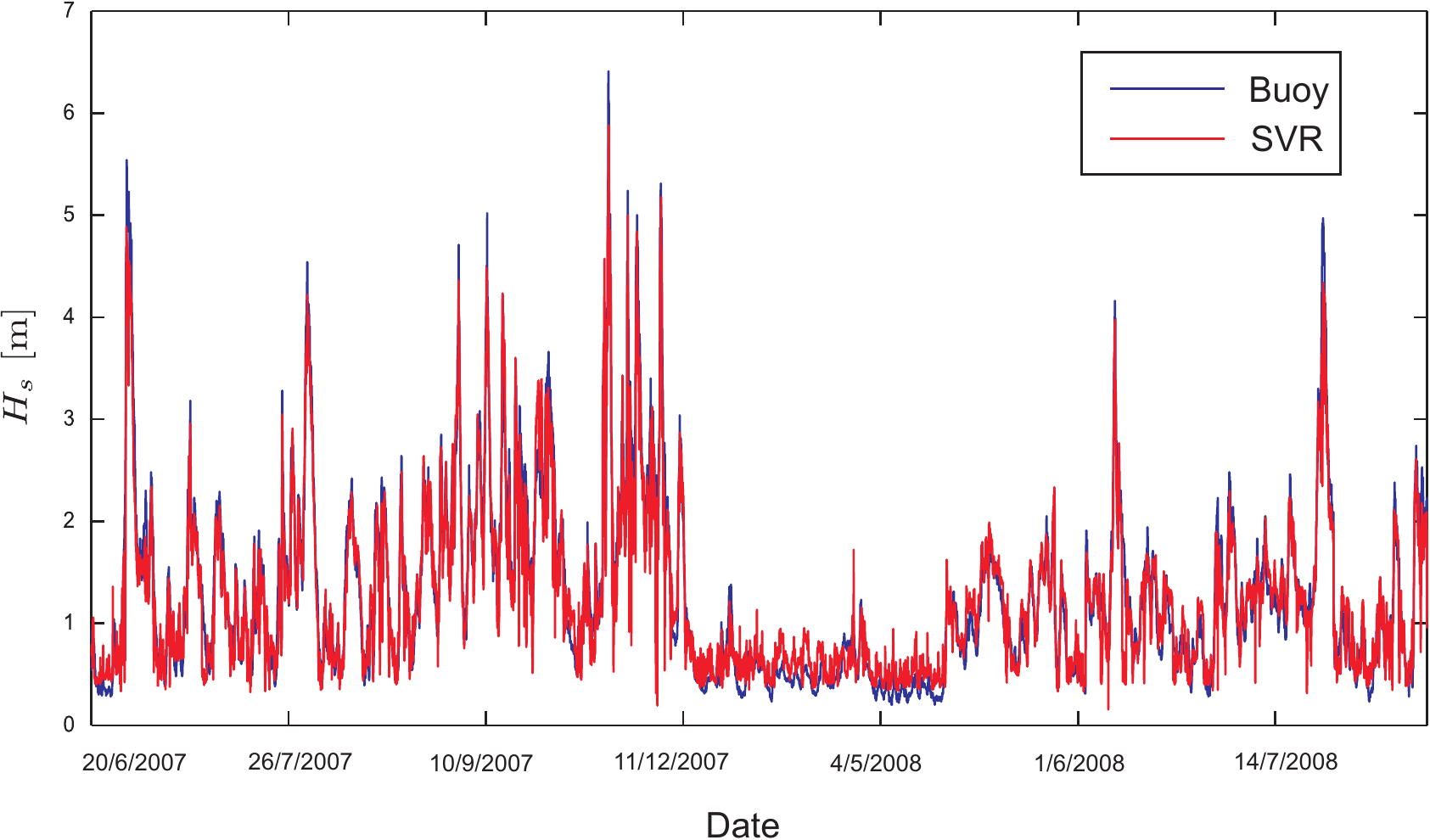}}
    \subfigure[]{\includegraphics[width=0.7\textwidth]{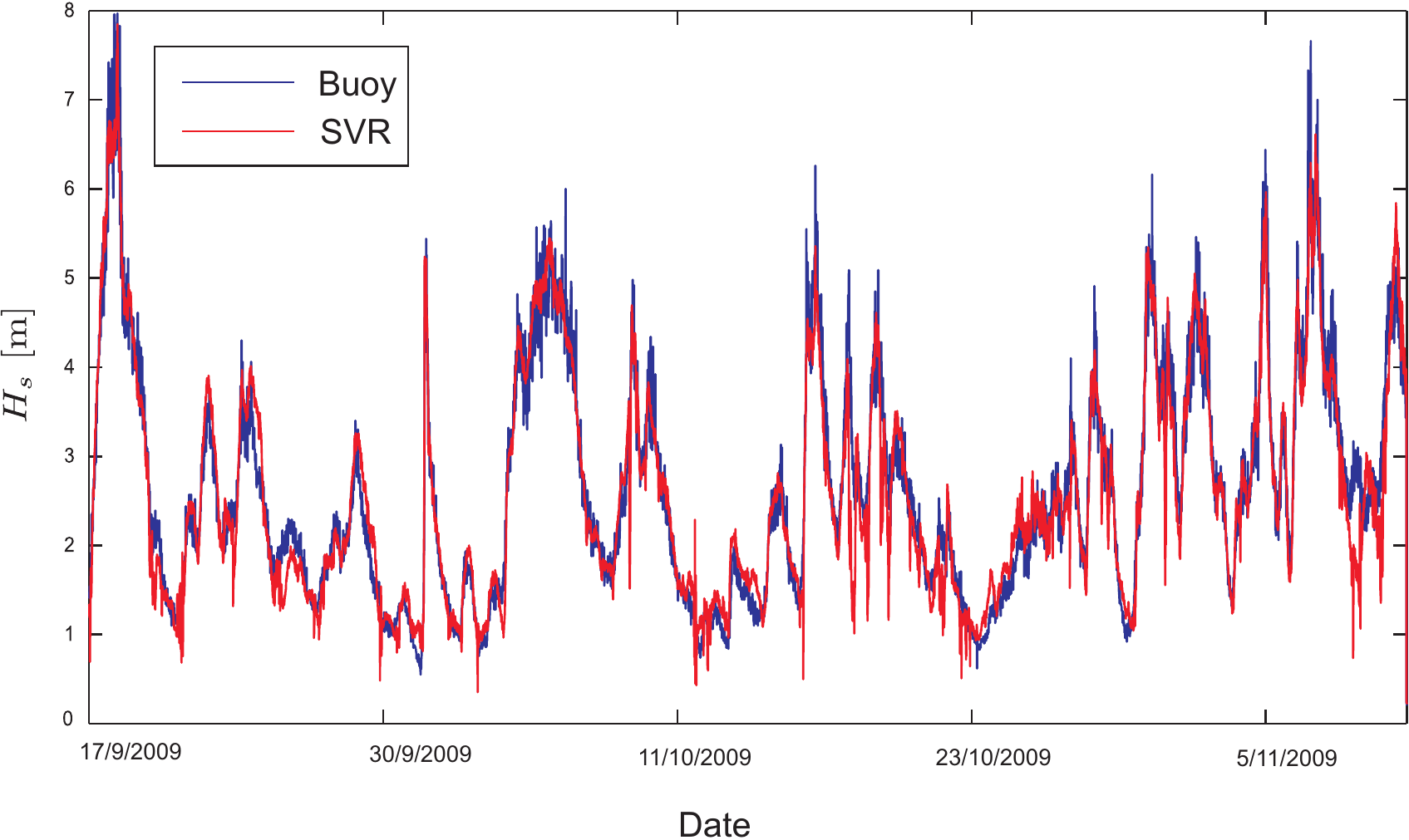}}
    \subfigure[]{\includegraphics[width=0.7\textwidth]{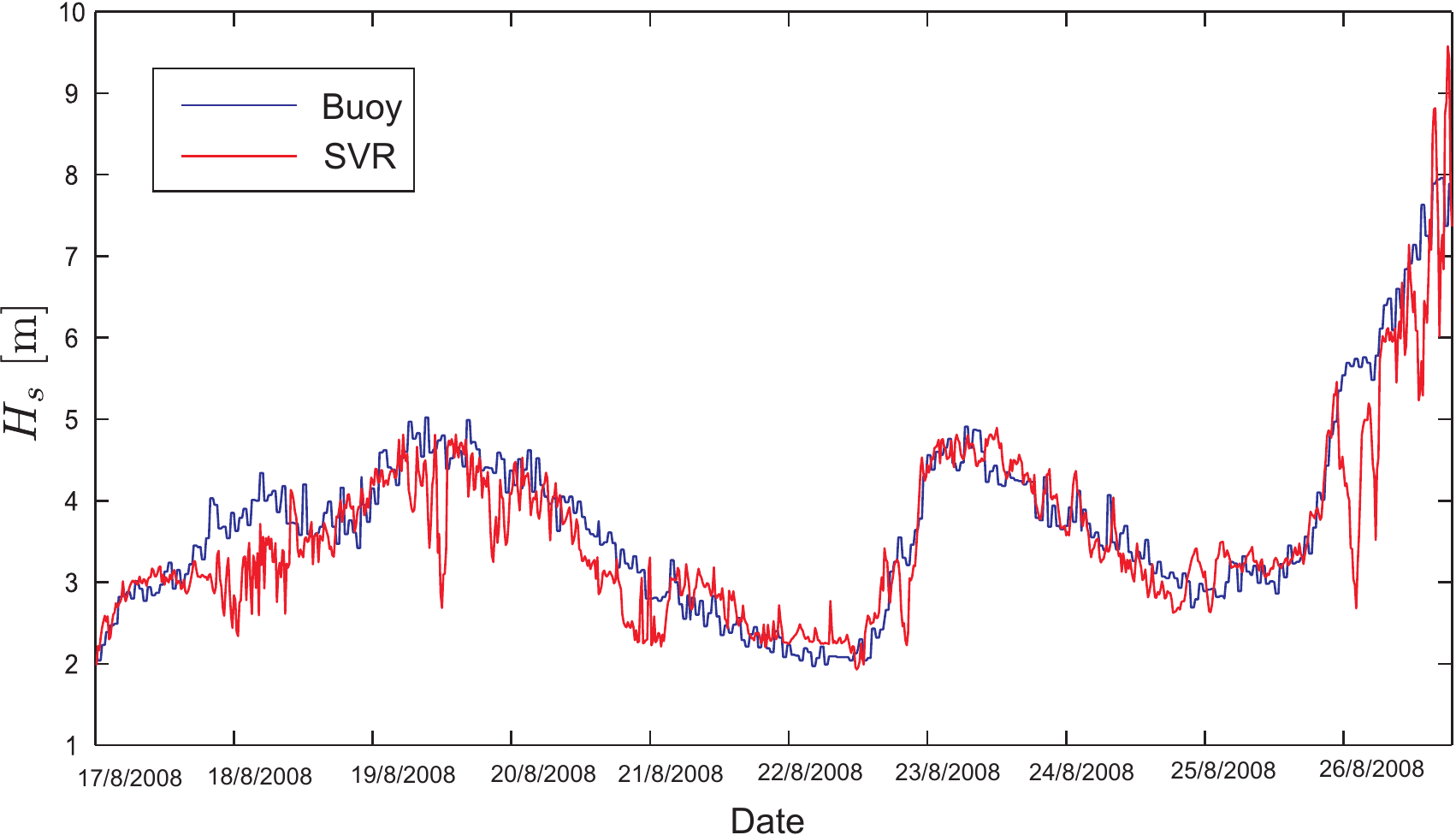}}
  \end{center}
  \caption{\label{Temporal_evolution} Temporal evolution of the $H_s$ estimation obtained with the SVR in the different platforms considered; (a) Fino~1; (b) Ekofisk; (c) Glas Dowr.}
\end{figure}

Figure \ref{Diferencias} complements the temporal figures before by including a direct comparison in terms of $H_s$ differences (measured minus predicted, $H_s-\hat{H}_s$) in all the locations considered.

\begin{figure}[ht]	
  \begin{center}
      \subfigure[]{\includegraphics[width=0.7\textwidth]{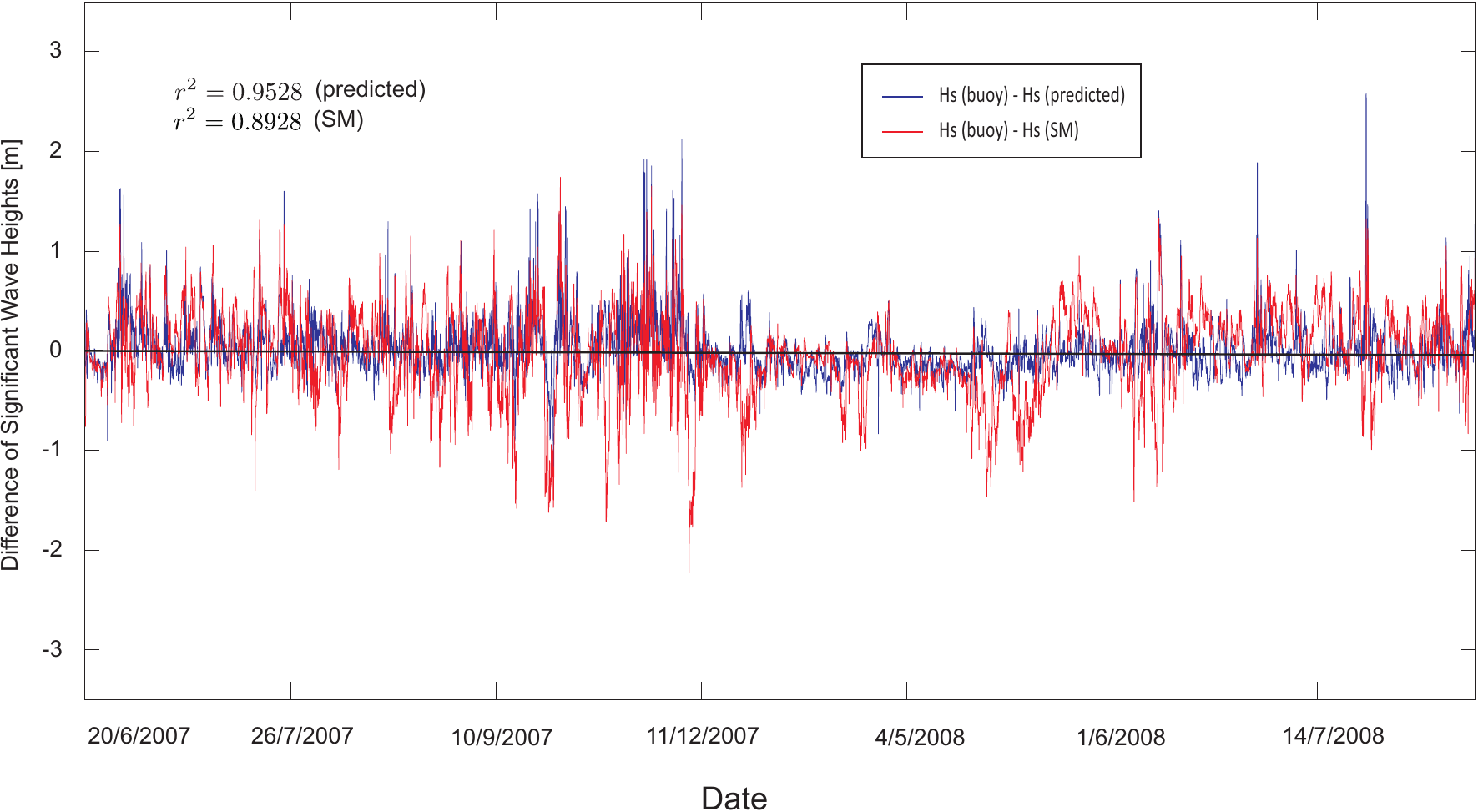}}
    \subfigure[]{\includegraphics[width=0.7\textwidth]{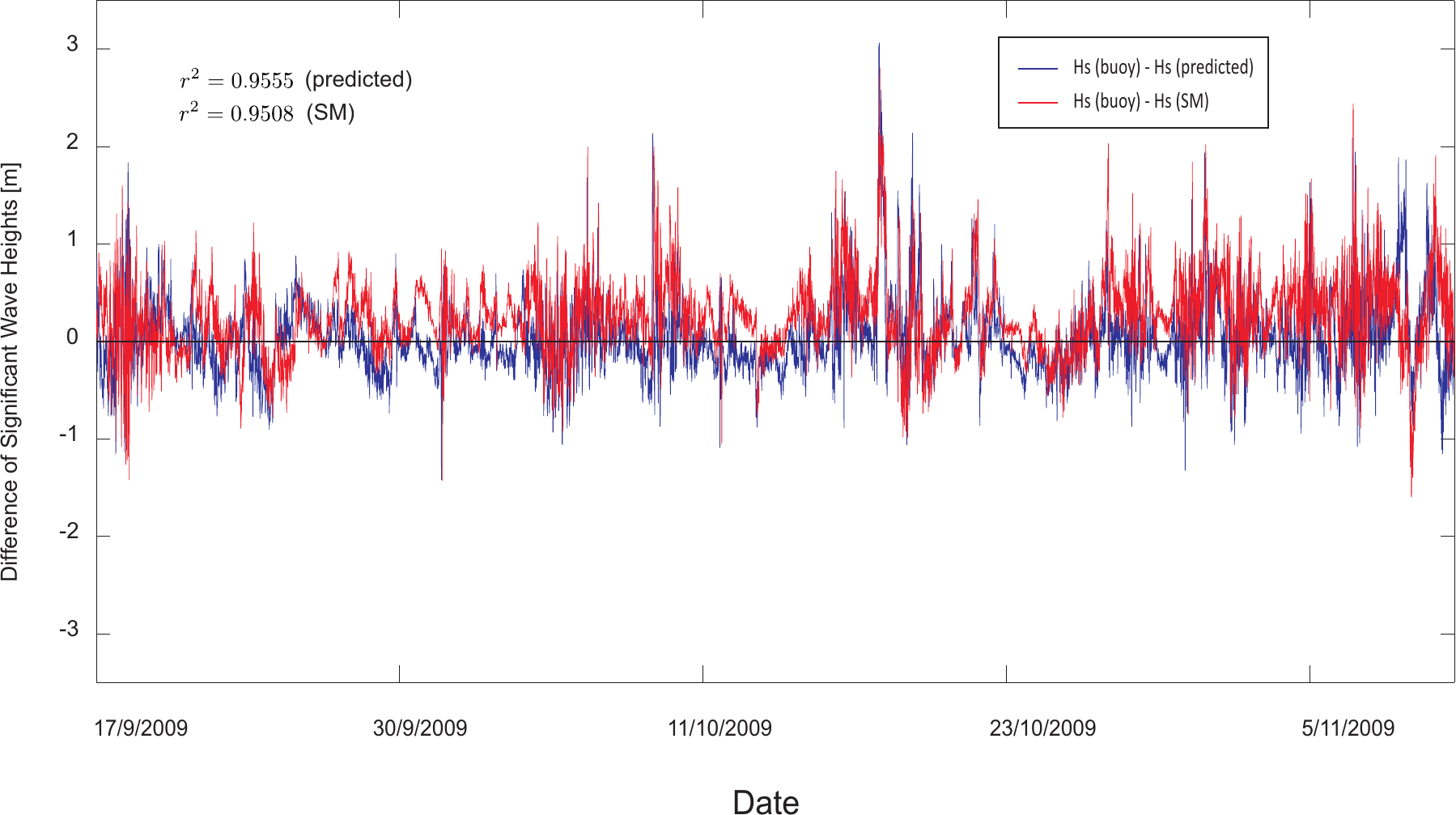}}
    \subfigure[]{\includegraphics[width=0.7\textwidth]{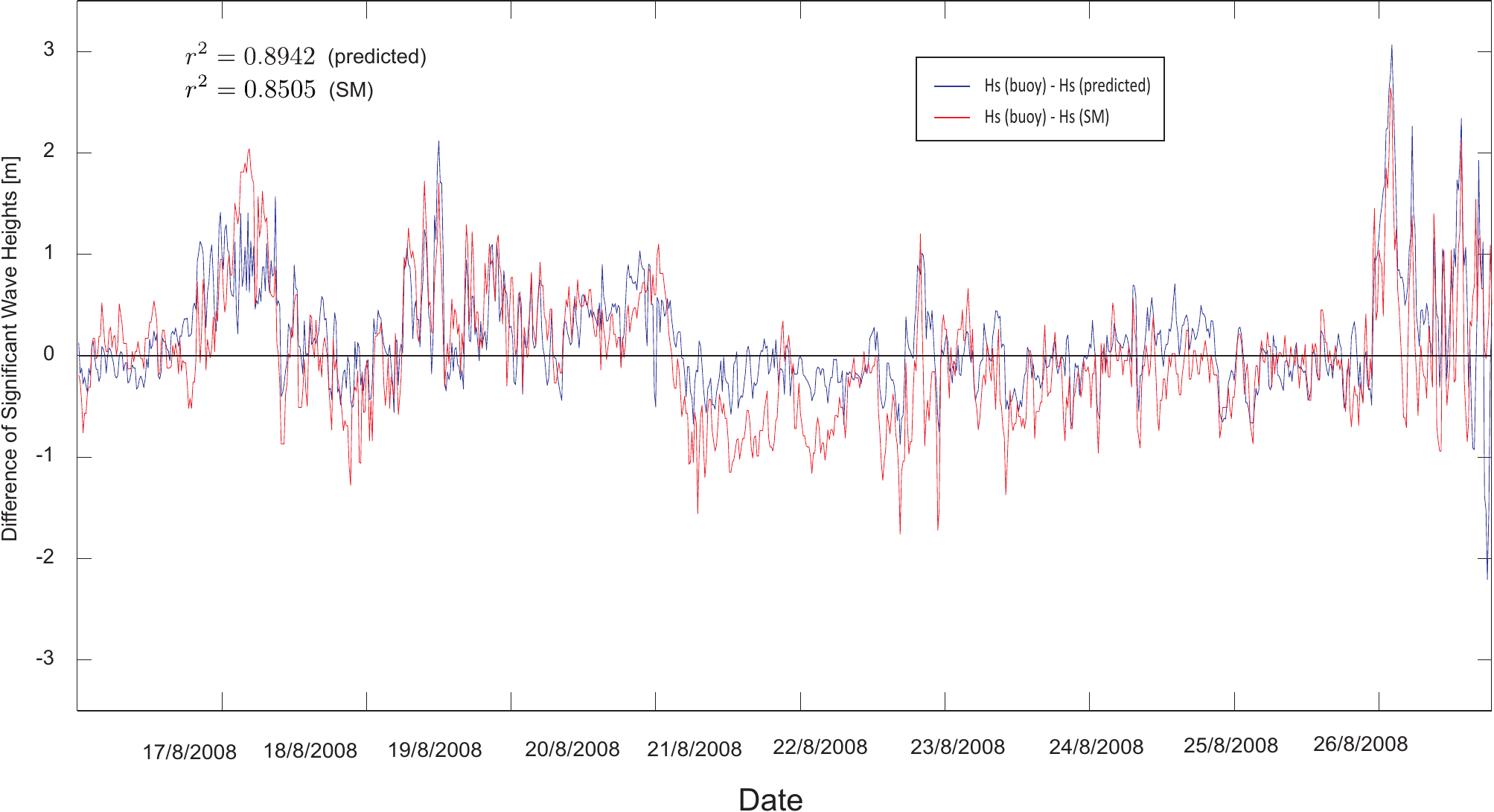}}
  \end{center}
  \caption{\label{Diferencias}Difference between measured and predicted $H_s$ with the SVR and SM in the locations considered; (a) Fino~1; (b) Ekofisk; (c) Glas Dowr.}
\end{figure}

As it can be seen, the performance of the SVR in Fino~1 and Ekofisk platforms databases is extremely good, following the trend and getting all the peaks in $H_s$. The performance in Glas Dowr is poorer, as previously reported. The SVR is able to catch the trend in $H_s$, but the reconstruction is not so accurate as in the other platforms. Note that the trend in $H_s$ is mainly due to storms occurred in the zone, so it is easy to see that the SVR is able to catch the behavior of the $H_s$ during these storms. The fact that the number of training samples is low in this platform seems to explain part of the poorer behavior of the SVR respect to the other measuring stations considered, as stated above. An additional analysis of the performance the proposed SVR-based method can better explain the SVR poor performance at Glas Dowr platform. The analysis is based on the calculation of bivariate histograms of the relative error in the $H_s$ estimation  ($[\hat{H_s} - H_s(\mbox{buoy})] / H_s(\mbox{buoy})$) with the corresponding significant wave steepness $\epsilon_s$ derived from the buoy data. The histograms for the three measuring stations considered in this work are shown in Figure~\ref{Histogramas}.

\begin{figure}[ht]	
  \begin{center}
  \includegraphics[width=0.96\textwidth]{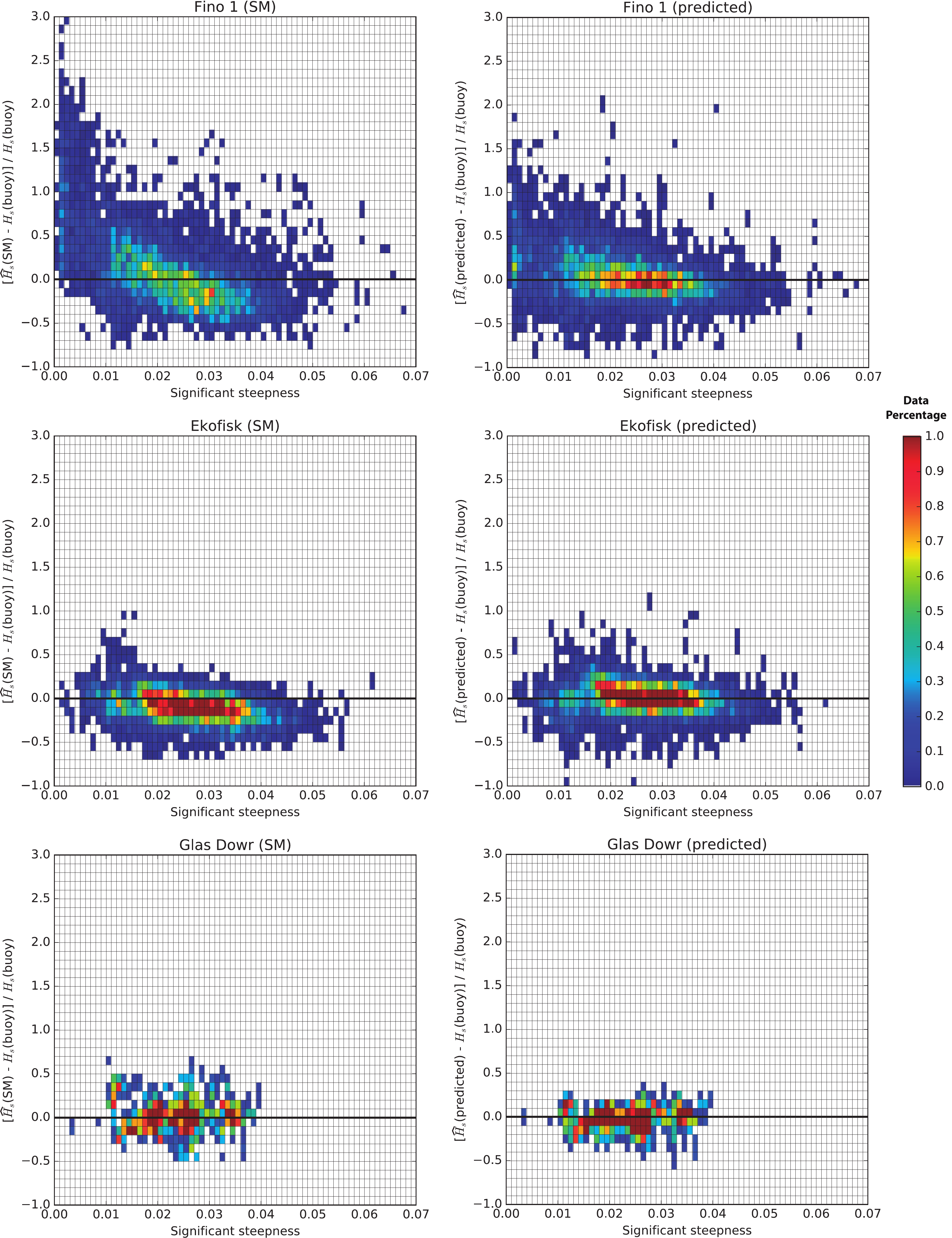}
  \end{center}
  \caption{\label{Histogramas} Bivariate histograms of significant wave steepness vs. relative error in the $H_s$ estimation for each measuring station: Fino~1 (top), Ekofisk (middle), and Glas Dowr (down). The results derived from the SM are plotted on the left, and the corresponding results from SVR (predicted) are located on le right part of the image. The color bar indicates the percentage of total data within the histogram for each case.}
\end{figure}

This figure shows the histograms for each measuring station and each $H_s$ estimation method (i.e. the results derived from SM in the left part of Figure~\ref{Histogramas}, and the results derived from SVR on the right part of that figure).
From these results, the following conclusions can be extracted separately for each station:

\begin{itemize}
\item Fino~1: These results are shown in the upper part of Figure~\ref{Histogramas}. The Fino~1 measurements cover a wider range of different sea state conditions than the other two stations. It can be seen that SM presents a higher scatter than SVR (predicted) results in the estimation of $H_s$. Furthermore, the SVR results present a higher percentage of data closer to the zero relative error than SM. In addition, it can be seen that, in most of the cases, the points where SM presents worse estimations of $H_s$ correspond to low values of $\epsilon_s$. It can be seen that in those cases SM overestimates more $H_s$.

\item Ekofisk: The results obtained for this station appear in the middle part of Figure~\ref{Histogramas}. It can be seen that, although SVR does not induce an improvement in the scatter presented in the histogram, the higher percentage of data are closer to the zero value of the relative error than SM, which has a higher bias than SVR.

\item Glas Dowr: The results corresponding to this station are shown in the lower part of Figure~\ref{Histogramas}. These data do not contain so higher values of $\epsilon_s$ than the other measuring stations. In addition, the scatter is reduced with SVR comparing with SM and there are a higher percentage of data closer to the zero value of the relative error for the SVR results.	
\end{itemize}

It is well known that swell presents in many cases smaller wave steepness than wind sea, because of the longer wave length that swell wave spectra normally contain. This indicates that SM should fail more when the wave steepness is low, but this is an implication only in one sense, i.e. it does not mean that for all the cases where the steepness is low, SM should fail. For example, in some cases of a very young wind sea, the steepness could be low (because the low values of $H_s$), but the roughness on the sea surface may be enough to get a proper value of $SNR$ for the $H_s$ estimation.
In addition, the wave steepness affects the radar imagery mechanisms, i.e. due to the effect of the tilt modulation caused by the wave slopes \cite{Alpers81,Bahar83,Feindt86,Schroeter86,West89,Ziemer94}.

\section{Conclusions}\label{sc:Conclusions}
In this work, a method for obtaining $H_s$ estimations from non-coherent X-band marine radars images has been presented.
This method is based on the use of the SVR methodology, for implementing a non-linear function that relates some selected input parameters with an objective value of $H_s$. After analyzing the results achieved by the SVR-based method and comparing them with the ones achieved by a SM, which is commonly used for $H_s$ estimation from non-coherent X-band marine radars, it can be observed that the proposed method presents better results reducing the scatter of the $H_s$ estimation. Hence, SVR method is able to outperform the SM by reducing the MSE error and increasing the correlation coefficient of the $H_s$ time series. Similar performances are achieved for the different platforms, which indicates that the performances presented here can be maintained for new data sets processed in the future for the same platforms.

\chapter{Efficient prediction of low-visibility events at airports}\label{cap:fog}
\section{Introduction}
According to the World Meteorological Organization \cite{WMO11}, fog is defined as the reduction in horizontal visibility to less than 1000 m. When the observed horizontal visibility is at least 1000 m, but not more than 5000 m, the phenomenon is called {\em mist}. Fog is typically classified according to the physical process that produces saturation or near-saturation of the air, such as strong evaporation, rain/post-frontal fogs or radiation and advective phenomena. The occurrence of fog and mist impact on a wide variety of human activities. Among them, air transportation is probably one the most affected sectors: foggy days dramatically restrict airport activities and cause flight delays, diversions and cancellations \cite{Rebollo14}, or accidents in the worst cases \cite{Ahmed14}. According to \cite{Bergot07}, the landing and take-off capacity at Paris-Charles de Gaulle International Airport is reduced by a factor of two during low-visibility conditions.

To aid the aeronautical community in dealing with low-visibility conditions at airports, meteorological services prepare terminal aerodrome forecasts for the local area around an airfield in accordance with the regulations of the International Civil Aviation Organization, specifically those provided in Annex 3 to the Convention on International Civil Aviation \cite{Jeppesen15}. Although terminal aerodrome forecasts are tailored to the needs of airlines and aircraft operators for flight planning (both pre-flight and intra-flight), they also assist air traffic and airport managers, for example, in activating specific procedures for ensuring safe operations during mist or fog conditions. Forecasting low-visibility conditions is frequently a difficult task requiring both knowledge of the meteorological causes of mist or fog formation, and a thorough awareness of the local topography. Consequently, aeronautical meteorological forecasters integrate different sources of information, such as observations, numerical weather prediction and other guidance tools to make a final robust decision on low-visibility forecasts. Hence, new techniques and methodologies are being developed to help forecasters improve the prediction of reduced-visibility events at airports facilities.

Numerical weather prediction is one of the most widely-used approaches by meteorological service providers for forecasting reduced-visibility conditions due to fog at airports. The most common procedure is to analyze the outputs of three-dimensional global models, such as the Global Forecast System from the National Oceanic and Atmospheric Administration \cite{Kanamitsu91}, the Integrated Forecasting System from the European Centre for Medium-Range Weather Forecasts \cite{Simmons89}, or mesoscale models, such as the Weather Research and Forecasting model \cite{Skamarock08} and the High-Resolution Limited-Area Model \cite{Unden02}. Nevertheless, as stated by many authors \cite{Van-der-Velde10,Zhou11,Roman-Gascon12,Steeneveld15}, the forecasting of fog events by numerical weather prediction is particularly difficult, in part because fog formation is extremely sensitive to small-scale variations of atmospheric variables, such as wind-shifts or changes in the low-level stability. One of the most significant drawbacks is the extremely high vertical resolution required to accurately simulate fog formation in the lower boundary layer \cite{Herman16}. To help overcome this problem, several single-column models have been developed for the forecasting of fog events \cite{Bergot94,Duynkerke98,Bott02,Terradellas06}, with a higher vertical grid resolution and a more comprehensive description of cloud microphysical processes. Some authors have also combined single-column models with three-dimensional models to provide a detailed numerical simulation of the thermo-hydrodynamic state of the atmosphere \cite{Bartok12,Fedorova13}.

Other research topics related to numerical weather prediction focus on the development of ensemble-prediction systems, since small differences in either the initial conditions, or in the model itself, increase and become significantly large after a certain time increment due to the chaotic and highly nonlinear nature of the atmospheric system \cite{Lorentz65}. Thus, ensemble prediction systems account for the uncertainty in weather forecasts, where, for example, an ensemble prediction combined with the Weather Research and Forecasting model was developed to forecast fog events in 13 cities in East China \cite{Zhou10}. Although some interesting results have been achieved with numerical models and ensemble prediction systems, the necessary computational and human resources, knowledge and facilities require major investments beyond that available.

Another interesting research topic consists of using statistical methods. One of the first such attempts was the use of linear regression \cite{Koziara83}. Subsequently, ANNs have also been used with statistical methods for fog prediction due to their capacity in dealing with complex nonlinear interactions among input and objective variables, and their performance in forecasting fog events based on more accessible observational variables. As an example, \cite{Fabbian07} successfully assessed the ability of a MLP with a back-propagation training algorithm to forecast fog events at Canberra International Airport. More recently, \cite{Dutta15} obtained good results with a MLP with a back-propagation learning technique to forecast 3-h visibility intervals during winter at Kolkata airport (India).  \cite{Colabone15} also used a very similar artificial neural network to predict the occurrence of fog events at the Academia da For\c{c}a A\'erea (Brasil). Other alternative artificial-intelligence techniques, such as FL \cite{Miao12} or Bayesian decision networks \cite{Boneh15}, have been applied to forecast low-visibility conditions.

Here, the performance of different machine-learning-based regressors are examined in forecasting low-visibility conditions at airports, and propose state-of-the-art regressors, which, to our knowledge, have not been previously applied to the problem of hourly forecasting of low-visibility events in terms of the runway visual range at airports. SVRs, ELMs and GPs have been evaluated in the prediction of low-visibility events at the Valladolid airport (Spain). Also the extent to which the atmospheric input variables may be pre-processed with the wavelet transform is investigated, and show how the machine-learning-based regressors obtain excellent results in the prediction of low-visibility conditions at airport facilities.

In the following section, the specific problem, including the processing of the relevant predictive variables is described. Section \ref{Methods} presents the main characteristics of the regression methods adapted to low-visibility prediction, as well as the wavelet methodology for data pre-processing. Section \ref{Experiments} details experimental results obtained in the prediction of low visibility at the Valladolid airport during several winter months. Finally, Section \ref{sc:Conclusions} provides some final remarks concerning this research and future work.

\section{Predictive Data and Objective Variables}\label{sc:UsedData}
The prediction of low-visibility events is considered at the Valladolid airport, Spain (41.70 N, 4.88 W), shown in Figure \ref{mapa}, which is the most important airport of the autonomous community of Castile-Leon in the ``Montes Torozos'' region (a very homogeneous and extensive area -- 800 km$^2$ -- on the northern plateau of the Iberian Peninsula), and is well-known for its foggy days. Due to the geographical and climatological characteristics of this area, radiation fog is by far the most frequent fog phenomenon \cite{Roman-Gascon16}, due partly to its proximity to the Duero river basin \cite{Morales94}. A detailed climatology for Valladolid airport of the most important aeronautical-meteorological variables can be found in \cite{AEMET12}, which analyzes, among other variables, the runway visual range for the period 1998--2011 to show November, December, January and February with the highest number of low-visibility events on average, while the summer months have the least.

\begin{figure}[ht]
\begin{center}
\includegraphics[draft=false, angle=0,width=13cm]{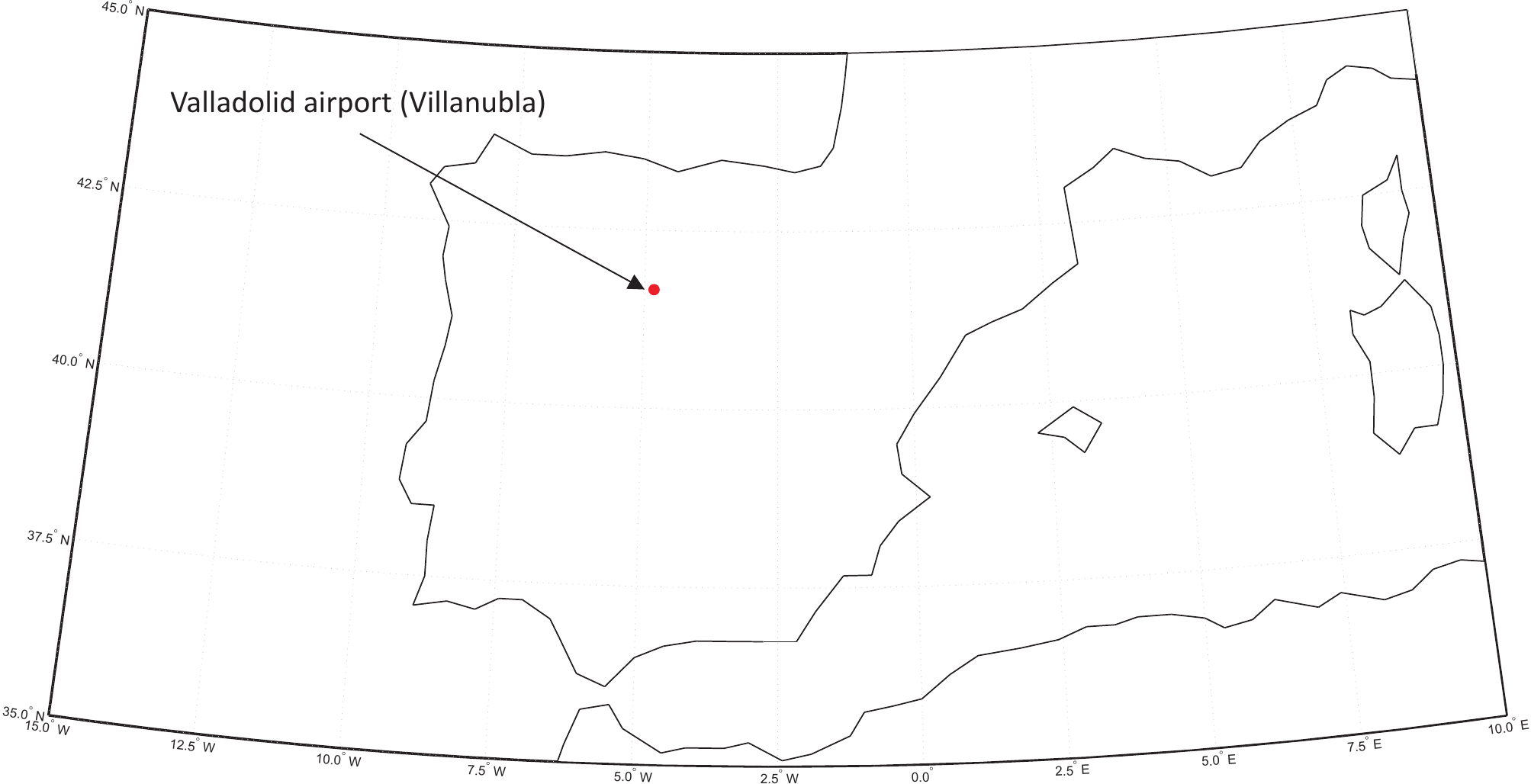}
\end{center}
\caption{ \label{mapa} Location of Valladolid airport (Villanubla), Spain, where the experiments to validate the proposed methodology for the prediction of low-visibility events have been carried out.}
\end{figure}

To study the occurrence of reduced-visibility conditions, it is used the data from a 100m meteorological tower located at the Research Centre for the Lower Atmosphere ``Jos\'e Casanova'' (CIBA), which is located about 13 km north-north-west of the airport. In situ information of the most basic parameters relevant to radiation fog at the airport is provided by meteorological data obtained from the two runway thresholds. The target variable is the runway visual range obtained  from three visibilimetres deployed along the airport runway (the touchdown zone, the mid-point and stop-end of the runway), which belong to the aeronautical observation network of the Meteorological State Agency of Spain. Note that these instruments are managed under a quality-management system certified by ISO 9001:2008, which guarantees measurement accuracy, and ensures the compliance of the measurements with international standards. It is also important to note that although meteorological airport reports (also commonly known as METAR reports) are prepared with human intervention at the Valladolid airport, they are not a good source of information, because the Valladolid airport is not a 24-h airport, which means observational information is lacking between 1930 and 0430 local time in summer, and from 2030 to 0530 local time in winter. The complete list of input and target variables considered are summarized in Table \ref{data}. Hourly data at the Valladolid airport from 2008 to 2013 is considered during the months when radiation fog is most intense according to \cite{AEMET12} (November, December, January and February).  The prediction time horizon has been set to 1 h, which requires successful prediction of the visibility at the airport 1 h later than the timestamp of input data (predictive variables), though experiments with a larger time horizon are also considered.

\begin{table}[ht]
\scriptsize{}
\begin{center}
\caption{\label{data} Data used in the study.} \vspace{0.3cm}
\resizebox{16cm}{!} {
\begin{tabular}{ccccc}
\hline
\vspace{0.1cm}

Variable         &Source       &Height above the ground (m)        &Units          &Instrument \\
\hline
\hline
Temperature      & CIBA      & 96.6, 35.5, 20.5, 10.5, 2.3         & $^{\circ}$C       & Riso P2448A and P2642A \\
                 & AEMET-Airport          & 2                      & $^{\circ}$C        & Vaisala HMP45D  \\
\hline
Relative Humidity/         & CIBA                & 97, 10                    &  \%          & Vaissala HMP45A \\
Dew point                 & AEMET-Airport          & 2                      & $^{\circ}$C        & Vaisala HMP45D \\
\hline
Wind speed       & CIBA      & 98.6, 74.6, 34.6, 9.6, 2.2          & m/s              & Riso P2548A \\
                 & AEMET-Airport          & 10                     &  m/s             & Vaisala WVA15  \\
\hline
Wind             & CIBA  & 98.6, 74.6, 34.6, 9.6, 2.2           & degrees true  & Riso P2021A   \\
direction        & AEMET-Airport          &  10                 & degrees true     & Vaisala WVA15  \\

\hline
Atmospheric      & CIBA                  & 2        &  hPa                          & Vaisala PA21\\
pressure         & AEMET-Airport         &2         & hPa                           & Vaisala PA21    \\

\hline
Runway visual range (target)       & AEMET-Airport         & 2        & m                             & Vaisala FD12  \\

\hline
\end{tabular}}
\end{center}
\end{table}

\section{Methods}\label{Methods}
The machine-learning regressors, including the SVRs, MLPs, ELMs and GP are used in this study, and they are also defined in Section \ref{sec:state_of_the_art}. Note that all the regressors considered are state-of-the-art methods in regression problems that have been demonstrated to give very good results in previous applications. Some general characteristics of the methods are well known: for example, the ELM is a very fast training algorithm, since it is based on random weights and a pseudo-inverse calculation. In contrast, the GP is usually the most computationally-demanding approach to be trained, and has shown poor performance for large data (though not the case here). While the MLP with the Levenberg-Marquardt training algorithm is a strong regression approach, it is computationally more demanding than the ELM. In terms of computational requirements in the training phase, the SVR is comparable to the MLP approach. The specific performances of these algorithms in the prediction of the runway visual range at Valladolid airport are detailed in Section \ref{Experiments}.

The description of the wavelet transform used is also included for pre-processing the input data in some of the experiments described below. Figure \ref{Ejemplo_Sist}a provides a general view of the proposed system structure, where the initial database is either directly passed to the regressors, or pre-processed with a wavelet methodology to give greater diversity in the variables. In any case, the considered regressors process this information to yield a final prediction. Figure \ref{Ejemplo_Sist}b shows the structure of the training/testing process. First, the initial database is split into training and testing datasets to train the regressors with the mathematical expressions given in Section \ref{sec:state_of_the_art}, and to evaluate the performances of the different methods, respectively.

\begin{figure}[ht]
\begin{center}
\subfigure[]{\includegraphics[draft=false, angle=0,width=13cm]{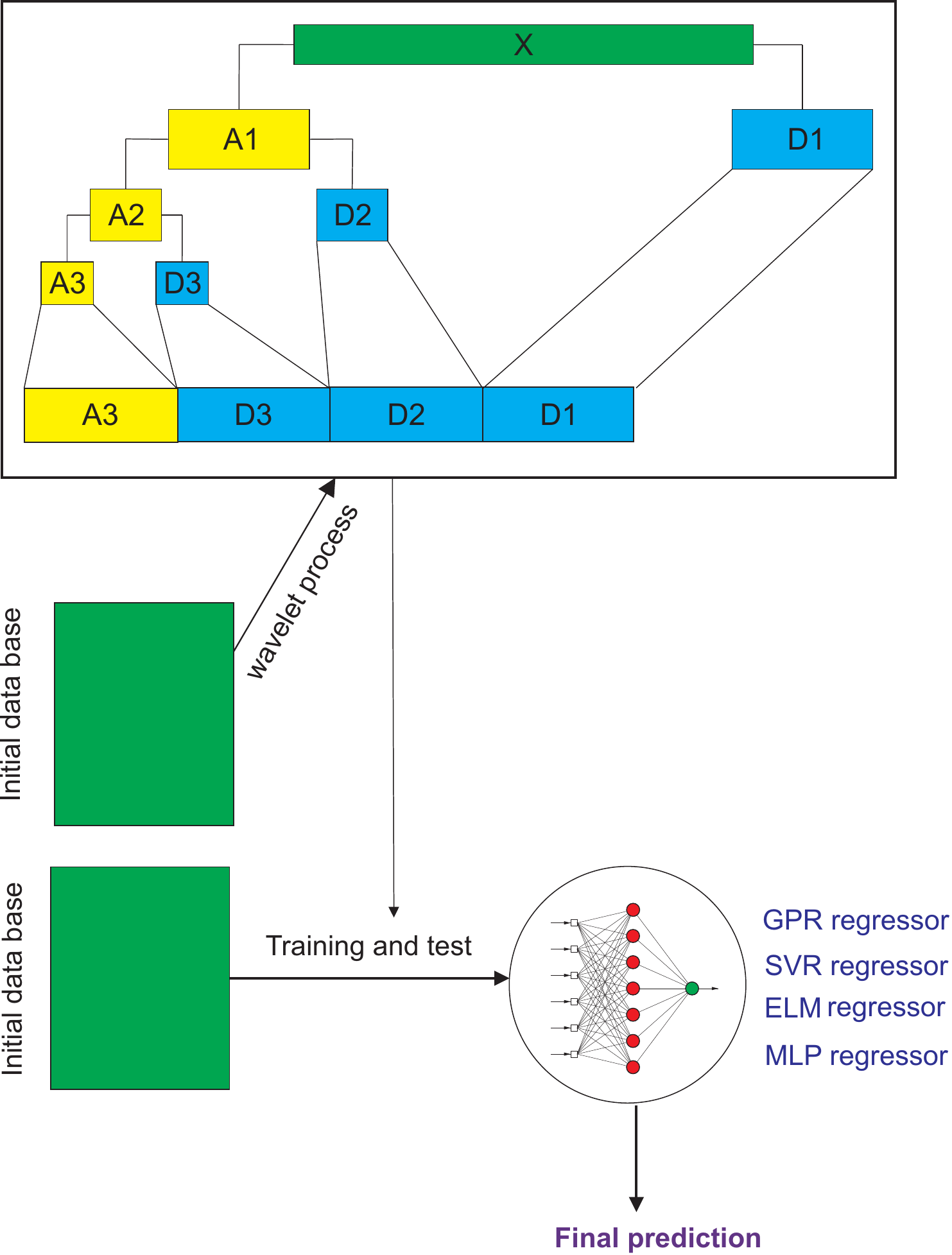}}
\subfigure[]{\includegraphics[draft=false, angle=0,width=10cm]{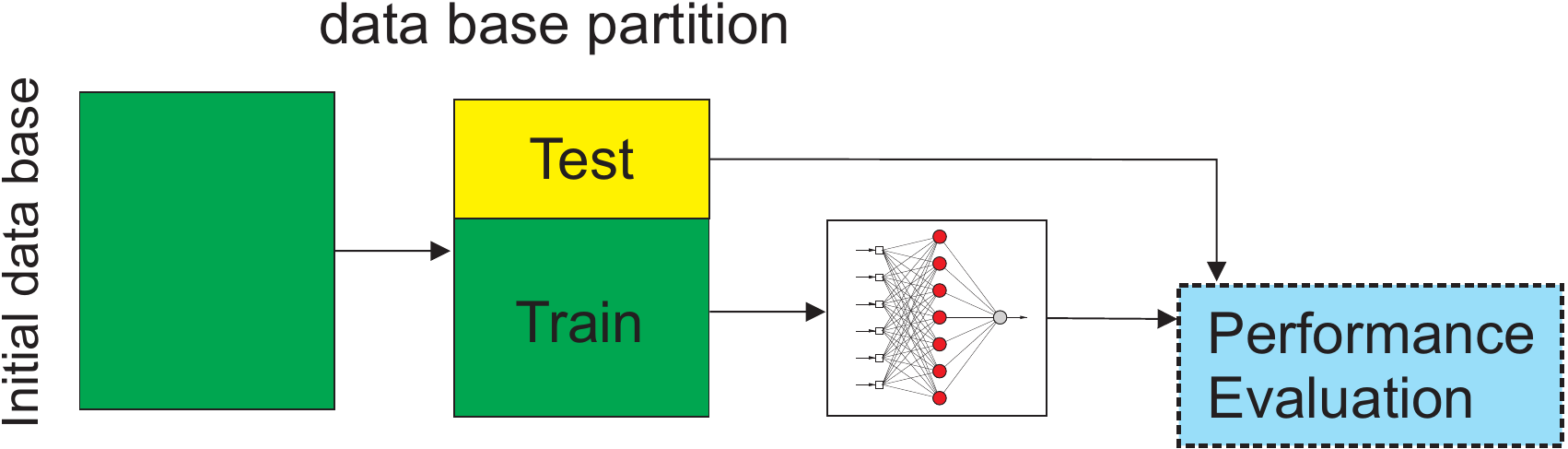}}
\end{center}
\caption{\label{Ejemplo_Sist} Example of the proposed prediction system with wavelet pre-processing for the test dataset and training step; (a) example of the prediction-system structure; (b) training phase and evaluation of the regressors.}
\end{figure}

The explanation of the wavelet transform will be explained directly below because the rest of the regression algorithms used in this study are explained in depth in chapter \ref{cap:introduction}.

\subsection{Discrete-Wavelet-Transformation Algorithm}\label{Wavelet}
In some regression problems, a specific pre-processing of the input data improves the performance of the regressors, such as the use of wavelet transforms \cite{Deo,Nourani}, whose basic aspects are outlined here. For further details, the interested reader may consult \cite{Mallat}.

For a continuous signal of interest $x(t)$, its wavelet transform is defined as \cite{Nourani}

\begin{equation}
T(a,b) = \frac{1}{\sqrt{a}}\int_{x=-\infty}^{x=+\infty}{g^*\left(\frac{t-b}{a}\right) x(t)\cdot dt},
\end{equation}
where $a$ is a scale factor, $b$ is the temporal translation of the function $g(t)$, $*$ denotes the complex conjugate, and $g(t)$ is the mother
wavelet transform. As input data are usually composed of discrete values $x_h$, it is necessary to use the discrete-wavelet transform to decompose the signal, for which the mother wavelet transform has the expression

\begin{equation}
g_{kl}(t) = \frac{1}{\sqrt{a_0^k}}\left(\frac{t-lb_0a_0^k}{a_0^k}\right),
\end{equation}
where $a_0$ is the specified fine-dilation step equal to 2 in most cases, $b_0$ is the location parameter set to 1 in most cases,
and $k$ and $l$ are integers that control the wavelet dilation and location, respectively. The discrete-wavelet transform usually considers the values of $a_0$ and $b_0$ based on powers of two. The mother wavelet in compact notation \cite{Mallat} is

\begin{equation}
g_{kl} = 2^{-k/2}g\left(2^{-k}h-l\right),
\end{equation}

whereby the wavelet coefficients with a scale $a = 2^k$ and location $b = 2^{k}l$ are written as

\begin{equation}
T_{kl} = 2^{-k/2}\sum_{i=0}^{N-1}{g\left(2^{-k}i-l\right)}x_h,
\end{equation}
where $x_h$ is the finite time series of interest, $i = 0, 1, 2, \ldots, N - 1$ and $N$ is an integer power of 2, i.e., $N = 2M$. Thus, the inverse discrete-wavelet transform (the reconstruction of the function $x_h$) is given by \cite{Nourani}

\begin{equation}
x_h = \bar{T}+\sum_{k=0}^{M}{\sum_{l=0}^{2^{M-m}-1}{T_{kl}2^{-k/2}}g\left(2^{-k}i-l\right)} = \bar{T} + \sum_{k=0}^{M}{W_{k}(t)},
\end{equation}
where $\bar{T}$ is the approximation sub-series at level $M$, and $W_{k}(t)$ is the detail of the sub-series at levels $k = 1, 2, \ldots, M$.
In this case, one level of approximation and three levels of detail sub-series for each predictor is considered.

Note that because of the wavelet coefficients, it is possible to analyze some details of the frequencies contained in the signal of interest in terms of the large scale (approximation) or small scale (detailed), resulting in a powerful pre-processing scheme in which different sub-series $W_k$ are generated and used to increase the information at the input of the prediction system. In this case, the wavelet pre-processing of the predictive variables for the problem of low-visibility prediction is carried out by specific functions contained in a Matlab toolbox \cite{Matlab}. Figure \ref{wavelet_process} shows an example of the signal decomposition into different sub-series (approximation and detailed parts). The wavelet transform is applied to the input data (predictive variables) to decompose the signal into approximation and detailed parts. An example is given below for a given variable of the prediction process (variable 1: air temperature at 96.6 m), which is carried out several times by applying the wavelet transform again to the approximation part resulting from the previous step, as seen in Figure \ref{wavelet_process}.

\begin{figure}[ht]
\begin{center}
\includegraphics[draft=false, angle=0,width=13cm]{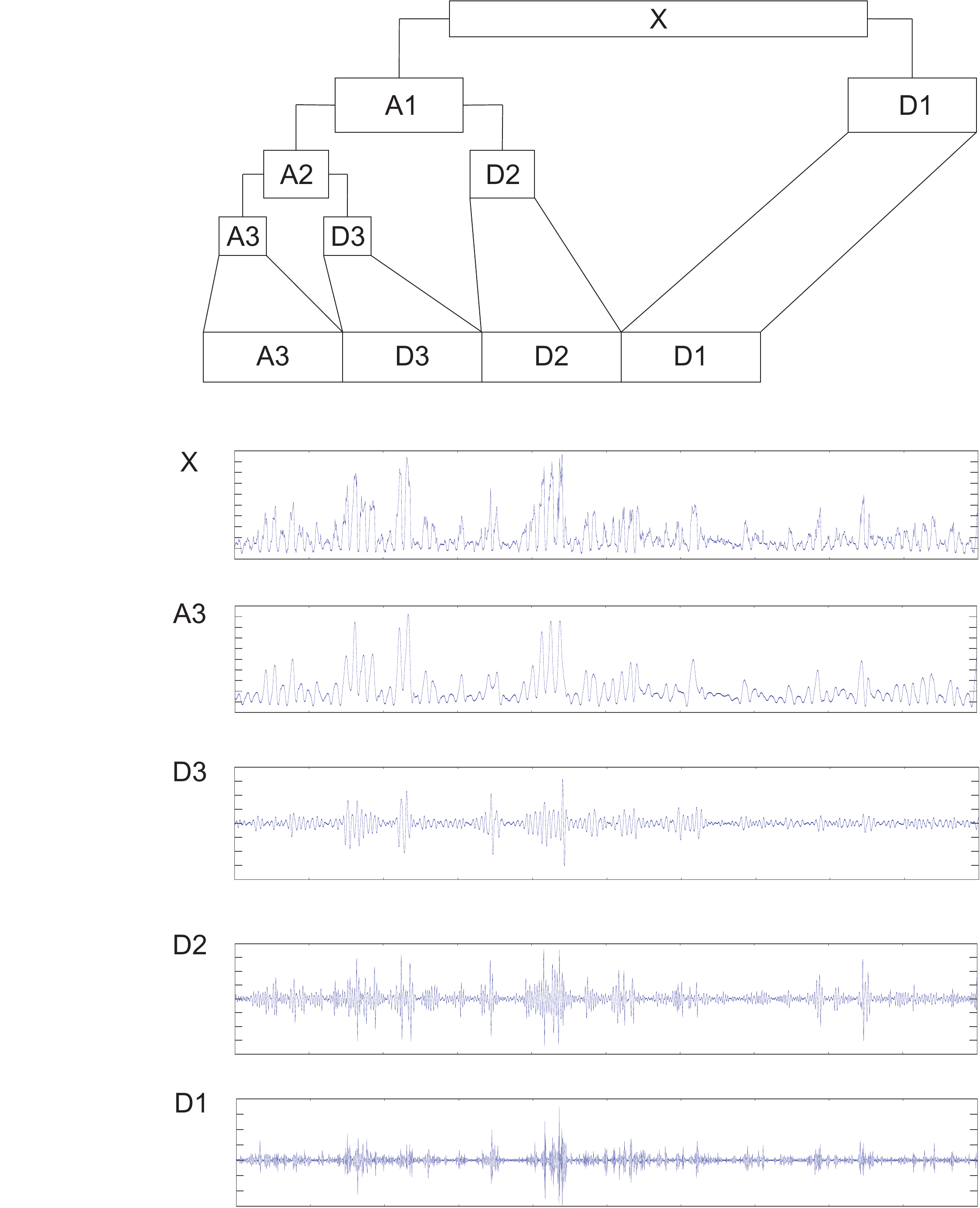}
\end{center}
\caption{ \label{wavelet_process} Decomposition of a signal of interest into approximation and detailed parts using a wavelet transform. In this case, the signals of interest are the predictive variables for the low-visibility prediction.}
\end{figure}

\section{Results}\label{Experiments}
Here it is presented the results obtained by the different regression algorithms in estimating the runway visual range at Valladolid airport. To provide an additional baseline for comparison, in addition to the several regressors already considered, the results are compared in terms of the root-mean-square error ($RMSE$) with that of the persistence model ($RMSE_{\mathcal{P}}$), for which a skill score with a persistence model as a reference is defined as

\begin{equation}
RMSE_{ss}=1-\frac{RMSE_{\mathcal{M}}}{RMSE_{\mathcal{P}}},
\end{equation}
where $\mathcal{M}$ is the error (RMSE) of the forecast of the proposed methods, and $\mathcal{P}$ is the RMSE of the persistence model. Note that the closer $RMSE_{ss}$ is to zero, the more similar the model $\mathcal{M}$ is to the persistence. Note that positive values of $RMSE_{ss}$ means the model $\mathcal{M}$ outperforms the persistence (baseline model for this problem), whereas negative values of $RMSE_{ss}$ imply the persistence is a better model for prediction than $\mathcal{M}$.

First the importance of including wavelets is considered, as well as the external variables from the CIBA meteorological tower, for the assessment of the algorithms with and without these variables. Table \ref{Comparativa} shows the best results obtained by the machine-learning approaches, both with and without the wavelet method, in which all the variables from the CIBA tower are included in terms of the $RMSE$, the $MAE$, correlation coefficient ($r^2$) and the skill score ($RMSE_{ss}$). In all cases, each algorithm is trained using 80\% of the data for train and 20\% for test, both sets randomly chosen. Without the wavelet method (left-hand column), it can be observed that the GP regression provides the best results for all the performance metrics, with a $RMSE_{ss}=0.6$ better than that of the MLP (second best), where $RMSE_{ss}=0.5$. In contrast, the ELM and SVR show poorer results than the MLP and GP, where the SVR delivers the worst error values. Regarding the prediction of low-visibility events including a pre-processing step with the wavelet method (right-hand column), the results of the ELM are worsened for all the metrics, which, however, remain fairly constant with the exception of the $r^2$ metric, which increases slightly, but not enough to improve the skill score. A significant improvement in the MLP and the GP is detected when applying pre-processing with wavelets, where all performance metrics improve similarly in both cases.

\begin{table}[ht]
\begin{center}
\caption{\label{Comparativa} Comparison of the best results (10 runs of the algorithms in different sets) for the estimation of low-visibility events (in terms of the runway visual range at the airport) by the ELM, SVR, MLP and GP, with and without wavelet pre-processing. The variance is given in parentheses.} \vspace{0.3cm}
\resizebox{16cm}{!}{
\begin{tabular}{lcccccccccrcrrrr}
\hline
& \multicolumn{5}{c}{Without Wavelet method}  & &\multicolumn{2}{c}{Wavelet method}\\
\cline{2-5}   \cline{7-10}
      &$RMSE$ [m]    &$MAE$ [m]      &$r^2$  &$RMSE_{ss}$  && $RMSE$ [m]   &$MAE$ [m]    &$r^2$  &$RMSE_{ss}$ \\
\hline
ELM\  &262.1(108.8)   &137.8(27.6)   &0.5  &0.5   &&286.9(45.0)  &155.0(6.5)  &0.4  &0.4\\
SVR\  &369.5(0.2)   &104.7(0.0)   &0.1  &0.3   &&365.8(0.0)  &105.0(0.0)  &0.2  &0.3\\
MLP\   &238.1(3.7)   &87.0(31.2)   &0.6  &0.5   &&204.8(2.7)  &70.2(26.1)   &0.7  &0.6\\
GP\  &207.8(0.0)   &88.0(0.0)    &0.6  &0.6   &&\textbf{185.7(0.0)}  &\textbf{71.7(0.0)}   &\textbf{0.8}  &\textbf{0.6}\\
\hline
\end{tabular}}
\end{center}
\end{table}

The $MAE$ is reduced by over 16 m in both approaches, which improves their accuracy. Similarly, the $RMSE$ is reduced, which improves the sensitivity of the MLP and GP to large errors. The $r^2$ for both methods shows the same improvement trend, leading to better results than in the case without the wavelet pre-processing method. The skill score improves from $RMSE_{ss}=0.5$ to $RMSE_{ss}=0.6$ for the MLP, and from $RMSE_{ss}=0.59$ to $RMSE_{ss}=0.63$ for the GP. Therefore, based on these the latter results, it can be concluded that the MLP and GP with the wavelet pre-processing step improve the prediction of low-visibility events than the other regression methods.

To complete this first aspect of the results, the performance of the different regressors is compared with wavelet pre-processing in Table \ref{Comparativa_sin_ciba} without the CIBA-tower variables. Note that the performance of all the methods is clearly affected by the removal of the CIBA-tower variables for all metrics. With respect to the skill score, note that the best algorithm (GP) worsens from $RMSE_{ss}=0.6$ to $RMSE_{ss}=0.5$, with the other algorithms suffering a similar performance when the CIBA-tower variables are removed, demonstrating the importance of considering these variables in predicting the runway visual range.

\begin{table}[ht]
\begin{center}
\caption{\label{Comparativa_sin_ciba} Comparison of the best results for the estimation of low-visibility events (in terms of the runway visual range at the airport) by the ELM, SVR, MLP and GP, for the wavelet pre-processing case, and without CIBA features.} \vspace{0.3cm}
\begin{tabular}{lcccccccccccccc}
\hline
      &$RMSE$ [m]    &$MAE$ [m]  &$r^2$  &$RMSE_{ss}$  \\
\hline
ELM\  &304.0  &163.6  &0.4 &0.4  \\
SVR\  &372.7  &105.6  &0.1 &0.3 \\
MLP\  &259.8   &104.3   &0.5  &0.5   \\
GP\  &260.5   &116.3   &0.5  &0.5   \\
\hline
\end{tabular}
\end{center}
\end{table}

The effect of considering different prediction time horizons (from 1 h to 4 h) is shown in Figure \ref{Time_h_pred}, with and without the CIBA-tower variables, where the performance of all the considered regressors is slightly affected by increasing the prediction time horizon, and thereby obtaining worse results. In the case without CIBA-tower variables, it is possible to see the degree to which the performance of the GP and the MLP are similar for all the prediction time horizons. The inclusion of the CIBA-tower variables improves the GP over the MLP, which indicates that the GP takes superior advantage of the information provided by these variables.

\begin{figure}[ht]
\begin{center}
\includegraphics[draft=false, angle=0,width=13cm]{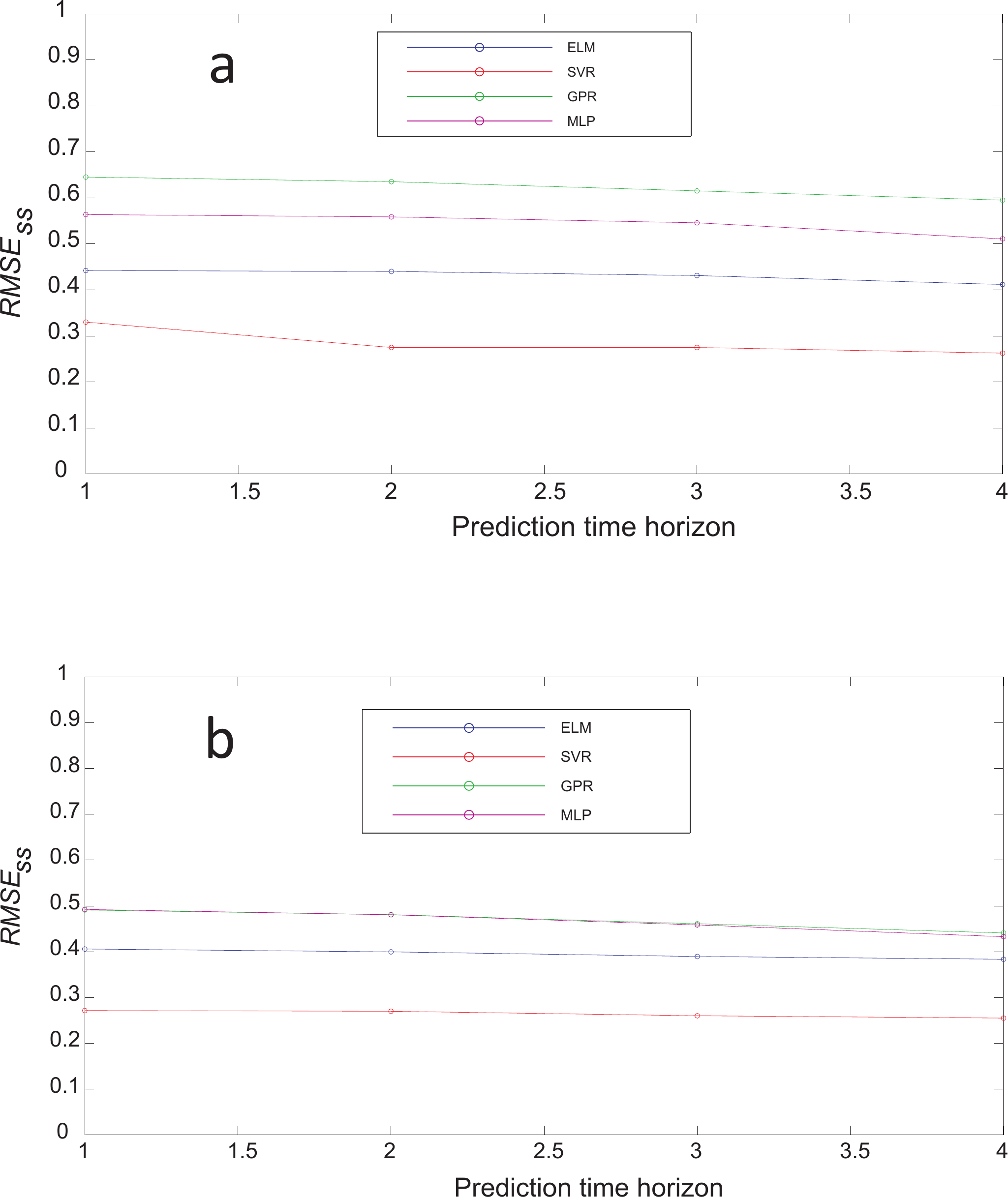}
\end{center}
\caption{\label{Time_h_pred}Skill score $RMSE_{ss}$ of the different regressors considered at different time-horizons for the prediction of low visibility; (a) with the CIBA tower; (b) without the CIBA tower.}
\end{figure}

Figures \ref{Prec_NN} and \ref{Prec_GPR} show two examples of the prediction of low-visibility events by the best algorithms tested including a wavelet pre-processing, where the prediction of the runway visual range and a normalized scatter plot are shown. Note that the MLP over-estimates the runway visual range in very low visibility conditions, whereas the GP gives a more accurate prediction, even in situations of very low visibility without a clear over-estimation of the runway visual range. Hence, it is this exceptional skill of the GP with wavelet pre-processing that makes it the best option for implementation in short-term low-visibility prediction systems in support of air navigation and airport services.

\begin{figure}[ht]
\begin{center}
\includegraphics[draft=false, angle=0,width=13cm]{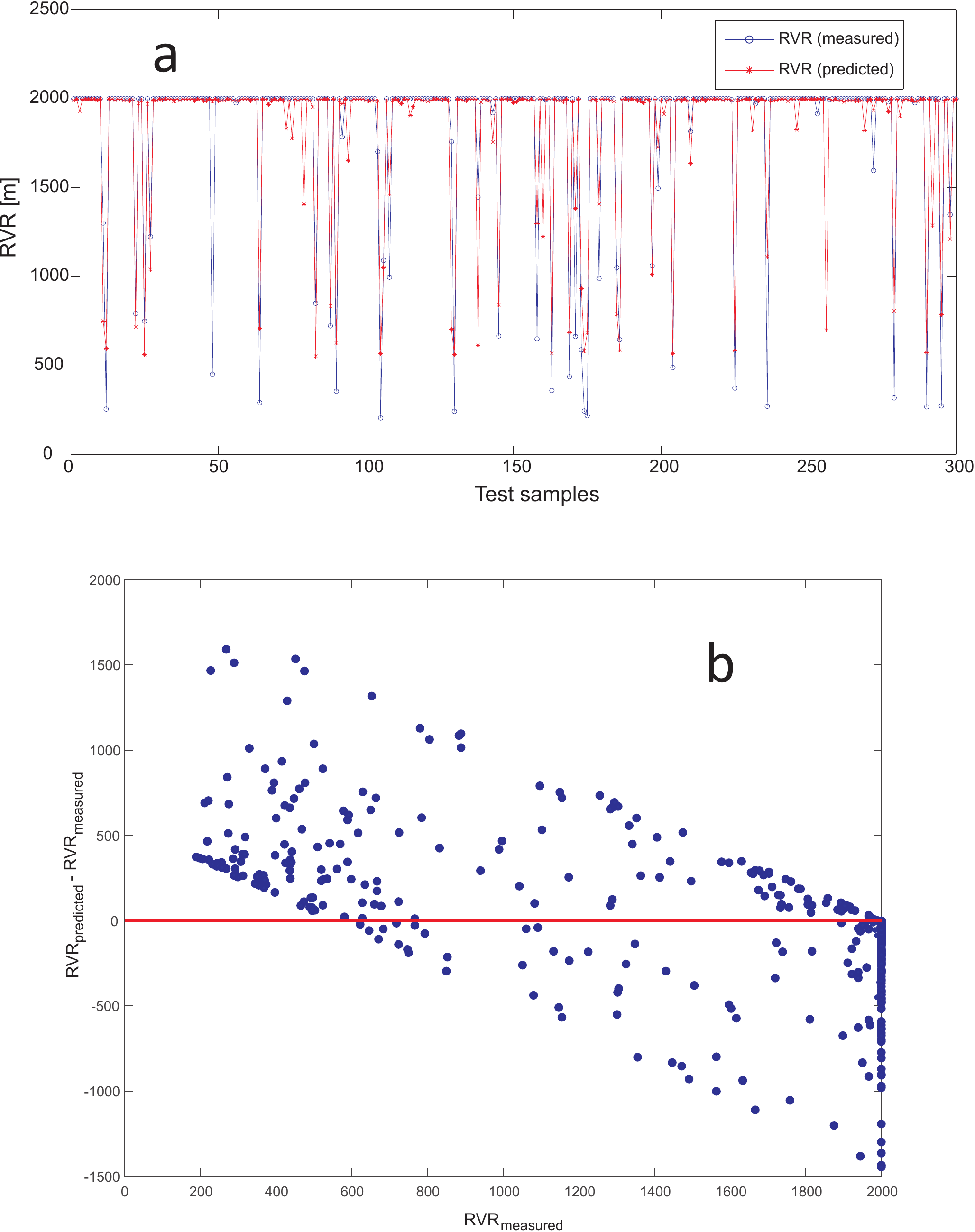}
\end{center}
\caption{\label{Prec_NN}Prediction of low-visibility events at Valladolid airport by the MLP approach with wavelet pre-processing for the test dataset; (a) direct prediction of the runway visual range (temporal); (b) Normalized scatter plot.}
\end{figure}

\begin{figure}[ht]
\begin{center}
\subfigure[]{\includegraphics[draft=false, angle=0,width=13cm]{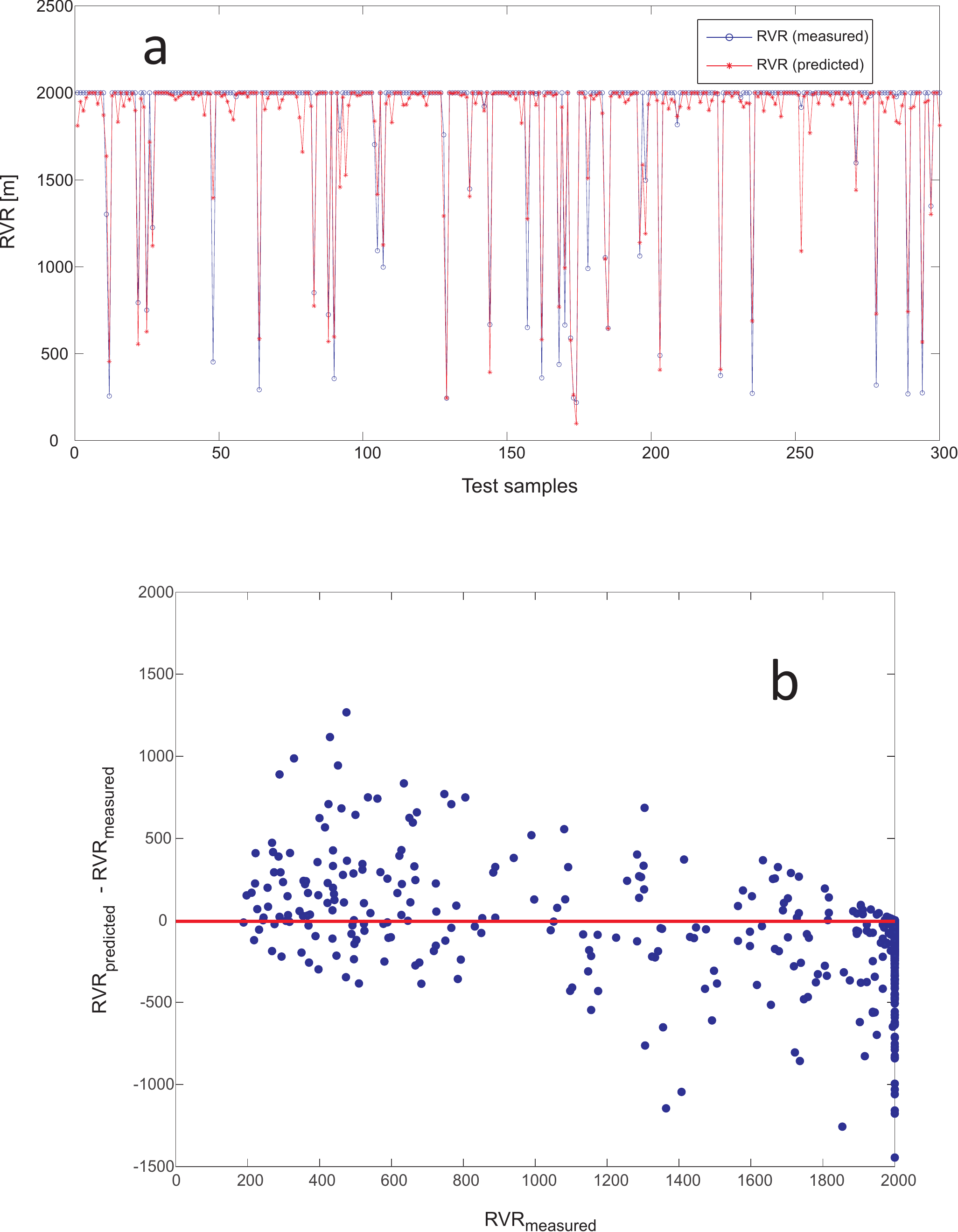}}
\end{center}
\caption{\label{Prec_GPR} Prediction of low-visibility events at Valladolid airport by the GP approach with wavelet pre-processing for the test dataset; (a) direct prediction of the runway visual range (temporal); (b) Normalized scatter plot.}
\end{figure}

More insight on the regressor performance is obtained by analyzing results when separating daytime (hours between sunrise and sunset) from nighttime (the remaining hours) cases. Since atmospheric conditions during the day and night are different in terms of boundary-layer stability, which conditions the degree of fog formation, the forecast time may impact the algorithm performance. Table \ref{Comparativa_diur_noct} shows results for day and night cases, which indicate the GP is better in terms of $RMSE_{ss}$ during the night, where $RMSE_{ss} = 0.6$ for the nighttime case, whereas $RMSE_{ss} = 0.5$ for the daytime. The other regressors seem to perform better in the nighttime case in terms of the $RMSE$, $MAE$ and $r^2$, though the skill score is quite similar for both the nighttime and daytime cases. Note that the GP is still the best performing algorithm, since it outperforms the other regressors for low-visibility events for both periods, indicating the superiority of the GP in the handling of the predictive variables compared with the other regressors.

\begin{table}[ht]
\begin{center}
\caption{\label{Comparativa_diur_noct} Comparison of the best results for the estimation of low-visibility events (in terms of the runway visual range at the airport) by the ELM, SVR, MLP and GP, for the wavelet pre-processing case, with nighttime and daytime samples.} \vspace{0.3cm}
\begin{tabular}{lcccccccccccccc}
\hline
      &$RMSE$ [m]    &$MAE$ [m]  &$r^2$  &$RMSE_{ss}$  \\
\hline
Day-time \\
\hline
ELM\  &281.9  &147.1  &0.4 &0.4   \\
SVR\  &373.0  &102.9  &0.2 &0.3   \\
MLP\  &215.1  &76.3   &0.6  &0.6   \\
GP\  &259.5  &121.6  &0.5  &0.5   \\
\hline
\hline
Night-time \\
\hline
ELM\  &271.4  &150.7  &0.5 &0.4   \\
SVR\  &360.3  &106.5  &0.3 &0.3   \\
MLP\  &203.9  &72.9   &0.7  &0.6   \\
GP\  &200.2  &87.6   &0.7  &0.6   \\
\hline
\end{tabular}
\end{center}
\end{table}

Algorithm performance in terms of classification accuracy is carried out by including thresholds at different runway visual ranges, whereby different procedures are triggered depending on the visibility conditions at the airport. The 1000-m threshold is considered because low-visibility procedures are activated at airports when the magnitude of the runway visual range is $\leq$ 1000 m. The 550-m and 300-m thresholds are also considered since they are the limits whereby category I and II precision-instrument approach and landing operations are performed, respectively. Table \ref{confusion_matrix} shows the confusion matrices obtained after applying different thresholds (for visibilities $<$ 1000~m, 550~m and 300~m) for the best algorithms (the MLP and GP), where the GP obtains better classification values than the MLP. The percentage of correct classification is $>$ 98\% (also for the MLP) in visibility situations over the 1000-m threshold, and over 80\% under the threshold (68\% for the MLP). When the visibility threshold is lower, the correct classification percentage is evidently lower, and the false negatives given by the algorithms are much more frequent. Note that for the 300-m threshold, the MLP is not able to correctly classify any low-visibility runway visual ranges. In contrast, the GP makes the correct classification in some cases, and thus is the most reliable method among those tested here for the prediction of low-visibility events at airports.

\begin{table}[ht]
\begin{center}
\caption{\label{confusion_matrix} Confusion matrix of the MLP and GP models with the wavelet method for 1000-m, 550-m and 300-m classification thresholds (Th).} \vspace{0.3cm}
\begin{tabular}{lccccccccccc}
\hline
& \multicolumn{2}{c}{Th. 1000m}  & &\multicolumn{2}{c}{Th. 550m}  & &\multicolumn{2}{c}{Th. 300m}\\
\cline{2-3}   \cline{5-6}  \cline{8-9}
      &Over Th.   &Under Th. &&Over Th.    &Under Th. &&Over Th.  &Under Th.\\
\hline
MLP\\
\hline
Over Th.\  &98.8\%  &1.2\%  &&55.6\%      &44.4\%      &&0\%     &100\%\\
Under Th.\  &31.8\%  &68.2\% &&4.0\%      &96.0\%      &&0\%     &100\%  \\
\hline
\hline
GP\\
\hline
Over Th.\  &98.8\%  &1.2\%  &&48.9\%      &51.1\%      &&34.5\%     &65.5\%\\
Under Th.\  &19.8\%  &80.2\% &&2.0\%      &98.0\%      &&0.8\%     &99.2\%  \\
\hline
\end{tabular}
\end{center}
\end{table}

\section{Conclusions}\label{sc:Conclusions}
A model for the prediction of low-visibility events at airport facilities based on machine-learning regression techniques is proposed. The performance of several state-of-the-art machine-learning regressors is examined for a real case study at the Valladolid airport (Spain). The input data are atmospheric variables obtained from local measurements at the airport, as well as a meteorological tower nearby. As the objective variable, the runway visual range at the airport is obtained from three visibilimetres deployed along the runways (the touchdown zone, the mid-point and stop-end of the runway). A study of the variables contributing the most to the prediction of low-visibility events is also carried out, together with the application of a wavelet transform to further exploit the information of the input variables. While excellent results in the prediction of low visibility at the study area are obtained with the proposed model, the method requires the use of an instrumented tower nearby. Since most airports are not equipped with such extra instrumentation, the applicability of the proposed machine-learning techniques may be limited. Therefore, future work is to evaluate the performance of alternative data sources concerning the vertical structure within the lower part of the boundary layer, such as conventional radio-soundings (ground-based), satellite-based atmospheric soundings or aircraft meteorological data relays. Additional research to study extremely low visibility is also required, since these situations have a greater impact on airport operations and aeronautical navigation than situations with merely reduced visibility. For example, extreme-event probability distributions or related techniques could be combined with machine-learning approaches.

\part{Final remarks and future research activities}\label{part:conclusiones}
\chapter*{Final remarks}
This Ph.D. thesis deals with the improvement of the optimization process in the exploitation of several renewable energies, as well as the study of the most important variables in meteorology whose influence is fundamental in the correct operation of facilities management in oceanic engineering and airports. The use of soft computing techniques, in particular neural approaches and EAs, are key in the development of the experiments carried out in this work.

From the results of the research activity developed within this work, several conclusions can be extracted, and they are summarized in this chapter.

\begin{itemize}
\item A hybrid prediction system for wave energy prediction has been proposed, and improved by means of a BO methodology. A FS method is applied to obtain the best features for the final prediction, the wave energy flux and significant wave height in the case under study. This procedure demonstrates that is possible to obtain good results without a high computational load. Once the selection process is done, the final prediction is carried out with ELM or SVR approach in order to compare the performance of both algorithms. In any case, the application of BO is able to improve the performance of the system. This improvement is related to the optimal selection of parameters carried out, being the increase of computational time the only inconvenient of the BO proposal. Nevertheless, this increase only affects the training phase and not the operation phase, in which predictions are made after training, therefore, this limitation is not an issue.

\item A hybrid approach is proposed for the prediction of WPREs in this thesis. In this case the combination of data from numerical-physical models (reanalysis) and state-of-the-art statistical ML regressors is proposed. The first contribution of this proposal is the use of the regressors to predict the WPREs, because these methods has not been previously applied directly to WPRE prediction. The second contribution is the use of direct reanalysis data as input (predictive) variables of the ML regression techniques. The results show good performance, especially those corresponding to ELM and GP approaches. In fact, GP exhibits the best results, outperforming clearly the rest of the ML regressors tested. Moreover, the use of reanalysis data is specially relevant in this problem, making easier the training of ML regressors since the ERA-Interim reanalysis provides robust meteorological variable estimation back to 1979, with high spatial and enough temporal resolution to tackle this problem.

\item A method for obtaining $H_s$ estimations from non-coherent X-band marine radars images has been presented as first contribution of the thesis to facilities management. After analyzing the results achieved by the SVR-based method and comparing them with the ones achieved by a standard method, which is commonly used for $H_s$ estimation from non-coherent X-band marine radars, it can be observed that the proposed method presents better results reducing the scatter of the $H_s$ estimation. The SVR methodology is able to increase the correlation coefficient of the $H_s$ time series.

\item A model for prediction of low-visibility at airports is finally presented in this thesis. The performance of several state-of-the-art ML regressors is examined for a real case study at the Valladolid airport (Spain). A study of the variables contributing the most to the prediction of low-visibility events is also carried out, together with the application of a wavelet transform to further exploit the information of the input variables. The method proposed requires the use of an instrumented tower nearby. Since many small airports are not equipped with such an extra instrumentation, the applicability of the proposed machine-learning techniques may be limited.

\end{itemize}

The results obtained in this research work have been presented at several international events and accepted or published in scientific publications in the Journal Citation Reports (JCR). In particular, during the last 3 years, 9 papers have been accepted for publication in relevant international journals, and other ones are currently under review. In addition, 6 papers have been presented in International conferences, and another one in a national congress. A complete list of the papers related to the research work performed in this Ph.D. thesis can be seen in \ref{apendix}.

\paginalimpia
\chapter*{Future research lines}
Despite the different results obtained from this Ph.D. thesis, there are several directions in which subsequent studies could progress. Some of the detected areas to be addressed in depth in near future are:

\begin{itemize}
\item Due to the generality of the approaches used in the majority of problems tackled in this work, the methodology used in the case of ocean wave features prediction can be extended to alternative prediction systems and other problems. Specially to hybrid approaches involving ML algorithms with a high number of parameters to be tuned.

\item In this work several approaches belonging to the state-of-the-art ML techniques are used. However, there are many advances in this field, for instance, the \ac{CNN}s. A CNN consists of a number of convolutional and subsampling layers optionally followed by fully connected layers. The input of a CNN used to be an image, therefore, the study of ocean wave parameters can be done by means of images of the wave's spectrum. The idea is to obtain the prediction of the main parameters ($H_s$, $T_m$, etc.) as of images which contain the spectrum in frequency of the wave data from oceanographic buoys.

\item In the case of low-visibility prediction at airports, since many small airports are not equipped with a nearby measuring tower, the applicability of the proposed machine-learning techniques may be limited. Therefore, future work is to evaluate the performance of alternative data sources concerning the vertical structure within the lower part of the boundary layer, such as conventional radio-soundings (ground-based), satellite-based atmospheric soundings or aircraft meteorological data relays. Additional research to study extremely low visibility is also required, since these situations have a greater impact on airport operations and aeronautical navigation than situations with merely reduced visibility. For example, extreme-event probability distributions or related techniques could be combined with machine-learning approaches.

\item Other research activities made during the Ph.D period which does not appear in this work, but which are published in journals, are related with the estimation of solar radiation by means of neuro-evolutionary hybrid mechanisms. The modelling system at daily forecast horizons could provide real-time energy utilization in power grids at a short-term temporal scale. However, a future study could validate the model for longer-term horizon, including seasonal scales that may enable energy experts in decision-making in relation to longer-term energy stainability projects. In addition, in real-time systems, the data behavior issues (e.g., non-stationarities, trends and jumps in input time-series) due to dynamical or stochastic nature of climate variables could also be considered to improve the model. There is opportunity to apply the model to some of the other solar-rich cities and regional sites (incorporating the universally available satellite data) to help enhance the practicality of the neuro-evolutionary wrapper methodology proposed in this study.
\end{itemize}

\part{Appendix}
\chapter*{Apendix A. List of publications} \label{apendix}
This section presents a summary of scientific publications obtained during the research in this thesis.

\section*{Papers in International Journals}
\begin{enumerate}
\item \textbf{L. Cornejo-Bueno}, J. C. Nieto Borge, E. Alexandre, K. Hessner, S. Salcedo-Sanz, ``Accurate Estimation of Significant Wave Height with Support Vector Regression Algorithms and Marine Radar Images'', Coastal Engineering, vol. 114, pp. 233-243, 2016 (JCR 2016: 3.221)

\item A. Aybar-Ruiz, S. Jim\'enez-Fern\'andez, \textbf{L. Cornejo-Bueno}, C. Casanova-Mateo, J. Sanz-Justo, P. Salvador-Gonz\'alez, S. Salcedo-Sanz, ``A novel Grouping Genetic Algorithm-Extreme Learning Machine Approach for Global Solar Radiation Prediction from Numerical Weather Models Inputs'', Solar Energy, vol. 132, pp. 129-142, 2016 (JCR 2016: 4.018)

\item \textbf{L. Cornejo-Bueno}, J. C. Nieto Borge, P. Garc\'aa-D\'iaz, G. Rodr\'iguez, S. Salcedo-Sanz, ``Significant Wave Height and Energy Flux Prediction for Marine Energy Applications: A Grouping Genetic Algorithm-Extreme Learning Machine Approach'', Renewable Energy, vol. 97, pp. 380-389, 2016 (JCR 2016: 4.357)

\item M. Dorado-Moreno, \textbf{L. Cornejo-Bueno}, P.A. Guti\'errez, L. Prieto, C. Herv\'as-Mart\'inez, S. Salcedo-Sanz, ``Robust Estimation of Wind Power Ramp Events with Reservoir Computing'', Renewable Energy, vol. 11, pp. 428-437, 2017 (JCR 2016: 4.357)

\item \textbf{L. Cornejo-Bueno}, C. Casanova-Mateo, J. Sanz-Justo, E. Cerro-Prada, S. Salcedo-Sanz, ``Efficient Low-Visibility Event Prediction at Airports using Machine-Learning Regression Techniques'', Boundary-Layer Meteorology, vol. 165, no. 2, pp. 349-370, 2017 (JCR 2016: 2.573)

\item \textbf{L. Cornejo-Bueno}, E.C. Garrido-Merch\'an, D. Hern\'andez-Lobato, S. Salcedo-Sanz, ``Bayesian Optimization of a Hybrid System for Robust Ocean Wave Features Prediction'', Neurocomputing, 2017 (JCR 2016: 3.317)

\item Z.M. Yaseena, R.C. Deo, A. Hilald, A.M. Abde, \textbf{L. Cornejo-Bueno}, S. Salcedo-Sanz, M.L. Nehdig, ``Predicting Compressive Strength of Lightweight Foamed Concrete using Extreme Learning Machine Model'', Advances in Engineering Software, 2017 (JCR 2016: 3.000)

\item S. Salcedo-Sanz, R.C. Deo, \textbf{L. Cornejo-Bueno}, C. Camacho-G\'omez, S. Ghimire, ``An Efficient Neuro-Evolutionary Hybrid Modelling Mechanism for the Estimation of Daily Global Solar Radiation in Sunshine State of Australia'', Applied Energy, vol. 209, pp. 79-94, 2017 (JCR 2016: 7.182)

\item \textbf{L. Cornejo-Bueno}, L. Cuadra, S. Jim\'enez-Fern\'andez, J. Acevedo-Rodrí\'iguez, L. Prieto, S. Salcedo-Sanz, ``Wind Power Ramp Events Prediction with Hybrid Machine Learning'', Energies, vol. 10, no. 11, pp. 1784-1811, 2017 (JCR 2016: 2.262)

\end{enumerate}

\section*{Papers in international conferences}
\begin{enumerate}
\item R. Mallol-Poyato, S. Jim\'enez-Fern\'andez, \textbf{L. Cornejo-Bueno}, P. D\'iaz-Villar and S. Salcedo-Sanz, ``Nested Evolutionary Algorithms for Joint Structure Design and Operation of Micro-grids under Variable Electricity Prices Scenarios'', 10th edition of INISTA, Madrid, Espa'na, pp. 114-118, 2015

\item P.A. Guti\'errez, J.C. Fern\'andez, M. P\'erez-Ortiz, \textbf{L. Cornejo-Bueno}, E. Alexandre-Cortizo, S. Salcedo-Sanz, and C. Herv\'as-Mart\'inez, ``Energy Flux Range Classification by using a Dynamic Window Autoregressive Model'',  13th International Work Conference on Artificial Neural Networks, IWANN 2015. Lecture Notes in Computer Science, vol. 9095, pp. 92-102, 2015

\item \textbf{L. Cornejo-Bueno}, A. Aybar-Ruiz, S. Jim\'enez-Fern\'andez, E. Alexandre, J. C. Nieto-Borge and S. Salcedo-Sanz, ``A Grouping Genetic Algorithm-Extreme Learning Machine Approach for Optimal Wave Energy Prediction'', IEEE World Congress on Computational Intelligence, Vancouver, Canad\'a, pp. 3817-3823, 2016

\item C. Camacho-G\'omez, R. Mallol-Poyato, S. Jim\'enez-Fern\'andez, \textbf{L. Cornejo-Bueno} and S. Salcedo-Sanz, ``Optimal Placement of Distributed Generation in Micro-grids with Binary and Integer-encoding Evolutionary Algorithms'', IEEE World Congress on Computational Intelligence, Vancouver, Canad\'a, pp. 3630-3637, 2016

\item \textbf{L. Cornejo-Bueno}, E. Garrido-Merch\'an, D. Hern\'andez-Lobato and S. Salcedo-Sanz, ``Bayesian Optimization of a Hybrid Prediction System for Optimal Wave Energy Estimation Problems'', IWANN, Cádiz, Spain, pp. 648-660, 2017

\item \textbf{L. Cornejo-Bueno}, A. Aybar-Ruiz, C. Camacho-G\'omez, L. Prieto, A. Barea-Ropero and S. Salcedo-Sanz, ``A Hybrid Neuro-Evolutionary Algorithm for Wind Power Ramp Events Detection'', IWANN, Cádiz, Spain, pp. 745-756, 2017

\end{enumerate}

\section*{Invited talks}
\begin{enumerate}
\item \textbf{L. Cornejo-Bueno}, A. Aybar-Ruiz, J. C. Nieto-Borge and S. Salcedo-Sanz, ``A New Hybrid GGA-ELM Approach for Significant Wave Height Prediction in Marine Energy Applications'', 15th EU/ME Workshop on Metaheuristic Applications, Madrid, Espa'na, 2015

\end{enumerate}

\section*{National conferences}
\begin{enumerate}
\item \textbf{Laura Cornejo-Bueno}, Carlos Camacho-G\'omez, Adri\'an Aybar-Ruiz, Luis Prieto and Sancho Salcedo-Sanz, ``Feature Selection with a Grouping Genetic Algorithm - Extreme Learning Machine Approach for Wind Power Prediction'', XI Congreso Espa'nol de Metaheur\'isticas, Algoritmos Evolutivos Y Bioinspirados (MAEB 2016), Salamanca, Espa'na, pp. 373-382, 2016

\end{enumerate}

\chapter*{Apendix B. Awards}
\begin{enumerate}
\item Accesit (second prize) of the Competition ``Thesis in 3 minutes'', for the branch of knowledge ``Engineering and Architecture''. Universidad Rey Juan Carlos and Universidad de Alcal\'a, July, 2016

\end{enumerate}

\part{Bibliography}

\end{document}